\pdfoutput=1

\documentclass[11pt]{article}

\usepackage[table,dvipsnames]{xcolor}

\usepackage[final]{acl}

\usepackage{booktabs,arydshln}
\makeatletter
\def\adl@drawiv#1#2#3{%
        \hskip.5\tabcolsep
        \xleaders#3{#2.5\@tempdimb #1{1}#2.5\@tempdimb}%
                #2\z@ plus1fil minus1fil\relax
        \hskip.5\tabcolsep}
\newcommand{\cdashlinelr}[1]{%
  \noalign{\vskip\aboverulesep
           \global\let\@dashdrawstore\adl@draw
           \global\let\adl@draw\adl@drawiv}
  \cdashline{#1}
  \noalign{\global\let\adl@draw\@dashdrawstore
           \vskip\belowrulesep}}
\makeatother

\usepackage{times}
\usepackage{latexsym}

\usepackage[T1]{fontenc}

\usepackage[utf8]{inputenc}

\usepackage{microtype}

\usepackage[most]{tcolorbox}

\usepackage{tikz}
\usetikzlibrary{bayesnet}

\usepackage[utf8]{inputenc} 
\usepackage[T1]{fontenc}    
\usepackage{hyperref}       
\usepackage{url}            
\usepackage{booktabs}       
\usepackage{amsfonts}       
\usepackage{nicefrac}       
\usepackage{multirow}
\usepackage{multirow, makecell}
\usepackage{tabularx}
\newcommand\setrow[1]{\gdef\rowmac{#1}#1\ignorespaces}
\newcommand\clearrow{\global\let\rowmac\relax}
\clearrow

\usepackage{times}
\usepackage{latexsym}
\usepackage{soul,color}
\usepackage{todonotes} 
\usepackage[normalem]{ulem}
\usepackage{tikz}
\usepackage{array}
\usepackage[T1]{fontenc}
\usepackage{wrapfig}

\usepackage{fixltx2e}

\usepackage{amsmath}
\usepackage{textgreek}
\usepackage{enumitem,bbding,etoolbox,calc,pifont}

\usepackage[capitalise]{cleveref}

\usepackage{soul}

\usepackage{xspace,mfirstuc,tabulary}

\newcommand\yelp{\textsc{Yelp}\xspace}
\newcommand\amazon{\textsc{Amazon}\xspace}

\newcommand\gyafc{\textsc{GYAFC}\xspace}
\newcommand\shakespeare{\textsc{Shakespeare}\xspace}
\newcommand\jfleg{\textsc{JFLEG}\xspace}
\newcommand\symb{\textsc{Sym}\xspace}

\newcommand{\TODOMirac}[1]{\textcolor{red}{\bf[*TODO*]}}

\usepackage[font=small,labelfont=bf]{caption}

\title{
Prompt-and-Rerank: A Method for Zero-Shot and Few-Shot Arbitrary Textual Style Transfer with Small Language Models
}

\author{
Mirac Suzgun\textsuperscript{*} \\
Stanford University \\
\footnotesize{\texttt{msuzgun@cs.stanford.edu}} \\\And
Luke Melas-Kyriazi\textsuperscript{*} \\
Oxford University \\
\footnotesize{\texttt{lukemk@robots.ox.ac.uk}} \\\And 
Dan Jurafsky \\
Stanford University \\
\footnotesize{\texttt{jurafsky@cs.stanford.edu}}
\\
}
\begin{document}
\maketitle
\begin{abstract}
\vspace{-1.0em}
We propose a method for arbitrary textual style transfer (TST)---the task of transforming a text into any given style---utilizing general-purpose pre-trained language models. 
Our method, \emph{Prompt-and-Rerank}, is based on a mathematical formulation of the TST task, decomposing it into three constituent components: \emph{textual similarity}, \emph{target style strength}, and \emph{fluency}. 
Specifically, our method first uses zero-shot or few-shot prompting to obtain a set of candidate generations in the target style, and then re-ranks these candidates according to a combination of the three components above. 
Empirically, our method enables small pre-trained language models to perform on par with state-of-the-art large-scale models while consuming two orders of magnitude less compute and memory.
Finally, we conduct a systematic investigation of the effect of model size and prompt design (e.g., prompt paraphrasing and delimiter-pair choice) on style transfer quality across seven diverse textual style transfer datasets.\footnote{Our code, data, and results are available at \url{https://github.com/suzgunmirac/prompt-and-rerank}}
\end{abstract}
\vspace{-1.5em}
\section{Introduction}
\vspace{-0.5em}
Textual style transfer (TST) refers to the task of transferring one stylistic aspect of a piece of text (e.g., sentiment attribute, formality, politeness, etc.) without changing its main semantic content, structure, or other attributes. 
Traditionally, the natural language generation (NLG) community has approached each instantiation of style transfer as a distinct task, designing and training specialized models on style-specific training corpora. For example, sentiment transfer has been studied extensively (\citet{li2018delete,sudhakar2019transforming,luo2019towards}, \emph{inter alia}). This paradigm has restricted TST research to a limited, simple set of style choices with parallel corpora, which can often be solved adequately by word replacement (e.g., replacing negative words with corresponding positive words for sentiment transfer). 

\begin{table}[t]
\small 
\centering
\scalebox{0.63}{
\begin{tabular}{c | p{0.63\textwidth}}
\toprule
\multicolumn{2}{c}{\textbf{Sentiment Transfer}} \\
\toprule
 \textcolor{gray}{\bf{Input}} & \textcolor{gray}{the food was \textcolor{cyan}{low quality} and thrown together quickly with \textcolor{cyan}{little} flavor.} \textcolor{OliveGreen}{[Neg]} \\
 \cdashlinelr{1-2}
 GPT-J-6B & the food was \textcolor{purple}{delicious} and \textcolor{purple}{prepared with a lot of care!} \textcolor{OliveGreen}{[Pos]} \\ \midrule
 \textcolor{gray}{\bf{Input}} & \textcolor{gray}{the staff is \textcolor{cyan}{awesome} and location is \textcolor{cyan}{right in the heart} of old town!} \textcolor{OliveGreen}{[Pos]} \\  
 \cdashlinelr{1-2}
 GPT-J-6B & the staff is \textcolor{purple}{terrible} and location is \textcolor{purple}{in the middle of nowhere!} \textcolor{OliveGreen}{[Neg]} \\ 
 \midrule
 \multicolumn{2}{c}{\textbf{Elizabethan $\to$ Modern English Translation}} \\
 \midrule
 \multirowcell{2}{\textcolor{gray}{\bf{Input}}} &  \textcolor{gray}{\textcolor{cyan}{thou hast quar rell'd} with a man \textcolor{cyan}{for coughing} in the street, because he \textcolor{cyan}{hath wakened thy} dog \textcolor{cyan}{that hath lain asleep} in the sun.} \textcolor{OliveGreen}{[Elizabethan]} \\  
 \cdashlinelr{1-2}
 \multirowcell{2}{GPT-J-6B} & \textcolor{purple}{you have cursed} a man because he\textcolor{purple}{'s coughed} in the street, because he \textcolor{purple}{has woken your} dog, \textcolor{purple}{which has been sleeping} in the sun. \textcolor{OliveGreen}{[ModernEng]} \\ 
  \midrule
 {\textcolor{gray}{\bf{Input}}} & \textcolor{gray}{if \textcolor{cyan}{thou art} fickle, what \textcolor{cyan}{dost thou} with him that is \textcolor{cyan}{renown'd} for faith?} \textcolor{OliveGreen}{[Elizabethan]} \\  
 \cdashlinelr{1-2}
 \multirowcell{2}{GPT-J-6B} & if \textcolor{purple}{you are} fickle, then what \textcolor{purple}{are you doing with that person} who is \textcolor{purple}{known for being reliable}? \textcolor{OliveGreen}{[ModernEng]} \\ 
 \midrule
 \multicolumn{2}{c}{\textbf{Fluency and Grammar Correction}} \\
 \midrule
 \textcolor{gray}{\bf{Input}} & \textcolor{gray}{\textcolor{cyan}{Forexample}, My cousin is \textcolor{cyan}{12years} old.} \textcolor{OliveGreen}{[Ungrammatical]} \\  
 \cdashlinelr{1-2}
 GPT-J-6B & \textcolor{purple}{For example}, My cousin is \textcolor{purple}{twelve years} old. \textcolor{OliveGreen}{[Grammatical]} \\ 
 \midrule
 \multicolumn{2}{c}{\textbf{Formality Transfer}} \\
 \midrule
 \textcolor{gray}{\bf{Input}} & \textcolor{gray}{\textcolor{cyan}{r u} talking about \textcolor{cyan}{ur} avatar?} \textcolor{OliveGreen}{[Informal]} \\  
 \cdashlinelr{1-2}
 GPT-J-6B & \textcolor{purple}{are you} talking about \textcolor{purple}{your} avatar? \textcolor{OliveGreen}{[Formal]} \\ 
\bottomrule
\end{tabular}
}
\caption{
Qualitative examples of few-shot style transfer on the \yelp, \shakespeare, \jfleg, and \gyafc datasets. Coupling off-the-shelf ``small'' language models with our prompt-and-reranking method enables us to perform arbitrary textual style transfer without any model training or prompt-tuning. Compared to the extremely large language models (viz., ones with more than 100 billion parameters) used by \citet{reif2021recipe}, our models obtain similar performance using almost two orders of magnitude less compute and memory.}
\label{tab:mainpage}
\end{table}

With the recent success of general-purpose language modeling (LM), it is, however, natural to ask whether one can tackle a more general formulation of style transfer: \emph{arbitrary} TST, in which one aims to transform a reference text into an arbitrary style specified by the user at inference-time. 

Inspired by the success of natural-language prompting in other domains~\citep{radford2019language,petroni2019language,brownetal2020,gao2021making}, we consider a prompting-based zero- and few-shot approach to arbitrary TST. Under this setup, we specify the desired type of style transfer problem using a natural-language prompt containing the source text (and optionally a few examples, in the few-shot case), and then use a pre-trained LM to generate the stylized target text. Thus, the source text may be transformed into any user-specified style without additional training or fine-tuning. 

Recent work~\cite{reif2021recipe} has found that extremely large language models (LLMs), namely the 175 billion-parameter GPT-3~\cite{brownetal2020} model and the proprietary 137 billion-parameter \emph{LLM} model, are capable of sentiment and formality transfer. 
However, language models at this scale are not accessible to most researchers and practitioners, even in inference-only settings, due to their large memory consumption and slow generation times.
Thus far, to the best of our knowledge, there has not been any research on the capabilities of reasonably-sized models for style transfer domain, nor any systematic study of how the precise construction of the prompt affects model performance.

Differently from past work, this present paper takes a first-principles approach to arbitrary TST using pretrained language models. We first mathematically formalize the task, showing how it can be formulated as the combination of \emph{textual similarity}, \emph{target style strength}, and \emph{fluency}. This framework naturally leads us to propose a new method for arbitrary TST, which we call ``\textbf{Prompt-and-Rerank}.'' Using this method, we demonstrate, for the first time, that it is possible to perform arbitrary TST using small language models; prior work indicated that only enormous (i.e., GPT-3-scale) language models were capable of this task. 

We summarize the main contributions and insights of this paper as follows: 
(i) We provide the first mathematical formalization of the arbitrary TST task. 
(ii) We propose Prompt-and-Rerank, a novel prompting-based method for arbitrary TST which follows naturally from our mathematical formulation. 
(iii) Our method matches and sometimes even exceeds state-of-the-art performance on arbitrary TST while using small language models such as GPT-2, which consume two orders of magnitude less memory and compute than prior work. 
(iv) We conduct a nuanced investigation of the influence of prompt design, such as task phrasing and delimiter-pair choice, on the quality of style transfer generations. 
(v) In order to encourage and facilitate further research in the area, we establish a set of benchmarks for arbitrary TST (including cleaned versions of the popular sentiment transfer datasets \amazon and \yelp) along with accompanying automatic evaluation metrics.

\section{Background and Related Work}
\vspace{-0.5em}

\textbf{Background.} TST is a long-standing problem in NLP which encompasses many popular sub-tasks, such as sentiment and formality transfer. Prior to the advent of large-scale pre-training in recent years, it was common practice to consider each of these sub-tasks separately, and to train separate models on different supervised datasets for each task. 
These models generally performed well within the limited scope of their task, but failed to generalize to new tasks or to texts outside of their training distribution. Here we show that the modern paradigm of pre-training large models and then prompting (or fine-tuning) them can be applied to many sub-tasks of TST in a unified, zero-shot manner, even with relatively small Transformers.

\textbf{Related Work.}
Traditional approaches to TST can be broadly categorized into two families. The first family involves identifying and replacing distinctive style-related phrases (\citet{li2018delete,sudhakar2019transforming,wu2019mask,madaan2020politeness,malmi2020unsupervised,reid2021lewis}, \emph{inter alia}). For example, \citet{madaan2020politeness} perform the task of politeness transfer by first identifying words with stylistic attributes using TF-IDF and then training a model to replace or augment these stylistic words with ones associated with the target attribute. In general, these approaches perform well for very simple style edits (e.g., negating a sentence by adding the word \textit{not}), but they struggle in scenarios that require more complex syntactic and semantic changes. 

The second family of approaches involves disentangling latent representations of style and content, such that a text can be encoded into a style-invariant representation and then decoded in a desired style~\cite{hu2017toward,shen2017style,fu2018style,luo2019towards}. For example, \citet{hu2017toward} encodes into and decodes from a style-agnostic latent space using a VAE alongside attribute discriminators. These approaches are often theoretically well-grounded, but they generally require large quantities of labeled data and struggle to scale beyond a small number of styles.

Differently from these two families, one recent work~\citep{reif2021recipe} uses enormous pre-trained language models to tackle TST, an idea motivated by the remarkable performance of pre-trained LMs in other areas of NLP \citep{radford2019language,devlinetal2019bert,yang2019xlnet,liuetal2019_roberta}. Specifically, they use \emph{LLM}, \emph{LLM-Dialog}, and GPT-3, each of which has over 100 billion parameters, to rewrite texts in a variety of styles. However, they perform minimal analysis of their prompting setup, deferring such analysis to future work, and they suggest that this prompting-based approach is only feasible with LLMs.\footnote{
    A note on \emph{terminology}: We shall refer to GPT-3~\citep{brownetal2020} and similar models with 100 billion model parameters as \emph{large} language models (LLMs), and the versions of GPT-2 and GPT-J---ranging from OpenAI's GPT-2-Small (117M) to EleutherAI's GPT-J-6B---which are two-to-three orders of magnitude smaller than GPT-3, as \emph{small} language models (SLMs).
}

This paper presents a novel prompt-and-rerank approach to the general task of textual style transfer using pre-trained language models. Alongside our method, we present the first systematic study of prompt formulation and model size for the task of textual style transfer. Contrary to expectations, using our method we find that even small LMs are able to effectively perform arbitrary style transfer. In fact we match the performance of \citet{reif2021recipe} on multiple datasets using two orders of magnitude less memory and compute.

\vspace{-0.5em}

\section{Method: Prompt-Based Arbitrary TST}
\vspace{-0.5em}
This section begins with a mathematical formalization of the task of textual style transfer.\footnote{Despite its important role in NLG, we are not aware of any prior formal statement of the (textual) style transfer problem. Here, we hope to solidify the problem formulation and illustrate the a connection between this problem formulation and the automatic metrics used in the field to evaluate TST models.} Our formalization elucidates the three underlying components of the task, namely \emph{text similarity}, \emph{target style strength}, and \emph{fluency}, and naturally leads us to  \emph{Prompt-and-Rerank}, our prompt-based re-ranking algorithm for solving TST.

\subsection{Problem Formulation}
\label{sec:problem_formulation}
Let $\boldsymbol{x} \in \Sigma^{*}$ denote a text over a vocabulary $\Sigma$, and $\mathcal{S}$ the set of all possible text style choices. Let us further use $\boldsymbol{x}^{(s_1)} \in \Sigma^{*}$ to denote a text $\boldsymbol{x}$ written in the style $s_1 \in \mathcal{S}$. Informally speaking, the goal of TST is to transfer the style of a text $\boldsymbol{x}^{(s_1)}$ (usually, a sentence) from $s_1$ to $s_2$ without changing the main semantic content of the text. We can formally express this transformation via a function $f: \Sigma^{*} \times \mathcal{S} \times \mathcal{S} \to \Sigma^{*}$, which takes an input text (say $\boldsymbol{x}^{(s_1)}$) and its corresponding style ($s_1)$, as well as a target style ($s_2$), and outputs a modified version of the input  written in the style of $s_2$ (namely, $\boldsymbol{\tilde{x}}^{(s_2)}$).\footnote{In cases where the original style of the input text might not be known a priori, one can either estimate the style of the input using a statistical classifier or assume that the input is written in a neutral style.} Ideally, we would want the generated output $\boldsymbol{\tilde{x}}^{(s_2)}=f(\boldsymbol{x}^{(s_1)}, s_1, s_2)$ to be ``close'' (both semantically and syntactically) to the ground-truth $\boldsymbol{x}^{(s_2)}$ as much as possible.

\begin{figure}[t]
  \centering
  \begin{tikzpicture}[scale=0.75, transform shape]
    \node[latent] (x) {$\boldsymbol{x}$} ; %
    \node[latent, right=of x] (s2) {$s_2$} ; %
    \node[latent, left=of x] (s1) {$s_1$} ; %
    \node[latent, below=of x, xshift=+1.0cm] (xs2) {$\boldsymbol{\tilde{x}}^{(s_2)}$} ; %
    \node[latent, below=of x, xshift=-1.0cm] (xs1) {$\boldsymbol{x}^{(s_1)}$} ; %
    \plate[inner sep=0.25cm, xshift=-0.12cm, yshift=0.12cm] {plate1} {(x) (s1) (s2) (xs1) (xs2)} {}; %
    \edge {x} {xs1} ; %
    \edge {x} {xs2} ; %
    \edge {s1} {xs1} ; %
    \edge {s2} {xs2} ; %
  \end{tikzpicture}
  \begin{tikzpicture}[scale=0.75, transform shape]
    \node[latent] (x) {$\boldsymbol{x}$} ; %
    \node[latent, right=of x] (s2) {$s_2$} ; %
    \node[latent, left=of x] (s1) {$s_1$} ; %
    \node[latent, below=of x, xshift=+1.0cm] (xs2) {$\boldsymbol{\tilde{x}}^{(s_2)}$} ; %
    \node[latent, below=of x, xshift=-1.0cm] (xs1) {$\boldsymbol{x}^{(s_1)}$} ; %
    \plate[inner sep=0.25cm, xshift=-0.12cm, yshift=0.12cm] {plate1} {(x) (s1) (s2) (xs1) (xs2)} {}; %
    \edge {x} {xs1} ; %
    \edge {xs1} {xs2} ; %
    \edge {s1} {xs1} ; %
    \edge {s2} {xs2} ; %
  \end{tikzpicture}
  \vspace{-0.5em}
  \caption{Two different but equally meaningful and valid interpretations of the textual style transfer task. Here $\boldsymbol{x}$ can be thought as the universal (abstract) meaning of a text, $\boldsymbol{x}^{(s_1)}$ a rewrite of $\boldsymbol{x}$ in the style of $s_1$. Depending on which graphical model one adheres to, $\boldsymbol{x}^{(s_2)}$ can be said to generated by $\boldsymbol{x}$ and $s_2$ (left model) or by $\boldsymbol{x}^{(s_1)}$ and $s_2$ (right model). In this paper, we follow the second interpretation.}
  \label{fig:tst_graphicalmodels}
  \vspace{1mm}
\end{figure}
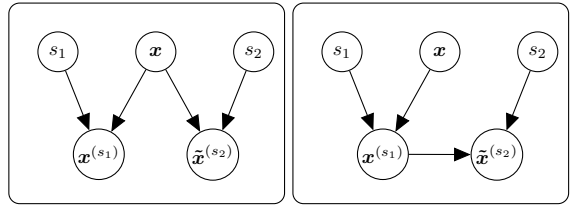

The graphical models depicted in Figure~\ref{fig:tst_graphicalmodels} provide two different ways of formulating the task of TST (and of machine translation for that matter). Both models have valid and meaningful implications and interpretations; the main generative difference between them is that the parents of $\boldsymbol{\tilde{x}}^{(s_2)}$ are $\boldsymbol{x}$ and $s_2$ in the former (left), whereas the parents of $\boldsymbol{\tilde{x}}^{(s_2)}$ are $\boldsymbol{x}^{(s_1)}$ and $s_2$ in the latter (right).

\begin{figure*}[t]
\centering
\includegraphics[width=0.96\textwidth]{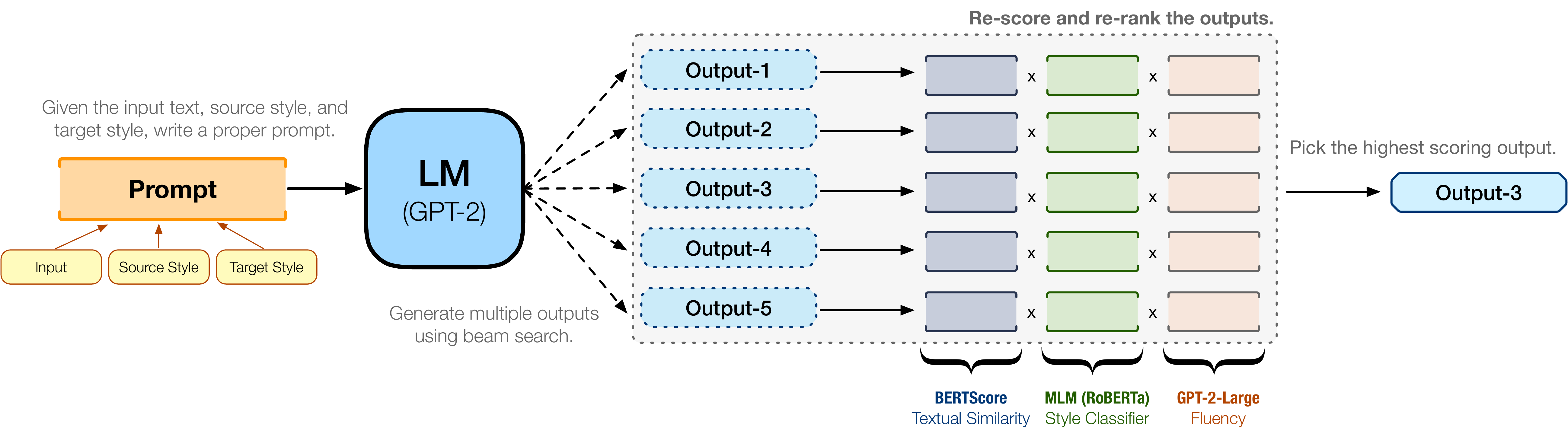}
\vspace*{-0.9em}
\caption{An illustration of our \emph{Prompt-and-Rerank} method. 
Given an input text and the style transformation, we first compose a prompt and feed it to a pretrained language model to generate multiple output texts---conditioned on the prompt---using beam search. We then re-score each candidate output along the three axes from \cref{eq:condprob}: \emph{textual similarity}, \emph{style transfer strength}, and \emph{fluency}. We choose the candidate with the highest re-ranked score as our final output.}
\label{fig:renraking_algorithm}
\end{figure*}

Due to the inherent difficulty of collecting diverse supervised data for arbitrary TST, most prior studies considered a simplified version of the task, wherein the source ($s_1$) and target ($s_2$) style choices are fixed beforehand. In this work, we consider a broad formulation of the task, make no assumptions about the source and target style choices a priori, and explain how one can leverage the power of off-the-shelf LMs to perform arbitrary TST. 

Given an input text $\boldsymbol{x}^{(s_1)}$ written in the style of $s_1$ and the target style $s_2$, we decompose the conditional likelihood of a generated output $\boldsymbol{\tilde{x}}^{(s_2)}$ into three terms:\footnote{We make use of the brackets ``$[\cdot]$'' only to group relevant terms (e.g., $\boldsymbol{x}^{(s_1)}, s_1$) together; they do not have any statistical significance in this context.}
{
\begin{flalign} \label{eq:condprob}
& p(\boldsymbol{\tilde{x}}^{(s_2)} \mid [\boldsymbol{x}^{(s_1)}, s_1], s_2) \\
& \quad = \frac{p(\boldsymbol{\tilde{x}}^{(s_2)}, [\boldsymbol{x}^{(s_1)}, s_1], s_2)}{p([\boldsymbol{x}^{(s_1)}, s_1], s_2)} \notag \\
& \quad  \propto p([\boldsymbol{x}^{(s_1)}, s_1], [\boldsymbol{\tilde{x}}^{(s_2)},s_2]) \notag \\
& \quad = p([\boldsymbol{x}^{(s_1)}, s_1] \mid[\boldsymbol{\tilde{x}}^{(s_2)},s_2]) \ p([\boldsymbol{\tilde{x}}^{(s_2)},s_2] ) \notag \\
& \quad = \underbrace{p([\boldsymbol{x}^{(s_1)}, s_1] \mid[\boldsymbol{\tilde{x}}^{(s_2)},s_2])}_{\text{textual similarity}}  \underbrace{p(s_2 \mid \boldsymbol{\tilde{x}}^{(s_2)} )}_{\text{transfer strength}} \underbrace{p(\boldsymbol{\tilde{x}}^{(s_2)})}_{\text{fluency}} \notag
\end{flalign}
}
The first term, $p([\boldsymbol{x}^{(s_1)}, s_1] \mid[\boldsymbol{\tilde{x}}^{(s_2)},s_2])$, can be thought as a measure of \emph{textual similarity} between the input text and the generated output. 
The second term, $p(s_2 \mid \boldsymbol{\tilde{x}}^{(s_2)} )$, measures the \emph{transfer strength} of the output (i.e., determines whether the output is written in the target style). The last term, $p(\boldsymbol{\tilde{x}}^{(s_2)})$, measures the overall \emph{fluency} of the output.

\subsection{\textit{Prompt-and-Rerank} for Arbitrary TST}
\label{subsec:reranking_alg}
The problem formulation above naturally leads us to a method for (textual) style transfer, which we denote \textit{Prompt-and-Rerank} (P\&R). 

The foundation of our method is use of \textit{prompt templates} to convert TST into a natural-language generation problem. Formally, we use a predefined template $\tau \in \mathcal{T}$ to convert an input text $\boldsymbol{x}^{(s_1)}$ and the desired style transformation (i.e., $s_1 \to s_2$) into a natural-language prefix $\tau(\boldsymbol{x}, s_1, s_2)$.
The template $\tau$ serves to not only contextualize the task for the model but also incorporate all the necessary conditional information (that is, input sentence, source style, and target style) in the input context. 
The precise design and composition of the templates is the topic of the following section (\S\ref{subsec:promt_choices}).\footnote{Additionally, in the few-shot case, where we have a number of few-shot exemplars, we convert these exemplars into meaningful prompts using the same template structure $\tau$ and prepend them to the main prompt.}

Next, we feed the prompt into a pre-trained LM (e.g., GPT-2) and use the model to generate $k$ different outputs $\boldsymbol{\tilde{x}}_{k}^{(s_2)}$ conditioned on the prompt, each sampled independently without updating any parameters of the model. These outputs are taken to be our \textit{candidate} outputs for re-ranking.

Finally, we re-rank our $k$ candidate outputs according to the decomposition in Equation~\ref{eq:condprob}:
{
\begin{flalign} \label{eq:reranking}
& p_{\text{reranking}} (\boldsymbol{\tilde{x}}^{(s_2)}_{i} | [\boldsymbol{x}^{(s_1)}, s_1], s_2) \\ 
& \propto p([\boldsymbol{x}^{(s_1)}, s_1] |[\boldsymbol{\tilde{x}}^{(s_2)}_{i},s_2]) p(s_2 | \boldsymbol{\tilde{x}}^{(s_2)}_{i}) p(\boldsymbol{\tilde{x}}^{(s_2)}_{i}). \notag
\end{flalign}
}
\noindent And finally, we pick the output $\boldsymbol{\tilde{x}}^{(s_2)}_{i} \in \tilde{\mathcal{X}}$ with the highest re-ranking score. (Figure~\ref{fig:renraking_algorithm} provides an abstract illustration of our re-ranking algorithm.)

All that remains is to describe how to calculate each term in the re-ranking pass. 
(i) To calculate the first (textual similarity) term, we avail ourselves of $\texttt{BERTScore}$ \citep{bertscore2020}, which utilizes pre-trained contextual embeddings from BERT to measure the cosine similarity between two texts.\footnote{Let us note that $\texttt{BERTScore}$ is a symmetric function, i.e., $\texttt{BERTScore}(\boldsymbol{x}^{(s_1)}, \boldsymbol{x}^{(s_2)}) = \texttt{BERTScore}(\boldsymbol{x}^{(s_2)}, \boldsymbol{x}^{(s_1)})$. Furthermore, we acknowledge that $\texttt{BERTScore}$ technically neglects the additional style information, but we believe that this is a reasonable simplification under our framework.} We presume  $p([\boldsymbol{x}^{(s_1)}, s_1] |[\boldsymbol{\tilde{x}}^{(s_2)},s_2])= \texttt{BERTScore}(\boldsymbol{x}^{(s_1)}, \boldsymbol{x}^{(s_2)})$. (ii) In order to calculate the second term, we deliberately turn a \emph{masked} LM (MLM), in our case a pre-trained RoBERTa model, into a \emph{style} classifier as follows: Given $\boldsymbol{\tilde{x}}^{(s_2)}_{i} \in \tilde{\mathcal{X}}^{(s_2)}$ and $\mathcal{S} =\{s_1, s_2\}$, we convert $\boldsymbol{\tilde{x}}^{(s_2)}_{i}$ into a ``fill-in-the-blank'' cloze statement via a pre-defined cloze template, that is, we rewrite it as ``The following text is $\texttt{<mask>}$: $[\boldsymbol{\tilde{x}}^{(s_2)}_{i}]$.'' We then query the MLM to predict the masked token,
\footnote{One limitation of this framework is that it assumes the styles are associated with distinct tokens in the vocabulary.}
but instead of looking at the probability distribution over the original model vocabulary, we restrict our attention to the elements in $\mathcal{S}$ and thus consider the likelihood of the missing token being $s_1$ or  $s_2$. We then normalize these probabilities by $l_1$-normalization and get a proper probability distribution for $p(s_2 | \boldsymbol{\tilde{x}}^{(s_2)})$.\footnote{Of course, a more sophisticated normalization technique can be employed in this setup, but this basic normalization method seemed to be sufficient in our experiments.} (iii) As for the last term, we use GPT-2-Large (774M) to determine the overall likelihood of each candidate text.\footnote{Given a text $x:=x_{1:t}$ of length $t$, we calculate its probability under a model $\theta$ as $p_{\theta}(x) = \prod_{i=1}^{t} p_{\theta} (x_i \mid x_{<i})$} (iv) Afterwards, we compute the score for each candidate by multiplying (i), (ii), and (iii) accordingly; re-rank all the candidates; and pick the one with the highest score as the final output.\footnote{Since the calculation of (iii) penalizes long sequences or sequences involving rare words, we also consider the re-ranking method in which we ignore the \emph{fluency} factor, assuming that the sentences generated by the models are always fluent, which, we are aware that, is a faulty assumption.}

Overall, our approach is model-agnostic, allowing pre-trained LMs to be used out-of-the-box. Furthermore, our experiments show that with well-designed prompts, one does \emph{not} need a massive language model for this approach to be successful. 

\begin{table*}[!t]
\small 
\centering
\vspace{-2mm}
\scalebox{0.90}{
\begin{tabular}{cccc}
\toprule
\bf{Dataset} & \bf{Styles} & \bf{Example Sentence-Pairs} & \bf{Test Set Size} \\
\midrule
Yelp Restaurant Reviews & Negative & ever since joes has changed hands it's just gotten worse and worse. & \multirow{2}{*}{1000} \\ 
\citep{zhang2015character} & Positive & ever since joes has changed hands it's gotten better and better. &  \\
\midrule
Amazon Product Reviews & Negative & if your bike had a kickstand on the plate it won't lock down. & \multirow{2}{*}{1000} \\ 
\citep{he2016ups} & Positive & if your bike had a kickstand on the plate it would lock down. &  \\
\midrule
GYAFC Formality Dataset & Informal & and so what if it is a rebound relationship for both of you? & \multirow{2}{*}{1000} \\ 
\citep{raotetreault2018gyafc} & Formal & what if it is a rebound relationship for both of you? & \\
\midrule
Shakespearean English Dataset & Elizabethan & is rosaline, whom thou didst love so dear, so soon forsaken? & \multirow{2}{*}{599} \\ 
\citep{xu2012paraphrasing} & Modern & have you given up so quickly on rosaline, whom you loved so much? & \\
\midrule
JFLEG Corpus & Ungrammatical & Forexample, My cousin is 12years old. & \multirow{2}{*}{747} \\ 
\citep{napolesetal2017jfleg} & Grammatical & For example, my cousin is 12 years old. & \\
\midrule
Symbolic Manipulation  & Symbolic & olive $>$ cat & \multirow{2}{*}{1000} \\ 
(\emph{Ours}) & English & olive is greater than cat  & \\
\bottomrule
\end{tabular}
}
\vspace{-0.6mm}
\caption{Overview of the textual style transfer datasets used in this paper.}
\label{tab:datasets}
\end{table*}

\vspace{-0.5em}

\section{Prompt Construction}
\label{sec:prompt_construction}
In practice, we found the specific syntax and semantics of the prompt template significantly impact model performance. Thus, we conducted a systematic investigation of the impact of different prompt design choices on the quality of TST generations.

\subsection{Delimiter-Pairs}
We experimented with ten different text boundary markers (delimiter pairs), which may be divided into two categories: those whose opening and closing markers are identical (known as \emph{indistinguishable} delimiters), and those whose markers are different (known as \emph{complementary} delimiters). 
Specifically, we considered two indistinguishable pairs (viz., quotes and dashes) and eight complementary pairs:
(1) curly brackets $\{ \cdot \}$, 
(2) square brackets $[ \cdot ]$, 
(3) angle brackets $\langle \cdot  \rangle$, 
(4) parentheses $( \cdot )$, 
(5) quotes $\text{"} \cdot \text{"}$, 
(6) dashes $\text{--} \cdot  \text{--}$, 
(7) triple angle brackets $\langle\langle\langle \cdot \rangle\rangle\rangle$, 
(8) bracket quotes $\rangle \text{ "} \cdot \text{"}$, 
(9) asterisk quotes $\text{* "} \cdot  \text{"}$, and 
(10) double curly bracket $\{ \{ \cdot \} \}$.\footnote{We use (8), (9), and (10) to emulate blockquotes, bullet points, and liquid tags in Markdown, respectively.} In their experiments, \citet{reif2021recipe} uses only curly brackets.\footnote{We hypothesized that the complementary delimiter-pairs might yield better results than the indistinguishable ones, since it is categorically easier for models to distinguish and understand where sentences start and end. We also speculated that delimiter-pairs that were more likely to be used as text-separators in the training data in various contexts (e.g., in code snippets) might yield better results.}

\subsection{Prompt Phrasing}
\label{subsec:promt_choices}
We considered four manually-written template formats $t \in \mathcal{T}$ for our discrete prompts:

\textbf{(a)} \emph{Vanilla}: ``Here is a text: $[d_1][\boldsymbol{x}^{(s_1)}][d_2]$ Here is a rewrite of the text, which is $[s_2]$: $[d_1]$'',

\textbf{(b)} \emph{Contrastive}: ``Here is a text, which is $[s_1]$: $[d_1][\boldsymbol{x}^{(s_1)}][d_2]$ Here is a rewrite of the text, which is $[s_2]$: $[d_1]$'',

\textbf{(c)} \emph{Negation-v1}: ``Here is a text, which is $[s_1]$: $[d_1][\boldsymbol{x}^{(s_1)}][d_2]$ Here is a rewrite of the text, which is not $[s_1]$: $[d_1]$'', and

\textbf{(d)} \emph{Negation-v2}: ``Here is a text, which is not $[s_2]$: $[d_1][\boldsymbol{x}^{(s_1)}][d_2]$ Here is a rewrite of the text, which is $[s_2]$: $[d_1]$''.

Note that $[d_1]$ and $[d_2]$ denote the opening and closing elements of the chosen delimiter-pair, respectively. In their experiments, \citet{reif2021recipe} exclusively made use of the \emph{vanilla} setting, which only specifies the target style ($s_2$) in the second half of the prompt; however, we initially speculated that providing useful information about the source style ($s_1$) and creating a clear contrast between the source and target styles in the prompt semantics might help pre-trained LMs to have a better understanding of the underlying nature of the task and improve their performance; hence, we decided to look at the \emph{contrastive} setting as well. As for the other two \emph{negation} templates, we wanted to test how specifying the source style as the negation of the target style (viz., $s1$:=``not $s_2$'') and vice versa might affect the model performance.\footnote{The last two formats might be useful especially when we do not have access to either the source or the target style.}

\paragraph{Example.} Finally, to make our prompting setup more concrete, let us give a concrete and brief example of how we formulate a prompt. We consider the \emph{contrastive} template with \textit{curly brackets} as our delimiter. 
If we have an input sentence $\boldsymbol{x}^{(s_1)}$=``I love \emph{The Sound of Music}; it is the best movie ever!!’’ with $s_1$=\textit{positive} and $s_2$=\textit{negative}, then the prompt under this template would be ``Here is a text, which is positive: \{I love \emph{The Sound of Music}; it is the best movie ever!!\} Here is a rewrite of the text, which is negative: \{'' The language model would then generate an output by autoregressively decoding after the last delimiter, to produce a sentence such as: ``I hate \emph{The Sound of Music}; it is the worst movie ever!!\}''\footnote{Table~\ref{tab:example_prompts} in the Appendix provides a complete set of examples of prompts used in each task.}

\vspace{-0.4em}
\subsection{Zero-Shot vs. Few-Shot Settings}
\vspace{-0.2em}
In recent years, LLMs, such as GPT-3, have proven themselves to be resourceful few-shot learners. In a few-shot learning setting, a model is presented with a small set of illustrative examples, oftentimes along with a natural-language prompt describing the task, and expected to understand the underlying task and make accurate predictions without performing any gradient updates to the weights of the model at inference time. We wanted to explore how the number of demonstrations affects the performance of our models. To that end, we also tested the performances of our models under the zero-shot and four-shot settings.

\vspace{-1mm}
\section{Experiments and Results}
\vspace{-1mm}
\subsection{Datasets}
\vspace{-1mm}
\label{subsec:datasets_and_tasks}
Differently from most previous work, which focused on single TST subtasks or datasets, we present experiments on a wide range of TST subtasks (also described in \cref{tab:datasets}): 
\begin{itemize}[noitemsep,topsep=0pt,leftmargin=6mm]
    \item \textbf{\yelp}: Sentiment transfer for Yelp reviews \citep{zhang2015character}.
    \item \textbf{\amazon}: Sentiment transfer for Amazon reviews \citep{li2018delete}.
    \item \textbf{\shakespeare}: \emph{Elizabethan}-to-modern translation for Shakespeare \citep{xu2012paraphrasing}.
    \item \textbf{\gyafc}: Formality transfer for Yahoo Answers responses \citep{li2018delete}.
    \item \textbf{\jfleg}: Grammar error correction for student essays ~\citep{napolesetal2017jfleg}.
    \item \textbf{\symb}: Symbol-to-natural-language translation on a new custom synthetic dataset.
\end{itemize}
In the initial stages of our research, we noticed that all of these datasets, with the exception of \symb (which is synthetic), contain various tokenization issues (e.g., sentences sometimes contain extra white-space or have their punctuation marks separated out by spaces). We did not wish these tokenization artifacts to diminish the quality of our generations from general-purpose LMs---neither did we want this issue to negatively impact our evaluation scheme. To that end, we used a simple text-cleaning procedure to clean the texts.\footnote{We release both the original and cleaned versions of the datasets alongside this paper to help facilitate future research. In the Appendix, we also present results for both the original and cleaned datasets.}

\subsection{Evaluation Metrics}
Prior studies on style and sentiment transfer have typically evaluated models across three dimensions: content/meaning preservation (textual similarity), style transfer strength, and fluency \cite{mir2019evaluating,briakou2021evaluating}. We note that remarkably, these dimensions correspond exactly to the criteria that appear in Equation \ref{eq:condprob} in \S\ref{sec:problem_formulation}. 

\textbf{Content Preservation.} BLEU~\citep{papineni2002bleu} is the standard metric for measuring semantic content preservation. We use the SacreBLEU (sBLEU) implementation~\citep{post2018sacrebleu} to compute both \emph{reference}-BLEU (\emph{r}-sBLEU) and \emph{self}-sBLEU (\emph{s}-sBLEU) scores. Whereas \emph{r}-sBLEU helps measure the distance of generated sentences from the ground-truth references, \emph{s}-sBLEU indicates the degree to which the model directly copies the source. 

\textbf{Transfer Strength.} 
In order to determine whether outputs generated by a TST model have the attributes of their target styles, we follow the standard classifier-based approach: we train a (binary) style classifier on the corpus of interest and use it to estimate the fraction of generated outputs whose styles match their target styles. 

\textbf{Fluency.} 
To measure the fluency of generated texts, we compute their average token-level perplexity (PPL) using a pre-trained LM (in our case, GPT-2-Large). We note that, whilst this PPL-driven approach has the advantage of being automated and practical, it still contains considerable drawbacks, including biases towards shorter texts.

\vspace{-1mm}
\subsection{Model Choices.} We used four GPT-2 models~\citep{radford2019language} of varying sizes (viz., GPT-2-Small (117M params), GPT-2-Medium (345M), GPT-2-Large (774M), and GPT-2-XL (1.6B)), GPT-Neo-1.3B~\citet{gptneo}, GPT-Neo-2.7B, and GPT-J-6B~\cite{gpt-j}. 
We highlight that none of these models were fine-tuned or prompt-tuned.

\begin{table}[!t]
\small
\centering
\scalebox{0.83}{
\begin{tabular}{l | >{\rowmac}c >{\rowmac}c  >{\rowmac}c  >{\rowmac}c  >{\rowmac}c <{\clearrow}}
\toprule
\bf{Model} & \bf{Acc} & \bf{\emph{r}-sBLEU} & \bf{\emph{s}-sBLEU} & \bf{PPL} \\
\toprule
\multicolumn{5}{c}{{\scriptsize \textsc{Supervised}}} \\
\cdashlinelr{1-5}
{[1]} CrossAlignment    &  0.73 &  7.8 &  18.3 & 217 \\
{[2]} BackTrans         &  0.95 &  2.0 &  46.5 & 158 \\
{[3]} MultiDecoder      &  0.46 &  13.0 & 39.4 & 373 \\
{[4]} DeleteOnly        &  0.85 &  13.4 & 33.9 & 182 \\
{[4]} DeleteAndRetrieve &  0.90 &  14.7 & 36.4 & 180 \\
{[5]} UnpairedRL        &  0.49 &  16.8 & 45.7 & 385 \\
{[6]} DualRL            &  0.88 &  25.9 & 58.9 & 133 \\
{[7]} ST (Multi-Class) & 0.86 &  26.4 &  63.0 &  175 \\
{[7]} ST (Conditional) &  0.93 & 22.9 &  52.8 &  223 \\
{[8]} B-GST &  0.81 & 21.6 & 46.5 & 158 \\
\cdashlinelr{1-5}
\multicolumn{5}{c}{\scriptsize \textsc{Zero- or Few-Shot Inference Only}} \\
\cdashlinelr{1-5}
{[9]} LLM\textsubscript{ Aug-0S-FirstChoice} &  0.85 &  5.3 &  9.2 &  33 \\
{[9]} LLM\textsubscript{ 5S-FirstChoice} &  0.93 &  6.7 &  11.2 &  43 \\
{[9]} LLM\textsubscript{ Aug-0S-Best-sBLEU}\textsuperscript{\textdagger} &  0.63 &  19.8 &  45.1 & 55 \\
{[9]} LLM\textsubscript{ 5S-Best-sBLEU}\textsuperscript{\textdagger} &  0.78 &  23.2 &  48.3 &  77 \\
\cdashlinelr{1-5}
\emph{Ours} (GPT-2-XL) &  0.87 &  14.8 &  28.7 &  65 \\
\emph{Ours} (GPT-J-6B) &  0.87 &  23.0 &  47.7 &  80 \\
\bottomrule
\end{tabular}
}
\vspace{-0.5em}
\caption{
A comparison of our Prompt-and-Rerank approach with supervised sentiment transfer methods and the ultra-large-scale prompting-based method of \citet{reif2021recipe} on the \yelp-clean dataset. In order to compare fairly against previous studies, we applied our data-cleaning code to their publicly-available outputs and re-computed all evaluation metrics.  References: [1]~\citep{shen2017style}, [2]~\citep{prabhumoye2018style}, [3]~\citep{fu2018style}, [4]~\citep{li2018delete}, [5]~\citep{xu2018unpaired}, [6]~\citep{luo2019dual}, [7]~\citep{dai2019style}, [8]~\citep{sudhakar2019transforming}, [9]~\citep{reif2021recipe}. Note on \textsuperscript{\textdagger}: We used sBLEU to choose the best candidate, as opposed to BLEU that was used originally in \citep{reif2021recipe}.
}
\label{tab:AllStar1}
\end{table}

\begin{table}[!t]
\small 
\centering
\vspace{3mm}
\scalebox{0.80}{
\begin{tabular}{c | >{\rowmac}c | >{\rowmac}c  >{\rowmac}c  >{\rowmac}c  >{\rowmac}c <{\clearrow}}
\toprule
\bf{Dataset} & \bf{Model} & \bf{Acc$^{*}$} & \bf{\emph{r}-sBLEU} & \bf{\emph{s}-sBLEU} & \bf{PPL} \\ 
\toprule
\amazon & GPT-2-XL & \bf{0.70} & 11.5 & 17.2 & 77 \\
\emph{P$\to$N} & GPT-J-6B & 0.65 & \bf{21.5} & 31.4 & \bf{70}  \\
\midrule
\amazon & GPT-2-XL & \bf{0.56} & 13.2 & 19.9 & \bf{50} \\
\emph{N$\to$P} & GPT-J-6B & 0.52 & \bf{19.3} & 29.3 & 58 \\
\midrule
\yelp & GPT-2-XL & \bf{0.87} & 14.8 & 28.7 & \bf{65} \\
\emph{P$\to$N} & GPT-J-6B & \bf{0.87} & \bf{23.0} & 47.7 & 80 \\
\midrule
\yelp & GPT-2-XL & \bf{0.72} & 12.0 & 25.3 & \bf{55} \\
\emph{N$\to$P} & GPT-J-6B & 0.65 & \bf{20.2} & 44.6 & 58 \\
\midrule
\textsc{Shake-} & GPT-2-XL & 0.39 & 18.9 & 38.4 & 90 \\
\textsc{speare} & GPT-J-6B & \bf{0.78} & \bf{21.}9 & 31.8 & \bf{81} \\
\midrule
\multirowcell{2}{\jfleg} & GPT-2-XL & 35.9 & \bf{74.8} & {91.5} & 76 \\
& GPT-J-6B & \bf{40.0} & 64.8 & 59.1 & \bf{48} \\
\midrule
\multirowcell{2}{\gyafc} & GPT-2-XL & 0.82 & 32.7 & 41.9 & \bf{58} \\
& GPT-Neo-1.3B & \bf{0.85} & \bf{36.4} & {49.6} & 68 \\
\midrule
\multirowcell{2}{\symb} & GPT-2-XL & 0.56 & 68.5 & - & - \\
& GPT-J-6B & \bf{0.74} & \bf{81.9} & - & - \\
\bottomrule
\end{tabular}
}
\vspace{-1.0em}
\caption{Four-shot performances of GPT-2-XL and GPT-J across all style transfer tasks, using curly brackets as delimiters. Full results with all models and delimiter pairs are shown in the appendix. \emph{P$\to$N} stands for the positive $\to$ negative direction, and vice-versa for \emph{N$\to$P}. $^{*}$ for \jfleg, GLEU score is used in place of accuracy to measure grammar correction performance. Note that the {\emph{r}-sBLEU} column is not bolded because it is not necessarily desirable to have a higher {\emph{r}-sBLEU}.}
\label{tab:AllStar2}
\end{table}

\vspace{-2mm}
\subsection{Results}
Here, we present a summary of our key findings. For our complete results, we encourage the reader to see the Appendix (especially, Tables \ref{tab:AmazonCleanZeroShot}-\ref{tab:yelp_clean_different_phrasing}).

\cref{tab:AllStar1} juxtaposes our results on \yelp with those of prior studies.
Despite not training or fine-tuning, our method is competitive with prior models that were designed and trained specifically for these tasks.
In fact, compared to supervised methods, our models almost always generate more fluent outputs, as measured by perplexity.
Compared to \citet{reif2021recipe}, who utilize the proprietary 137-billion-parameter \textit{LLM} (LaMDA), we compare on-par or favorably despite using much smaller models; we obtain better sBLEU scores than their ``FirstChoice'' setting (which uses a single output) and better accuracy scores than their ``BestBLEU'' oracle setting (which takes the best of 16 outputs, as measured by sBLEU score).

\cref{tab:AllStar2} presents a summary of our results across all seven TST datasets for GPT-2-XL and GPT-J. For full results including all models (GPT-2-Small to GPT-J), please refer to the Appendix. Broadly, we find that all models are capable of TST to a reasonable degree---with the larger models (e.g., GPT-2-XL, GPT-Neo-2.7B, GPT-J) often performing better than the smaller models. The only model that consistently performs poorly is GPT-2-Small: Its high \emph{s}-sBLEU and low accuracy indicate that it \emph{copies} long sections of the input (without changing its style) more often than the other models.

\begin{table}[!t]
\small 
\centering
\scalebox{0.90}{
\begin{tabular}{c | >{\rowmac}c  >{\rowmac}c  >{\rowmac}c  >{\rowmac}c <{\clearrow}}
\toprule
\bf{Setting} & \bf{Acc} & \bf{\emph{r}-sBLEU} & \bf{\emph{s}-sBLEU} & \bf{PPL} \\ \toprule
Vanilla     &  78.0      &  14.7      &  31.0   &  \bf{58.5} \\
Contrastive &  \bf{79.5} &  13.4      &  {27.0} &  59.5 \\
Negation-v1 &  66.5      &  13.4      &  28.1   &  67.5 \\
Negation-v2 &  52.0      &  \bf{18.0} &  40.6   &  69.0 \\
\bottomrule
\end{tabular}
}
\vspace{-0.8em}
\caption{Four-shot performances of GPT-2-XL on the \yelp-clean dataset under different prompting protocols. We show the average of scores from \emph{P$\to$N} and \emph{N$\to$P} directions. A full table with all models is included in the Appendix. Across all models, the vanilla and contrastive prompting protocols tend to yield the most favourable results.}
\label{tab:yelp_clean_main_prompt_phrasing}
\end{table}

\begin{table}[!t]
\small 
\centering
\vspace{2mm}
\scalebox{0.90}{
\begin{tabular}{c | >{\rowmac}c  >{\rowmac}c  >{\rowmac}c  >{\rowmac}c <{\clearrow}}
\toprule
\bf{Delim.} & \bf{Acc} & \bf{\emph{r}-sBLEU} & \bf{\emph{s}-sBLEU} & \bf{PPL} \\ \toprule
$\langle \cdot \rangle$                             &  49.5      &  17.4      &   40.8  &  45 \\
$\text{* "} \cdot  \text{"}$                        &  55.0      &  12.0      &   29.8  &  37 \\
$\rangle \text{ "} \cdot \text{"}$                  &  53.0      &  10.7      &   25.4  &  35 \\
$\{ \cdot \}$                                       &  59.5      &  10.0      &   23.6  &  35 \\
$\text{--} \cdot  \text{--}$                        &  54.5      &  6.4       &   16.5  &  \bf{24} \\
$\{ \{ \cdot \} \}$                                 &  50.5      &  \bf{18.3} &  {43.9} &  65 \\
$( \cdot )$                                         &  55.5      &  12.4      &   28.1  &  43 \\
$\text{"} \cdot \text{"}$                           &  \bf{60.5} &  8.8       &   20.4  &  31 \\
$[ \cdot ]$                                         &  58.0      &  11.4      &   27.4  &  41 \\
\bottomrule
\end{tabular}
}
\vspace{-0.9em}
\caption{Zero-shot performances of GPT-2-XL on the \yelp-clean dataset using different delimiter pairs. Full tables with all models for all datasets are included in the Appendix.}
\label{tab:yelp_clean_main_delimiter}
\vspace{-0.5em}
\end{table}

Looking at individual tasks, we recognize that there remains substantial room for improvement on the \jfleg task: Most models underperformed a simple baseline that copied the input text without making any changes. The baseline achieved 37.2 GLEU, better than all models except GPT-J (which obtained 40.0).
Finally, on our new synthetic task \symb, we found that GPT-J performed significantly better than the rest: It achieved $74\%$ accuracy\footnote{Accuracy is measured via exact-string-matching.} whereas no other model exceeded $60\%$ accuracy.\footnote{When the models failed to generate the correct output, we found that a common failure case was copying the input words correctly but using the wrong logic (e.g., generating ``less than'' instead of ``greater than'').}

\vspace{-0.5em}
\subsection{Further Analysis and Discussion}
\vspace{-0.5em}

\textbf{Contrastive prompting generally improves style transfer quality.} As shown in \cref{tab:yelp_clean_main_prompt_phrasing} (and \cref{tab:yelp_clean_different_phrasing} in the Appendix), amongst the four prompting protocols considered in this paper, contrastive prompting generally yielded the best accuracy, albeit not always the best sBLEU scores. 

\textbf{Delimiter-pair choice has a large impact on model performance.} 
Our systematic analysis of ten different delimiter-pairs shows that delimiter choice substantially affects the quality of generated outputs. 
Although there is not a single pair which performs best across all setups, certain delimiters, such as the curly brackets $\{ \cdot \}$, square brackets $[ \cdot ]$, parentheses $( \cdot )$, and quotes $\text{"} \cdot \text{"}$, yielded consistently better results on both \amazon and \yelp (see Tables~\ref{tab:AmazonOriginalZeroShot}-\ref{tab:YelpCleanZeroShot}).
We hypothesize that the strong performance of these markers is attributable to the fact that they are often used as text separators (or dividers) in different textual contexts, such as essays, dialogues, and code snippets, which compose part of the pre-training data of our models. 

\textbf{Re-ranking method improves overall performance.} 
We considered two re-ranking approaches, one in which we picked the generated output with the highest beam score and one in which we sampled three outputs from the model using beam search and then re-scored them according to three criteria discussed in \S\ref{subsec:reranking_alg}. As shown in Tables~\ref{tab:AmazonCleanFewShotFullResults} and \ref{tab:YelpCleanFewShotFullResults}, the re-ranking method can boost the sentiment accuracy by $10$-$30\%$. It often, but not always, leads to better sBLEU and fluency scores. Also, as Table~\ref{tab:promptrerank4thewin} illustrates, if we have access to a classifier trained on paired data, it might be more convenient to use it in our style transfer accuracy measurements, instead of an MLM as a proxy-classifier, in the re-ranking process, as it empirically leads to higher accuracy and sBLEU scores.

\textbf{Analysis of bias and transfer performance in opposite directions.} We find that pre-trained models have strong directional biases: 
None of the models performed the same when going in the negative$\to$positive (\emph{N$\to$P}) and positive$\to$negative (\emph{P$\to$N}) directions on \amazon and \yelp. We offer three possible explanations for this phenomenon: (i) The inherent differences in the linguistic difficulty of the tasks, (ii) the potential biases in pre-training dataset(s), and (iii) the poor quality of annotations in certain style transfer directions. Regarding (i), a qualitative inspection of the sentiment transfer datasets illustrates that in some cases, good \emph{P$\to$N} performance can be achieved by simply adding a negation (e.g., ``not'') into the text. Regarding (ii), it is possible that the web-scraped pre-training data of these models contains more sentences that resemble the task of changing the sentiment from positive to negative than the reverse direction during their pre-training periods. Qualitatively, the GPT-2 models appear adept at negation; therefore, it may not be surprising that these models yield better results in the \emph{P$\to$N} direction. As for (iii), our inspection of the ground-truth data reveals that it contains some noisy labels and incorrect input-output pairs. 

\begin{table}[!t]
\small 
\centering
\scalebox{0.85}{
\begin{tabular}{c | >{\rowmac}c  >{\rowmac}c >{\rowmac}c  >{\rowmac}c  >{\rowmac}c <{\clearrow}}
\toprule
\bf{Model} & \bf{Setting} & \bf{Acc} & \bf{\emph{r}-sBLEU} & \bf{\emph{s}-sBLEU} & \bf{PPL} \\ 
\toprule
\multirowcell{3}{\bf{GPT-2-XL} \\ (1558M)} & Top Choice & 0.63 & 13.7 & 20.3 & 65 \\
&  {P\&R}\textsubscript{RoBERTa} & 0.87 & 14.8 & 28.7 & 65 \\
\cdashlinelr{2-6}
&  {P\&R}\textsubscript{Oracle Cl.} & 0.95 & 16.8 & 33.4 & 63 \\
\midrule
\multirowcell{3}{\bf{GPT-J-6B} \\ (6B)} & Top Choice & 0.81 & 25.3 & 50.5 & 107 \\
&  {P\&R}\textsubscript{RoBERTa} & 0.87 & 23.0 & 47.7 & 80 \\
\cdashlinelr{2-6}
&  {P\&R}\textsubscript{Oracle Cl.} & 0.95 & 25.4 & 52.4 & 87 \\
\bottomrule
\end{tabular}
}
\vspace{-1em}
\caption{
Comparison of vanilla four-shot performance of GPT-2 XL and GPT-J-6B models on \yelp-clean ($P \to N$) under three settings: (1) choosing the output with the highest beam score (TC), (2) Prompt-and-Rerank with RoBERTa used as a zero-shot style classifier ({P\&R}$_{\text{RoBERTa}}$), and (3) Prompt-and-Rerank with an oracle style classifier trained on paired data ({P\&R}$_{\text{Oracle}}$). For full results in \yelp-clean, see Table~\ref{tab:YelpCleanFewShotFullResults}.
}
\vspace{-0.2em}
\label{tab:promptrerank4thewin}
\end{table}

\textbf{Limitations.} 
The primary limitation of our re-ranking method is that it involves generating multiple outputs from an autoregressive LM, which requires multiple forward passes. Additionally, our approach relies on having access to a pre-trained bi-directional MLM. Compared to a simple zero-shot approach, these elements could potentially add complexity to deploying this model in practice. 
\vspace{-0.5em}
\section{Conclusion}
\vspace{-0.5em}
In this paper, we propose a novel formal framework for textual style transfer. This framework naturally leads us to a new method, which we denote \textit{Prompt-and-Rerank}, that utilizes general-purpose pretrained language models to transform text into in arbitrary styles. 
In our experiments, we use our method to demonstrate that off-the-shelf, pre-trained ``small'' language models, such as GPT-2, can perform arbitrary textual style transfer, without any additional model fine-tuning or prompt-tuning. 
Additionally, we conduct an extensive investigation prompt phrasing and delimiter choice on transfer quality. 
In total, we hope that our work makes further research in this area more accessible to a broad set of researchers, both by alleviating the computational constraints of hundred-billion-parameter language models and by establishing a standard set of clean datasets for arbitrary text style transfer. 
\newpage
\section{Acknowledgments}
We would like to thank Dora Demszky, Esin Durmus, Tatsu Hashimoto, John Hewitt, Faisal Ladhak, Percy Liang, Nelson Liu, Shikhar Murty, Tol\'{u}l\d{o}p\d{\'{e}} \`{O}g\'{u}nr\d{\`{e}}m\'{i}, Isabel Papadimitriou, Ashwin Paranjape, Suproteem K. Sarkar, Stuart M. Shieber, Kyle Swanson, Rose Wang, and Tianyi Zhang for their help and support, constructive comments, and valuable suggestions. We also thank Sudha Rao for help with the navigation of the \gyafc dataset and Yunli Wang for sharing their data splits and classifiers for \gyafc that were used in their paper. Melas-Kyriazi gratefully acknowledges the support of the Rhodes Trust.  
\bibliography{bib/anthology,bib/custom}

\begin{thebibliography}{41}
\expandafter\ifx\csname natexlab\endcsname\relax\def\natexlab#1{#1}\fi

\bibitem[{Black et~al.(2021)Black, Gao, Wang, Leahy, and Biderman}]{gptneo}
Sid Black, Leo Gao, Phil Wang, Connor Leahy, and Stella Biderman. 2021.
\newblock \href {https://doi.org/10.5281/zenodo.5297715} {{GPT-Neo: Large Scale
  Autoregressive Language Modeling with Mesh-Tensorflow}}.

\bibitem[{Briakou et~al.(2021)Briakou, Agrawal, Tetreault, and
  Carpuat}]{briakou2021evaluating}
Eleftheria Briakou, Sweta Agrawal, Joel Tetreault, and Marine Carpuat. 2021.
\newblock {Evaluating the Evaluation Metrics for Style Transfer: A Case Study
  in Multilingual Formality Transfer}.
\newblock \emph{arXiv preprint arXiv:2110.10668}.

\bibitem[{Brown et~al.(2020)Brown, Mann, Ryder, Subbiah, Kaplan, Dhariwal,
  Neelakantan, Shyam, Sastry, Askell, Agarwal, Herbert{-}Voss, Krueger,
  Henighan, Child, Ramesh, Ziegler, Wu, Winter, Hesse, Chen, Sigler, Litwin,
  Gray, Chess, Clark, Berner, McCandlish, Radford, Sutskever, and
  Amodei}]{brownetal2020}
Tom~B. Brown, Benjamin Mann, Nick Ryder, Melanie Subbiah, Jared Kaplan,
  Prafulla Dhariwal, Arvind Neelakantan, Pranav Shyam, Girish Sastry, Amanda
  Askell, Sandhini Agarwal, Ariel Herbert{-}Voss, Gretchen Krueger, Tom
  Henighan, Rewon Child, Aditya Ramesh, Daniel~M. Ziegler, Jeffrey Wu, Clemens
  Winter, Christopher Hesse, Mark Chen, Eric Sigler, Mateusz Litwin, Scott
  Gray, Benjamin Chess, Jack Clark, Christopher Berner, Sam McCandlish, Alec
  Radford, Ilya Sutskever, and Dario Amodei. 2020.
\newblock \href
  {https://proceedings.neurips.cc/paper/2020/hash/1457c0d6bfcb4967418bfb8ac142f64a-Abstract.html}
  {{Language Models are Few-Shot Learners}}.
\newblock In \emph{Advances in Neural Information Processing Systems 33: Annual
  Conference on Neural Information Processing Systems 2020, NeurIPS 2020,
  December 6-12, 2020, virtual}.

\bibitem[{Dai et~al.(2019)Dai, Liang, Qiu, and Huang}]{dai2019style}
Ning Dai, Jianze Liang, Xipeng Qiu, and Xuan-Jing Huang. 2019.
\newblock {Style Transformer: Unpaired Text Style Transfer without Disentangled
  Latent Representation}.
\newblock In \emph{Proceedings of the 57th Annual Meeting of the Association
  for Computational Linguistics}, pages 5997--6007.

\bibitem[{Devlin et~al.(2019)Devlin, Chang, Lee, and
  Toutanova}]{devlinetal2019bert}
Jacob Devlin, Ming-Wei Chang, Kenton Lee, and Kristina Toutanova. 2019.
\newblock \href {https://doi.org/10.18653/v1/N19-1423} {{{BERT}: Pre-training
  of Deep Bidirectional Transformers for Language Understanding}}.
\newblock In \emph{Proceedings of the 2019 Conference of the North {A}merican
  Chapter of the Association for Computational Linguistics: Human Language
  Technologies, Volume 1 (Long and Short Papers)}, pages 4171--4186,
  Minneapolis, Minnesota. Association for Computational Linguistics.

\bibitem[{Fu et~al.(2018)Fu, Tan, Peng, Zhao, and Yan}]{fu2018style}
Zhenxin Fu, Xiaoye Tan, Nanyun Peng, Dongyan Zhao, and Rui Yan. 2018.
\newblock {Style Transfer in Text: Exploration and Evaluation}.
\newblock In \emph{Proceedings of the AAAI Conference on Artificial
  Intelligence}, volume~32.

\bibitem[{Gao et~al.(2021)Gao, Fisch, and Chen}]{gao2021making}
Tianyu Gao, Adam Fisch, and Danqi Chen. 2021.
\newblock {Making Pre-trained Language Models Better Few-shot Learners}.
\newblock In \emph{Proceedings of the 59th Annual Meeting of the Association
  for Computational Linguistics and the 11th International Joint Conference on
  Natural Language Processing (Volume 1: Long Papers)}, pages 3816--3830.

\bibitem[{He and McAuley(2016)}]{he2016ups}
Ruining He and Julian McAuley. 2016.
\newblock {Ups and Downs: Modeling the Visual Evolution of Fashion Trends with
  One-Class Collaborative Filtering}.
\newblock In \emph{proceedings of the 25th international conference on world
  wide web}, pages 507--517.

\bibitem[{Hu et~al.(2017)Hu, Yang, Liang, Salakhutdinov, and
  Xing}]{hu2017toward}
Zhiting Hu, Zichao Yang, Xiaodan Liang, Ruslan Salakhutdinov, and Eric~P Xing.
  2017.
\newblock {Toward Controlled Generation of Text}.
\newblock In \emph{International conference on machine learning}, pages
  1587--1596. PMLR.

\bibitem[{Jin et~al.(2021)Jin, Jin, Hu, Vechtomova, and Mihalcea}]{jin2021deep}
Di~Jin, Zhijing Jin, Zhiting Hu, Olga Vechtomova, and Rada Mihalcea. 2021.
\newblock {Deep Learning for Text Style Transfer: A Survey}.
\newblock \emph{Computational Linguistics}, pages 1--51.

\bibitem[{Krishna et~al.(2020)Krishna, Wieting, and
  Iyyer}]{krishna-etal-2020-reformulating}
Kalpesh Krishna, John Wieting, and Mohit Iyyer. 2020.
\newblock \href {https://doi.org/10.18653/v1/2020.emnlp-main.55}
  {{Reformulating Unsupervised Style Transfer as Paraphrase Generation}}.
\newblock In \emph{Proceedings of the 2020 Conference on Empirical Methods in
  Natural Language Processing (EMNLP)}, pages 737--762, Online. Association for
  Computational Linguistics.

\bibitem[{Lample et~al.(2019)Lample, Subramanian, Smith, Denoyer, Ranzato, and
  Boureau}]{lample2018multipleattribute}
Guillaume Lample, Sandeep Subramanian, Eric Smith, Ludovic Denoyer,
  Marc'Aurelio Ranzato, and Y-Lan Boureau. 2019.
\newblock \href {https://openreview.net/forum?id=H1g2NhC5KQ}
  {{Multiple-Attribute Text Rewriting}}.
\newblock In \emph{International Conference on Learning Representations}.

\bibitem[{Li et~al.(2018)Li, Jia, He, and Liang}]{li2018delete}
Juncen Li, Robin Jia, He~He, and Percy Liang. 2018.
\newblock \href {https://doi.org/10.18653/v1/N18-1169} {{Delete, Retrieve,
  Generate: a Simple Approach to Sentiment and Style Transfer}}.
\newblock In \emph{Proceedings of the 2018 Conference of the North {A}merican
  Chapter of the Association for Computational Linguistics: Human Language
  Technologies, Volume 1 (Long Papers)}, pages 1865--1874, New Orleans,
  Louisiana. Association for Computational Linguistics.

\bibitem[{Liu et~al.(2019)Liu, Ott, Goyal, Du, Joshi, Chen, Levy, Lewis,
  Zettlemoyer, and Stoyanov}]{liuetal2019_roberta}
Yinhan Liu, Myle Ott, Naman Goyal, Jingfei Du, Mandar Joshi, Danqi Chen, Omer
  Levy, Mike Lewis, Luke Zettlemoyer, and Veselin Stoyanov. 2019.
\newblock \href {http://arxiv.org/abs/1907.11692} {{RoBERTa: {A} Robustly
  Optimized {BERT} Pretraining Approach}}.
\newblock \emph{CoRR}, abs/1907.11692.

\bibitem[{Luo et~al.(2019{\natexlab{a}})Luo, Li, Yang, Zhou, Tan, Chang, Sui,
  and Sun}]{luo2019towards}
Fuli Luo, Peng Li, Pengcheng Yang, Jie Zhou, Yutong Tan, Baobao Chang, Zhifang
  Sui, and Xu~Sun. 2019{\natexlab{a}}.
\newblock {Towards Fine-Grained Text Sentiment Transfer}.
\newblock In \emph{Proceedings of the 57th Annual Meeting of the Association
  for Computational Linguistics}, pages 2013--2022.

\bibitem[{Luo et~al.(2019{\natexlab{b}})Luo, Li, Zhou, Yang, Chang, Sui, and
  Sun}]{luo2019dual}
Fuli Luo, Peng Li, Jie Zhou, Pengcheng Yang, Baobao Chang, Zhifang Sui, and
  Xu~Sun. 2019{\natexlab{b}}.
\newblock {A Dual Reinforcement Learning Framework for Unsupervised Text Style
  Transfer}.
\newblock \emph{arXiv preprint arXiv:1905.10060}.

\bibitem[{Madaan et~al.(2020)Madaan, Setlur, Parekh, P{\'o}czos, Neubig, Yang,
  Salakhutdinov, Black, and Prabhumoye}]{madaan2020politeness}
Aman Madaan, Amrith Setlur, Tanmay Parekh, Barnab{\'a}s P{\'o}czos, Graham
  Neubig, Yiming Yang, Ruslan Salakhutdinov, Alan~W Black, and Shrimai
  Prabhumoye. 2020.
\newblock {Politeness Transfer: A Tag and Generate Approach}.
\newblock In \emph{Proceedings of the 58th Annual Meeting of the Association
  for Computational Linguistics}, pages 1869--1881.

\bibitem[{Malmi et~al.(2020)Malmi, Severyn, and Rothe}]{malmi2020unsupervised}
Eric Malmi, Aliaksei Severyn, and Sascha Rothe. 2020.
\newblock {Unsupervised Text Style Transfer with Padded Masked Language
  Models}.
\newblock \emph{arXiv preprint arXiv:2010.01054}.

\bibitem[{Mir et~al.(2019)Mir, Felbo, Obradovich, and
  Rahwan}]{mir2019evaluating}
Remi Mir, Bjarke Felbo, Nick Obradovich, and Iyad Rahwan. 2019.
\newblock {Evaluating Style Transfer for Text}.
\newblock \emph{arXiv preprint arXiv:1904.02295}.

\bibitem[{Napoles et~al.(2015)Napoles, Sakaguchi, Post, and
  Tetreault}]{napoles2015ground}
Courtney Napoles, Keisuke Sakaguchi, Matt Post, and Joel Tetreault. 2015.
\newblock {Ground Truth for Grammatical Error Correction Metrics}.
\newblock In \emph{Proceedings of the 53rd Annual Meeting of the Association
  for Computational Linguistics and the 7th International Joint Conference on
  Natural Language Processing (Volume 2: Short Papers)}, pages 588--593.

\bibitem[{Napoles et~al.(2017)Napoles, Sakaguchi, and
  Tetreault}]{napolesetal2017jfleg}
Courtney Napoles, Keisuke Sakaguchi, and Joel Tetreault. 2017.
\newblock \href {https://aclanthology.org/E17-2037} {{JFLEG}: A fluency corpus
  and benchmark for grammatical error correction}.
\newblock In \emph{Proceedings of the 15th Conference of the {E}uropean Chapter
  of the Association for Computational Linguistics: Volume 2, Short Papers},
  pages 229--234, Valencia, Spain. Association for Computational Linguistics.

\bibitem[{Papineni et~al.(2002)Papineni, Roukos, Ward, and
  Zhu}]{papineni2002bleu}
Kishore Papineni, Salim Roukos, Todd Ward, and Wei-Jing Zhu. 2002.
\newblock {BLEU: A method for Automatic Evaluation of Machine Translation}.
\newblock In \emph{Proceedings of the 40th annual meeting of the Association
  for Computational Linguistics}, pages 311--318.

\bibitem[{Petroni et~al.(2019)Petroni, Rockt{\"a}schel, Riedel, Lewis, Bakhtin,
  Wu, and Miller}]{petroni2019language}
Fabio Petroni, Tim Rockt{\"a}schel, Sebastian Riedel, Patrick Lewis, Anton
  Bakhtin, Yuxiang Wu, and Alexander Miller. 2019.
\newblock {Language Models as Knowledge Bases?}
\newblock In \emph{Proceedings of the 2019 Conference on Empirical Methods in
  Natural Language Processing and the 9th International Joint Conference on
  Natural Language Processing (EMNLP-IJCNLP)}, pages 2463--2473.

\bibitem[{Post(2018)}]{post2018sacrebleu}
Matt Post. 2018.
\newblock \href {https://doi.org/10.18653/v1/W18-6319} {{A Call for Clarity in
  Reporting {BLEU} Scores}}.
\newblock In \emph{Proceedings of the Third Conference on Machine Translation:
  Research Papers}, pages 186--191, Brussels, Belgium. Association for
  Computational Linguistics.

\bibitem[{Prabhumoye et~al.(2021)Prabhumoye, Kocielnik, Shoeybi, Anandkumar,
  and Catanzaro}]{prabhumoye2021few}
Shrimai Prabhumoye, Rafal Kocielnik, Mohammad Shoeybi, Anima Anandkumar, and
  Bryan Catanzaro. 2021.
\newblock {Few-shot Instruction Prompts for Pretrained Language Models to
  Detect Social Biases}.
\newblock \emph{arXiv preprint arXiv:2112.07868}.

\bibitem[{Prabhumoye et~al.(2018)Prabhumoye, Tsvetkov, Salakhutdinov, and
  Black}]{prabhumoye2018style}
Shrimai Prabhumoye, Yulia Tsvetkov, Ruslan Salakhutdinov, and Alan~W Black.
  2018.
\newblock {Style Transfer Through Back-Translation}.
\newblock In \emph{Proceedings of the 56th Annual Meeting of the Association
  for Computational Linguistics (Volume 1: Long Papers)}, pages 866--876.

\bibitem[{Radford et~al.(2019)Radford, Wu, Child, Luan, Amodei, Sutskever
  et~al.}]{radford2019language}
Alec Radford, Jeffrey Wu, Rewon Child, David Luan, Dario Amodei, Ilya
  Sutskever, et~al. 2019.
\newblock {Language Models are Unsupervised Multitask Learners}.
\newblock \emph{OpenAI blog}, 1(8):9.

\bibitem[{Rao and Tetreault(2018)}]{raotetreault2018gyafc}
Sudha Rao and Joel Tetreault. 2018.
\newblock \href {https://doi.org/10.18653/v1/N18-1012} {{Dear Sir or Madam, May
  {I} Introduce the {GYAFC} Dataset: Corpus, Benchmarks and Metrics for
  Formality Style Transfer}}.
\newblock In \emph{Proceedings of the 2018 Conference of the North {A}merican
  Chapter of the Association for Computational Linguistics: Human Language
  Technologies, Volume 1 (Long Papers)}, pages 129--140, New Orleans,
  Louisiana. Association for Computational Linguistics.

\bibitem[{Reid and Zhong(2021)}]{reid2021lewis}
Machel Reid and Victor Zhong. 2021.
\newblock {LEWIS: Levenshtein Editing for Unsupervised Text Style Transfer}.
\newblock \emph{arXiv preprint arXiv:2105.08206}.

\bibitem[{Reif et~al.(2022)Reif, Ippolito, Yuan, Coenen, Callison-Burch, and
  Wei}]{reif2021recipe}
Emily Reif, Daphne Ippolito, Ann Yuan, Andy Coenen, Chris Callison-Burch, and
  Jason Wei. 2022.
\newblock \href {https://aclanthology.org/2022.acl-short.94} {{A Recipe for
  Arbitrary Text Style Transfer with Large Language Models}}.
\newblock In \emph{Proceedings of the 60th Annual Meeting of the Association
  for Computational Linguistics (Volume 2: Short Papers)}, pages 837--848,
  Dublin, Ireland. Association for Computational Linguistics.

\bibitem[{Shen et~al.(2017)Shen, Lei, Barzilay, and Jaakkola}]{shen2017style}
Tianxiao Shen, Tao Lei, Regina Barzilay, and Tommi Jaakkola. 2017.
\newblock {Style Transfer from Non-Parallel Text by Cross-Alignment}.
\newblock \emph{Advances in neural information processing systems}, 30.

\bibitem[{Sudhakar et~al.(2019)Sudhakar, Upadhyay, and
  Maheswaran}]{sudhakar2019transforming}
Akhilesh Sudhakar, Bhargav Upadhyay, and Arjun Maheswaran. 2019.
\newblock {"Transforming" Delete, Retrieve, Generate Approach for Controlled
  Text Style Transfer}.
\newblock In \emph{Proceedings of the 2019 Conference on Empirical Methods in
  Natural Language Processing and the 9th International Joint Conference on
  Natural Language Processing (EMNLP-IJCNLP)}, pages 3269--3279.

\bibitem[{Wang and Komatsuzaki(2021)}]{gpt-j}
Ben Wang and Aran Komatsuzaki. 2021.
\newblock {GPT-J-6B: A 6 Billion Parameter Autoregressive Language Model}.
\newblock \url{https://github.com/kingoflolz/mesh-transformer-jax}.

\bibitem[{Wang et~al.(2019)Wang, Wu, Mou, Li, and
  Chao}]{wang-etal-2019-harnessing}
Yunli Wang, Yu~Wu, Lili Mou, Zhoujun Li, and Wenhan Chao. 2019.
\newblock \href {https://doi.org/10.18653/v1/D19-1365} {{Harnessing Pre-Trained
  Neural Networks with Rules for Formality Style Transfer}}.
\newblock In \emph{Proceedings of the 2019 Conference on Empirical Methods in
  Natural Language Processing and the 9th International Joint Conference on
  Natural Language Processing (EMNLP-IJCNLP)}, pages 3573--3578, Hong Kong,
  China. Association for Computational Linguistics.

\bibitem[{Wang et~al.(2020)Wang, Wu, Mou, Li, and Chao}]{wang2020formality}
Yunli Wang, Yu~Wu, Lili Mou, Zhoujun Li, and Wenhan Chao. 2020.
\newblock {Formality Style Transfer with Shared Latent Space}.
\newblock In \emph{Proceedings of the 28th International Conference on
  Computational Linguistics}, pages 2236--2249.

\bibitem[{Wu et~al.(2019)Wu, Zhang, Zang, Han, and Hu}]{wu2019mask}
Xing Wu, Tao Zhang, Liangjun Zang, Jizhong Han, and Songlin Hu. 2019.
\newblock {"Mask and Infill": Applying Masked Language Model to Sentiment
  Transfer}.
\newblock \emph{arXiv preprint arXiv:1908.08039}.

\bibitem[{Xu et~al.(2018)Xu, Sun, Zeng, Zhang, Ren, Wang, and
  Li}]{xu2018unpaired}
Jingjing Xu, Xu~Sun, Qi~Zeng, Xiaodong Zhang, Xuancheng Ren, Houfeng Wang, and
  Wenjie Li. 2018.
\newblock {Unpaired Sentiment-to-Sentiment Translation: A Cycled Reinforcement
  Learning Approach}.
\newblock In \emph{Proceedings of the 56th Annual Meeting of the Association
  for Computational Linguistics (Volume 1: Long Papers)}, pages 979--988.

\bibitem[{Xu et~al.(2012)Xu, Ritter, Dolan, Grishman, and
  Cherry}]{xu2012paraphrasing}
Wei Xu, Alan Ritter, Bill Dolan, Ralph Grishman, and Colin Cherry. 2012.
\newblock {Paraphrasing for Style}.
\newblock In \emph{COLING}, pages 2899--2914.

\bibitem[{Yang et~al.(2019)Yang, Dai, Yang, Carbonell, Salakhutdinov, and
  Le}]{yang2019xlnet}
Zhilin Yang, Zihang Dai, Yiming Yang, Jaime Carbonell, Russ~R Salakhutdinov,
  and Quoc~V Le. 2019.
\newblock {XLNet: Generalized Autoregressive Pretraining for Language
  Understanding}.
\newblock \emph{Advances in neural information processing systems}, 32.

\bibitem[{Zhang et~al.(2020)Zhang, Kishore, Wu, Weinberger, and
  Artzi}]{bertscore2020}
Tianyi Zhang, Varsha Kishore, Felix Wu, Kilian~Q. Weinberger, and Yoav Artzi.
  2020.
\newblock \href {https://openreview.net/forum?id=SkeHuCVFDr} {{BERTScore:
  Evaluating Text Generation with BERT}}.
\newblock In \emph{International Conference on Learning Representations}.

\bibitem[{Zhang et~al.(2015)Zhang, Zhao, and LeCun}]{zhang2015character}
Xiang Zhang, Junbo Zhao, and Yann LeCun. 2015.
\newblock {Character-Level Convolutional Networks for Text Classification}.
\newblock \emph{Advances in neural information processing systems},
  28:649--657.

\end{thebibliography}
\bibliographystyle{acl_natbib}
\clearpage
\appendix
\section{Appendix}
\label{sec:appendix}

\subsection{Additional Details about Datasets}
Previous TST studies have often chosen to focus on particular subtasks (such as changing the sentiment of a text from positive to negative) or particular datasets (such as \yelp or \amazon). In contrast, in our experiments, we decided to focus on a variety of TST datasets, some of which are known and widely used datasets in the field and some of which are new and synthetic. In the first half of this section, we present and discuss these datasets.\footnote{We chose these datasets to broaden the semantic diversity of the TST tasks and to establish benchmarks for new TST studies. We share both the original and clean versions of some of the widely-used but poorly-tokenized datasets, such as \amazon and \yelp. In doing so, we hope to help address the recent call-to-action on reproducibility in TST from \citet{jin2021deep}; they encouraged researchers to share their data and evaluation codes in order to establish reliable benchmarks and facilitate easier comparison of new studies with existing work. We hope that our efforts will be a constructive step towards this goal.} 

\paragraph{\textbf{Yelp Sentiment Dataset.}} \yelp is a subset of the Yelp Review Polarity Dataset that was first used by \citet{zhang2015character} for a text classification task. It consists restaurant and other business reviews from Yelp, along with a label---either \emph{positive} or \emph{negative}---for each review. We used the version of the dataset that was curated by \citet{li2018delete} in our experiments. The test set contains 500 positive and 500 negative samples, with one human reference (ground-truth) for each sample.

\textbf{Amazon Sentiment Dataset.} \amazon is similar to \yelp in its nature, but it contains product reviews that were obtained from Amazon. Each review is labeled either \emph{positive} or \emph{negative}. As before, we used the version of the dataset that was used by \citet{li2018delete}. The test set contains 500 positive and 500 negative sentences, with one human reference output for each sample.

\textbf{Shakespearean English Dataset.} We additionally used a small subset of the dataset that was used by \citet{xu2012paraphrasing} originally for phrase-based machine translation, and experimented with ``translating'' sentences written in \emph{Elizabethan} English to \emph{modern} English. This small test set, which we call \shakespeare, contains 599 paired sentences from William Shakespeare's \emph{Romeo and Juliet}, written in Elizabethan and modern English.\footnote{All the input sentences in \shakespeare contain at least 10 and at most 25 words (inclusive).}

\textbf{GYAFC Formality Dataset.} Grammarly's Yahoo Answers Formality corpus (\gyafc; \citet{raotetreault2018gyafc}) contains paired \emph{informal} and \emph{formal} sentences. Following \citet{luo2019dual}, we used the samples from the ``Family \& Relationship'' (F\&R) domain and restricted our focus to the informal to formal direction. The test set contains 500 formal and 500 informal sentences.

\textbf{JFLEG Corpus.} The JHU FLuency-Extended GUG (\jfleg) Corpus was introduced by \citet{napolesetal2017jfleg} to train and evaluate models for automatic grammatical error correction. It contains paired \emph{grammatical} and \emph{ungrammatical} sentences (with three error types---namely, awkward, orthographic, and grammatical). In our experiments, we focused on the ungrammatical to grammatical direction and used the publicly available test set that contains 747 sentences.

\textbf{Symbolic Manipulation Task.} We designed this small synthetic dataset to investigate how skillful the off-the-shelf SLMs are at writing symbolic expressions as natural English-language sentences. This dataset contains 1,000 example pairs, in which each input sample is written in a symbolic form (as either ``$\alpha$ > $\beta$'' or ``$\alpha$ < $\beta$'', where $\alpha$ and $\beta$ are two different single words from the animal color, fruit, and number categories) and its corresponding output is basically the spoken utterance in English.

\emph{Remark.} We realized that the original versions of all the aforementioned real-world TST datasets contain various tokenization issues (for instance, sentences sometimes contain extra whitespaces or have their puntuation marks separated out by spaces). We did not wish these tokenization artifacts to diminish the quality of our generations. To that end, we used a simple text-cleaning procedure to clean the texts before feeding them to our models.\footnote{For the \amazon and \yelp dataset, we show the benefit of data-cleaning on overall performance. We also publicly release our text-cleaning code.}

\subsection{Additional Evaluation Metrics} \label{sec:appendix_automatic_evaluation}
Here, we describe in greater detail the standard automatic evaluation metrics used in the assessment of the performance of TST models. 

\textbf{Content Preservation.} The standard metric for measuring semantic content preservation (or textual similarity, as we call it) has been BLEU \citep{papineni2002bleu}: If reference (ground-truth) sentences are available, then \emph{reference}-BLEU scores are calculated by comparing model outputs to human-written ground-truth outputs using \textit{n}-grams. Some recent studies \citep{lample2018multipleattribute,dai2019style} further look at \emph{self}-BLEU scores, comparing model outputs to input sentences---this is particularly done when reference sentences are not directly available. In our evaluations, we primarily used the SacreBLEU metric \citep{post2018sacrebleu}---as SacreBLEU has been shown to be a more reliable and accessible metric than BLEU---and considered both \emph{reference}-SacreBLEU (\emph{r}-sBLEU) and \emph{self}-SacreBLEU (\emph{s}-sBLEU) scores.\footnote{We used the SacreBLEU metric implemented in Hugging Face's \emph{Metrics} library and lowered all the texts---both predictions and references---before calculating the scores.} When evaluating the performances of models on the \jfleg corpus, we also used the sentence-level GLEU metric \citep{napoles2015ground}, a variant of BLEU that was specifically designed for evaluating grammatical error correction (GEC) models.

\textbf{Transfer Strength.} To determine whether outputs generated by a TST model have the attributes of their target styles, the most common approach has been to train a (binary) classifier on the training set of the corpus of focus, where the sentences are taken as the inputs and their corresponding styles as the labels, and then to use this trained classifier to predict the percentage of the generated outputs for which the styles predicted by the model match their target styles.\footnote{This method of measuring transfer accuracy demands access to either paired data for training a classifier or a pre-trained classier that can accurately estimate the style of an input text. It is, therefore, difficult to measure transfer accuracy for arbitrary or unknown styles, because there may not be any specific data to train a classifier.} In our sentiment transfer experiments, we measured transfer strength (sentiment accuracy) by fine-tuning pre-trained RoBERTa classifiers \citep{liuetal2019_roberta} on the training data in each case. In our experiments on \shakespeare, we used the RoBERTa-based Shakespeare classifier of \citet{krishna-etal-2020-reformulating}. Finally, in our experiments on \gyafc, we fine-tuned a pre-trained RoBERTa classifier on a subset of F\&R examples.\footnote{We release all our fine-tuned classifiers on our codebase.}

\textbf{Fluency.} With the emergence of successful LMs at our disposal, most recent TST models measure the fluency of their generated texts by computing perplexity (PPL) via a pre-trained LM like GPT-2.\footnote{Early work used to measure fluency of sentences using an \emph{n}-gram (typically trigram) Kneser-Ney language model.} Whilst this PPL-driven approach has the advantage of being automated and practical, it still contains considerable drawbacks, among which biases towards short texts and more frequent tokens can be listed right away. In our evaluations, we reported the average token-level PPL of generated texts using GPT-2-Large (774M). 

\subsection{Full Results}

In the tables below, we include zero-shot results for the clean versions of \amazon (\cref{tab:AmazonCleanZeroShot}) and \yelp (\cref{tab:YelpCleanZeroShot}), as well as the original versions of \amazon (\cref{tab:AmazonOriginalZeroShot}) and \yelp (\cref{tab:YelpOriginalZeroShot}). 
We also include four-shot results for the clean versions of \amazon (\cref{tab:ShakespeareCleanFewShotFullResults}), \yelp (\cref{tab:AmazonCleanFewShotFullResults}), \shakespeare (\cref{tab:YelpCleanFewShotFullResults}), \jfleg (\cref{tab:JFLEG_clean_few_shot_results}), \gyafc (\cref{tab:GYAFC_clean_few_shot_results}), and \symb (\cref{tab:SymbolicManipulation_few_shot_results}).

\subsection{Further Discussion}

\textbf{Sentiment Transfer.} Table~\ref{tab:AmazonCleanFewShotFullResults} and Table~\ref{tab:YelpCleanFewShotFullResults} show the results for the clean versions of \amazon and \yelp, respectively. In terms of sentiment accuracy, GPT-2-XL yielded the best performance on both datasets, achieving 70\% (87\%) positive $\to$ negative accuracy and 56\% (72\%) negative $\to$ positive on \amazon (\yelp). In both cases, however, the sBLEU scores of GPT-2-XL were relatively lower than those of other models, indicating that it copied more from the source text. The GPT-Neo models had higher \emph{r}-sBLEU and \emph{s}-sBLEU scores than GPT-2-XL on both \amazon and \yelp, with only slightly worse accuracy scores. In the case of \yelp-clean especially, the GPT-Neo/J models achieved good balances of sentiment accuracy, textual similarity, and fluency.

\textbf{Shakespeare-to-Modern English Translation.} As shown in Table~\ref{tab:ShakespeareCleanFewShotFullResults}, model performance generally improves with model size, with GPT-J-6B achieving almost 80\% accuracy (according to the supervised classifier) and 21.9 \emph{r}-sBLEU. Also notable is the difference between GPT-2-Small's high \emph{s}-sBLEU score and low classifier accuracy, relative to the other models. Together, these indicate that the model copies large parts of the input text more often than the other GPT models.

\textbf{Formality Transfer and Grammatical Error Correction}. For \gyafc (Table~\ref{tab:GYAFC_clean_few_shot_results}), most models achieved accuracy scores above $80\%$, with increasing model size correlating with BLEU score. Notably, GPT-Neo-2.7B achieved an accuracy score of $81\%$ and and a \emph{r}-sBLEU score of 50 in the informal to formal direction. For \jfleg (Table~\ref{tab:JFLEG_clean_few_shot_results}), on the other hand, most models failed to outperform a simple baseline, which automatically copied the input text without making any changes. This baseline achieves a GLEU score of 37.2, better than all models except GPT-J (which obtains 40.0). Broadly, there remains substantial room for improvement on \jfleg. 

\textbf{Symbolic Manipulation.} Our final task is designed to measure the ability of these language models to copy and manipulate tokens under a refined synthetic experimental setup. With the exception of GPT-J, no model exceeded 60\% accuracy on this synthetic dataset. GPT-J, by contrast, achieved 74\% accuracy.

\subsection{Additional Qualitative Examples}

We provide additional qualitative examples from our language models in Tables~\ref{tab:generations_amaozon}-\ref{tab:generations_shakespeare}.

\subsection{Additional Related Work}
Here, we describe additional related work on different subtasks of textual style transfer that could not be included in the main component of the paper due to space constraints. 

These works can be broadly categorized into two families. The first family of approaches involves identifying and replacing distinctive style-related phrases (\citet{li2018delete,sudhakar2019transforming,wu2019mask,madaan2020politeness,malmi2020unsupervised,reid2021lewis}, \emph{inter alia}). For instance, \citet{madaan2020politeness} tackle the task of politeness transfer with a two-step text-editing approach, first identifying words with stylistic attributes using a $n$-gram TF-IDF method and then training a model to replace or augment these stylistic words with ones associated with the target attribute. Similarly, \citet{li2018delete} propose a simple approach to sentiment and style transfer based on the idea that these attributes can often be identified by certain distinctive phrases. They identify these phrases, replace them with phrases associated with the target attribute, and combine them with an RNN to improve the fluency of the output text. Recently, \citet{reid2021lewis} propose to minimize the Levenshtein edit-distance between source and target texts, using a fine-tuned LM to make targeted edits. In general, these approaches perform well for very simple types of style transfer (e.g., negation by adding the word \textit{not} to a sentence), but they struggle in scenarios that require more complex syntactic and semantic changes. 

The second family of approaches involves disentangling latent representations of style and content \citet{hu2017toward,shen2017style,fu2018style,luo2019towards,wang2020formality} seek to learn a style-invariant representation for a piece of text, such that it can then be decoded in an arbitrary style. 
For example, \citet{hu2017toward} encoded sentences into a style-agnostic space and then decode them in a style-specific manner using a variational autoencoder alongside attribute discriminators. \citet{shen2017style,fu2018style,dai2019style,wang-etal-2019-harnessing} improved upon this methodology through the use of cross-alignment, style embeddings, rule-based systems, and new architectures. While these approaches are often theoretically well-grounded, they generally require large quantities of labeled data and struggle with scaling beyond a small number of styles.

\subsection{Computational Details}

The computational cost of our experiments were quite low, as they only involve running inference on pre-trained models. All experiments were conducted on a single GPU. We usde an NVidia V100 for all experiments except those with GPT-J-6B, for which we used an RTX 8000 due to memory requirements. We estimate that all experiments for this paper consumed fewer than 30 GPU-days. 

\subsection{License Details}

We will release all code for this experiment under an open-source license (MIT License). 

\subsection{Language Details}

All datasets used for this paper are in English. 

\subsection{Ethical Considerations}

Our work aims to advance the state of research on the task of arbitrary textual style transfer. As with many NLP applications, these methods may be used for negative purposes by malicious actors. For example, it would be possible to conceive of an instantiation of arbitrary textual style transfer which converts a non-sensationalist news headline into a sensationalist news headline, or one that converts a non-offensive piece of text into an offensive piece of text, in order to achieve a malicious goal. 

Our work also involves pretrained general-purpose language models, which bring up less-obvious ethical considerations than those discussed above. Since these language models are trained on text scraped from the web, they have acquired some of the biases present in web text. Such biases may be extracted by certain forms of prompting; recent work \citep{prabhumoye2021few} suggests that few-shot prompts can be used to detect social biases in pretrained language models. A large body of work is dedicated to understanding and de-biasing these models, but it is not the subject of our present work.

\newpage
\begin{table*}[h]
\small 
\centering
\scalebox{0.88}{
\begin{tabular}{c | p{0.95\textwidth} }
\toprule
\bf{Dataset} & \bf{\quad \quad\quad\quad\quad \quad\quad\quad\quad \quad \quad\quad \textcolor{gray}{[Few-Shot Examples]} and \textcolor{teal}{[Test-Time Input]}} \\ 
\toprule
 \multirowcell{7}{\textbf{\textcolor{black}{\amazon}}} & 
 \noindent
\textcolor{gray}{Here is a text, which is positive: \{very small but it works great in the car.\} Here is a rewrite of the text, which is negative: \{very small and it works terribly in the car.\}}
\textbackslash{n}
\textcolor{gray}{\#\#\#}
\textbackslash{n}
\textcolor{gray}{Here is a text, which is positive: \{i really loved it and will use it alot.\} Here is a rewrite of the text, which is negative: \{i really disliked it and will not use it again.\}}
\textbackslash{n}
\textcolor{gray}{\#\#\#}
\textbackslash{n}
\textcolor{gray}{Here is a text, which is positive: \{it gets the job done and for the price you can t beat it.\} Here is a rewrite of the text, which is negative: \{it does not work well and it was expensive.\}}
\textbackslash{n}
\textcolor{gray}{\#\#\#}
\textbackslash{n}
\textcolor{gray}{Here is a text, which is negative: \{i will never buy anything from this brand again.\} Here is a rewrite of the text, which is positive: \{i will buy from this brand again.\}}
\textbackslash{n}
\textcolor{gray}{\#\#\#}
\textbackslash{n}
\textcolor{teal}{Here is a text, which is negative: \{if your bike had a kickstand on the plate it won't lock down. \} Here is a rewrite of the text, which is positive: \{ }
 \\  
 \midrule
 \multirowcell{8}{\textbf{\textcolor{black}{\yelp}}} & 
 \noindent
\textcolor{gray}{Here is a text, which is negative: \{this place is awful!\} Here is a rewrite of the text, which is positive: \{this place is amazing!\}}
\textbackslash{n}
\textcolor{gray}{\#\#\#}
\textbackslash{n}
\textcolor{gray}{Here is a text, which is positive: \{definitely will buy another pair of socks from this store--they have the best socks ever\} Here is a rewrite of the text, which is negative: \{definitely will NOT buy another pair of socks from this store--they have the worst socks ever\}}
\textbackslash{n}
\textcolor{gray}{\#\#\#}
\textbackslash{n}
\textcolor{gray}{Here is a text, which is negative: \{my wife and i were disappointed by the quality of the service--also, the food was pretty tasteless\} Here is a rewrite of the text, which is positive: \{my wife and i were impressive by the quality of the service--also, the food was pretty delicious\}}
\textbackslash{n}
\textcolor{gray}{\#\#\#}
\textbackslash{n}
\textcolor{gray}{Here is a text, which is positive: \{i loved their black tea and hot chocolate selections!\} Here is a rewrite of the text, which is negative: \{i hated their black tea and hot chocolate selections!\}}
\textbackslash{n}
\textcolor{gray}{\#\#\#}
\textbackslash{n}
\textcolor{teal}{Here is a text, which is positive: \{it's small yet they make you feel right at home.\} Here is a rewrite of the text, which is negative: \{ } 
\\
\midrule
 \multirowcell{8}{\textbf{\textcolor{black}{\shakespeare}}} & 
 \noindent
\textcolor{gray}{Here is a text, which is written in old English: \{what hast thou there?\} Here is a rewrite of the text, which is written in modern English: \{what have you got there?\}}
\textbackslash{n}
\textcolor{gray}{\#\#\#}
\textbackslash{n}
\textcolor{gray}{Here is a text, which is written in old English: \{what say'st thou, my dear nurse?\} Here is a rewrite of the text, which is written in modern English: \{what did you say, my dear nurse?\}}
\textbackslash{n}
\textcolor{gray}{\#\#\#}
\textbackslash{n}
\textcolor{gray}{Here is a text, which is written in old English: \{and how doth she?\} Here is a rewrite of the text, which is written in modern English: \{and how is she doing?\}}
\textbackslash{n}
\textcolor{gray}{\#\#\#}
\textbackslash{n}
\textcolor{gray}{Here is a text, which is written in old English: \{talk not to me, for i'll not speak a word.\} Here is a rewrite of the text, which is written in modern English: \{don't talk to me, because i won't answer you.\}}
\textbackslash{n}
\textcolor{gray}{\#\#\#}
\textbackslash{n}
\textcolor{teal}{Here is a text, which is old English: \{as mine on hers, so hers is set on mine, and all combined, save what thou must combine by holy marriage.\} Here is a rewrite of the text, which is modern English: \{} \\
\midrule
 \multirowcell{8}{\textbf{\textcolor{black}{\gyafc}}} & \textbackslash{n}
\textcolor{gray}{Here is a text, which is informal: \{sorry but donnt know if i can do this alone.\} Here is a rewrite of the text, which is formal: \{I am sorry, but I don't know if I can do this alone.\}}
\textbackslash{n}
\textcolor{gray}{\#\#\#}
\textbackslash{n}
\textcolor{gray}{Here is a text, which is formal: \{i am going to ask him to come to the concert with me, and i hope he accepts my invitation.\} Here is a rewrite of the text, which is informal: \{gonna ask him to come to the concert with me and hope he says yes :)\}}
\textbackslash{n}
\textcolor{gray}{\#\#\#}
\textbackslash{n}
\textcolor{gray}{Here is a text, which is informal: \{that sucks man but u gotta move on\} Here is a rewrite of the text, which is formal: \{that is unfortunate, but you need to move on\}}
\textbackslash{n}
\textcolor{gray}{\#\#\#}
\textbackslash{n}
\textcolor{gray}{Here is a text, which is formal: \{and i am sorry that you and your girlfriend broke up last week.\} Here is a rewrite of the text, which is informal: \{and im sorry that u and ur girlfriend broke up last week...\}}
\textbackslash{n}
\textcolor{gray}{\#\#\#}
\textbackslash{n}
\textcolor{teal}{Here is a text, which is formal: \{i mean that you have to really be her friend.\} Here is a rewrite of the text, which is informal: \{} \\
\midrule
 \multirowcell{7}{\textbf{\textcolor{black}{\jfleg}}} & \textbackslash{n}
\textcolor{gray}{Here is a text, which is ungrammatical: \{There are several reason.\} Here is a rewrite of the text, which is grammatical: \{There are several reasons.\}}
\textbackslash{n}
\textcolor{gray}{\#\#\#}
\textbackslash{n}
\textcolor{gray}{Here is a text, which is ungrammatical: \{To my surprize nothing happened.\} Here is a rewrite of the text, which is grammatical: \{To my surprise, nothing happened.\}}
\textbackslash{n}
\textcolor{gray}{\#\#\#}
\textbackslash{n}
\textcolor{gray}{Here is a text, which is ungrammatical: \{This is important thing.\} Here is a rewrite of the text, which is grammatical: \{This is an important thing.\}}
\textbackslash{n}
\textcolor{gray}{\#\#\#}
\textbackslash{n}
\textcolor{gray}{Here is a text, which is ungrammatical: \{Water is needed for alive.\} Here is a rewrite of the text, which is grammatical: \{Water is necessary to live.\}}
\textbackslash{n}
\textcolor{gray}{\#\#\#}
\textbackslash{n}
\textcolor{teal}{Here is a text, which is ungrammatical: \{New and new technology has been introduced to the society.\} Here is a rewrite of the text, which is grammatical: \{} \\
\midrule
 \multirowcell{7}{\textbf{\textcolor{black}{\symb}}} & \noindent
\textcolor{gray}{Here is a text, which is symbolic: \{apple > seven\} Here is a rewrite of the text, which is English: \{apple is greater than seven\}}
\textbackslash{n}
\textcolor{gray}{\#\#\#}
\textbackslash{n}
\textcolor{gray}{Here is a text, which is symbolic: \{tiger < robin\} Here is a rewrite of the text, which is English: \{tiger is less than robin\}}
\textbackslash{n}
\textcolor{gray}{\#\#\#}
\textbackslash{n}
\textcolor{gray}{Here is a text, which is symbolic: \{teal > green\} Here is a rewrite of the text, which is English: \{teal is greater than green\}}
\textbackslash{n}
\textcolor{gray}{\#\#\#}
\textbackslash{n}
\textcolor{gray}{Here is a text, which is symbolic: \{apple < dog\} Here is a rewrite of the text, which is English: \{apple is less than dog\}}
\textbackslash{n}
\textcolor{gray}{\#\#\#}
\textbackslash{n}
\textcolor{teal}{Here is a text, which is symbolic: \{yellow > gray\} Here is a rewrite of the text, which is English: \{} \\
\bottomrule
\end{tabular}
}
\caption{A complete list of example-prompts used in our few-shot experiments. Here, the color \textcolor{gray}{gray} is used to highlight the examples used in our setups and the color \textcolor{teal}{teal} an example test-time input in each specific TST task.}
\label{tab:example_prompts}
\end{table*}

\newpage
\begin{table*}[!h]
\small 
\centering
\scalebox{0.85}{
\begin{tabular}{c | >{\rowmac}c | >{\rowmac}c  >{\rowmac}c  >{\rowmac}c  >{\rowmac}c | >{\rowmac}c  >{\rowmac}c  >{\rowmac}c  >{\rowmac}c <{\clearrow}}
\toprule
&  & \multicolumn{4}{c|}{ \bf{Positive $\to$ Negative} } & \multicolumn{4}{c}{ \bf{Negative $\to$ Positive} } \\
\bf{Model} &  \bf{Delimiter-Pair} & \bf{Acc} & \bf{\emph{r}-sBLEU} & \bf{\emph{s}-sBLEU} & \bf{PPL} & \bf{Acc} & \bf{\emph{r}-sBLEU} & \bf{\emph{s}-sBLEU} & \bf{PPL} \\ 
\midrule
    \multirowcell{10}{\bf{GPT-2-Small} \\ (117M)} & $\langle \cdot \rangle$ & 0.35 & 12.4 & 22.7 & 34 & 0.19 & 11.4 & 23.6 & 33 \\
    & $\text{* "} \cdot  \text{"}$ & 0.43 & 9.0 & 15.9 & 42 & 0.24 & 7.3 & 14.5 & 40 \\
    & $\rangle \text{ "} \cdot \text{"}$ & 0.46 & 6.6 & 11.3 & 29 & 0.23 & 6.8 & 14.1 & 30 \\
    & $\{ \cdot \}$ & 0.33 & 14.1 & 26.4 & 35 & 0.18 & 15.0 & 31.3 & 39 \\
    & $\text{--} \cdot  \text{--}$ & 0.40 & 6.8 & 12.6 & 29 & 0.17 & 6.5 & 13.8 & 26 \\
    & $\{ \{ \cdot \} \}$ & 0.36 & 27.0 & 49.7 & 85 & 0.20 & 27.0 & 56.0 & 94 \\
    & $( \cdot )$ & 0.35 & 18.1 & 32.7 & 54 & 0.18 & 17.6 & 38.2 & 59 \\
    & $\text{"} \cdot \text{"}$ & 0.45 & 8.2 & 14.2 & 32 & 0.21 & 8.4 & 16.2 & 33 \\
    & $[ \cdot ]$ & 0.35 & 18.9 & 35.5 & 60 & 0.21 & 14.3 & 29.3 & 43 \\
    & $\langle\langle\langle \cdot \rangle\rangle\rangle$ & 0.42 & 6.4 & 12.1 & 24 & 0.19 & 6.7 & 14.1 & 26 \\
\midrule 
    \multirowcell{10}{\bf{GPT-2-Medium} \\ (345M)} & $\langle \cdot \rangle$ & 0.42 & 21.9 & 37.9 & 67 & 0.27 & 23.2 & 45.0 & 72 \\
    & $\text{* "} \cdot  \text{"}$ & 0.46 & 11.1 & 20.0 & 45 & 0.31 & 7.8 & 15.2 & 32 \\
    & $\rangle \text{ "} \cdot \text{"}$ & 0.45 & 13.4 & 22.4 & 43 & 0.29 & 6.2 & 13.4 & 27 \\
    & $\{ \cdot \}$ & 0.44 & 21.6 & 38.2 & 73 & 0.26 & 19.1 & 37.1 & 67 \\
    & $\text{--} \cdot  \text{--}$ & 0.63 & 4.2 & 7.0 & 22 & 0.31 & 3.7 & 7.6 & 21 \\
    & $\{ \{ \cdot \} \}$ & 0.45 & 25.3 & 43.2 & 69 & 0.27 & 20.2 & 39.8 & 67 \\
    & $( \cdot )$ & 0.49 & 19.4 & 32.4 & 69 & 0.31 & 18.1 & 35.5 & 69 \\
    & $\text{"} \cdot \text{"}$ & 0.47 & 11.3 & 19.2 & 35 & 0.28 & 9.2 & 17.5 & 34 \\
    & $[ \cdot ]$ & 0.54 & 17.1 & 28.6 & 63 & 0.32 & 13.1 & 26.3 & 52 \\
    & $\langle\langle\langle \cdot \rangle\rangle\rangle$ & 0.47 & 14.2 & 25.3 & 43 & 0.28 & 10.8 & 21.1 & 35 \\
\midrule 
    \multirowcell{10}{\bf{GPT-2-Large} \\ (774M)} & $\langle \cdot \rangle$ & 0.38 & 25.5 & 43.7 & 52 & 0.24 & 27.8 & 58.0 & 73 \\
    & $\text{* "} \cdot  \text{"}$ & 0.39 & 27.1 & 46.7 & 72 & 0.25 & 22.9 & 47.0 & 60 \\
    & $\rangle \text{ "} \cdot \text{"}$ & 0.39 & 27.1 & 46.9 & 66 & 0.23 & 26.5 & 53.7 & 70 \\
    & $\{ \cdot \}$ & 0.39 & 28.7 & 48.8 & 77 & 0.24 & 28.0 & 54.8 & 63 \\
    & $\text{--} \cdot  \text{--}$ & 0.50 & 8.8 & 15.4 & 22 & 0.22 & 6.5 & 13.3 & 18 \\
    & $\{ \{ \cdot \} \}$ & 0.43 & 27.7 & 46.5 & 63 & 0.25 & 36.6 & 69.3 & 113 \\
    & $( \cdot )$ & 0.41 & 22.2 & 38.9 & 48 & 0.26 & 22.9 & 45.1 & 59 \\
    & $\text{"} \cdot \text{"}$ & 0.52 & 19.7 & 31.4 & 57 & 0.30 & 17.7 & 34.2 & 48 \\
    & $[ \cdot ]$ & 0.44 & 22.0 & 36.8 & 53 & 0.26 & 19.4 & 38.7 & 44 \\
    & $\langle\langle\langle \cdot \rangle\rangle\rangle$ & 0.47 & 13.1 & 21.9 & 28 & 0.28 & 12.2 & 24.6 & 25 \\
\midrule 
    \multirowcell{10}{\bf{GPT-2-XL} \\ (1558M)} & $\langle \cdot \rangle$ & 0.40 & 26.3 & 43.0 & 81 & 0.31 & 25.9 & 48.2 & 82 \\
    & $\text{* "} \cdot  \text{"}$ & 0.42 & 22.7 & 39.3 & 60 & 0.29 & 17.6 & 33.3 & 44 \\
    & $\rangle \text{ "} \cdot \text{"}$ & 0.43 & 20.8 & 35.1 & 54 & 0.29 & 17.9 & 34.0 & 47 \\
    & $\{ \cdot \}$ & 0.47 & 23.8 & 37.6 & 73 & 0.31 & 22.7 & 42.5 & 80 \\
    & $\text{--} \cdot  \text{--}$ & 0.56 & 5.6 & 9.0 & 19 & 0.28 & 4.3 & 7.8 & 18 \\
    & $\{ \{ \cdot \} \}$ & 0.42 & 29.8 & 49.8 & 99 & 0.29 & 25.7 & 46.4 & 80 \\
    & $( \cdot )$ & 0.45 & 16.4 & 28.8 & 41 & 0.28 & 17.8 & 33.6 & 53 \\
    & $\text{"} \cdot \text{"}$ & 0.47 & 16.1 & 26.2 & 38 & 0.30 & 14.6 & 28.4 & 41 \\
    & $[ \cdot ]$ & 0.46 & 19.2 & 30.8 & 60 & 0.32 & 16.4 & 31.4 & 52 \\
    & $\langle\langle\langle \cdot \rangle\rangle\rangle$ & 0.51 & 9.0 & 13.9 & 25 & 0.37 & 7.4 & 12.7 & 26 \\
\midrule 
    \multirowcell{10}{\bf{GPT-Neo-1.3B} \\ (1.3B)} & $\langle \cdot \rangle$ & 0.48 & 14.9 & 26.1 & 48 & 0.29 & 11.0 & 21.5 & 40 \\
    & $\text{* "} \cdot  \text{"}$ & 0.41 & 15.1 & 26.8 & 36 & 0.25 & 13.8 & 28.5 & 44 \\
    & $\rangle \text{ "} \cdot \text{"}$ & 0.38 & 19.7 & 36.0 & 60 & 0.26 & 18.9 & 37.4 & 48 \\
    & $\{ \cdot \}$ & 0.48 & 11.0 & 18.7 & 32 & 0.30 & 8.8 & 16.6 & 31 \\
    & $\text{--} \cdot  \text{--}$ & 0.54 & 5.0 & 8.6 & 18 & 0.30 & 4.9 & 9.6 & 18 \\
    & $\{ \{ \cdot \} \}$ & 0.49 & 15.3 & 25.0 & 47 & 0.31 & 13.6 & 24.6 & 42 \\
    & $( \cdot )$ & 0.44 & 14.3 & 25.3 & 44 & 0.27 & 15.5 & 29.1 & 51 \\
    & $\text{"} \cdot \text{"}$ & 0.39 & 19.5 & 33.4 & 50 & 0.25 & 15.8 & 31.4 & 37 \\
    & $[ \cdot ]$ & 0.46 & 15.1 & 27.0 & 51.7 & 0.26 & 15.1 & 30.4 & 47 \\
    & $\langle\langle\langle \cdot \rangle\rangle\rangle$ & 0.56 & 5.9 & 9.2 & 28 & 0.32 & 4.6 & 8.1 & 22 \\
\midrule 
    \multirowcell{10}{\bf{GPT-Neo-2.7B} \\ (2.7B)} & $\langle \cdot \rangle$ & 0.43 & 20.0 & 36.1 & 53 & 0.27 & 22.2 & 43.8 & 59 \\
    & $\text{* "} \cdot  \text{"}$ & 0.37 & 26.2 & 46.7 & 65 & 0.21 & 26.8 & 54.1 & 65 \\
    & $\rangle \text{ "} \cdot \text{"}$ & 0.37 & 26.1 & 45.9 & 68 & 0.22 & 23.6 & 50.7 & 60 \\
    & $\{ \cdot \}$ & 0.44 & 21.6 & 37.7 & 61 & 0.29 & 27.1 & 51.4 & 80 \\
    & $\text{--} \cdot  \text{--}$ & 0.56 & 4.4 & 7.7 & 15 & 0.23 & 4.0 & 8.2 & 14 \\
    & $\{ \{ \cdot \} \}$ & 0.42 & 23.7 & 42.0 & 56 & 0.24 & 29.5 & 58.5 & 72 \\
    & $( \cdot )$ & 0.44 & 19.7 & 32.9 & 48 & 0.27 & 21.1 & 40.9 & 64 \\
    & $\text{"} \cdot \text{"}$ & 0.38 & 26.1 & 44.5 & 67 & 0.22 & 26.3 & 52.7 & 67 \\
    & $[ \cdot ]$ & 0.48 & 20.3 & 35.6 & 67 & 0.25 & 22.4 & 42.9 & 58 \\
    & $\langle\langle\langle \cdot \rangle\rangle\rangle$ & 0.45 & 14.1 & 24.8 & 32 & 0.22 & 21.0 & 42.1 & 55 \\
\midrule 
    \multirowcell{10}{\bf{GPT-J-6B} \\ (6B)} & $\langle \cdot \rangle$ & 0.40 & 27.3 & 47.0 & 74 & 0.32 & 17.0 & 32.7 & 51 \\
    & $\text{* "} \cdot  \text{"}$ & 0.38 & 29.2 & 49.5 & 82 & 0.28 & 23.4 & 42.9 & 61 \\
    & $\rangle \text{ "} \cdot \text{"}$ & 0.36 & 27.4 & 47.2 & 69 & 0.30 & 23.1 & 43.6 & 64 \\
    & $\{ \cdot \}$ & 0.41 & 27.8 & 47.6 & 80 & 0.32 & 24.9 & 45.6 & 78 \\
    & $\text{--} \cdot  \text{--}$ & 0.43 & 7.1 & 12.3 & 19 & 0.20 & 4.8 & 9.0 & 17 \\
    & $\{ \{ \cdot \} \}$ & 0.29 & 30.4 & 54.9 & 72 & 0.29 & 26.8 & 51.1 & 76 \\
    & $( \cdot )$ & 0.48 & 24.9 & 41.6 & 80 & 0.35 & 22.3 & 39.3 & 77 \\
    & $\text{"} \cdot \text{"}$ & 0.39 & 28.6 & 47.3 & 69 & 0.31 & 23.0 & 42.4 & 64 \\
    & $[ \cdot ]$ & 0.43 & 23.3 & 38.2 & 63 & 0.37 & 20.8 & 38.3 & 60 \\
    & $\langle\langle\langle \cdot \rangle\rangle\rangle$ & 0.34 & 30.3 & 55.6 & 98 & 0.31 & 23.6 & 44.2 & 67 \\
\bottomrule
\end{tabular}
}\vspace{-0.6em}
\caption{Zero-shot performances of the off-the-shelf ``small'' language models from the GPT-2 and GPT-Neo/J families on the original version of the \amazon dataset. Here, we also experimented with ten different delimiter-pairs, ranging from curly brackets to asterisk quotes: Overall, curly brackets $\{ \cdot \}$, square brackets $[ \cdot ]$, parentheses $( \cdot )$, and quotes $\text{"} \cdot \text{"}$ yielded consistently reliable and high-quality outputs. Most of the models could not go beyond 60\% accuracy in the positive to negative direction and 35\% accuracy in the negative to positive direction. As shown in Table~\ref{tab:AmazonCleanZeroShot}, most models performed marginally better (in terms of their accuracy, BLEU, and PPL scores) on the cleaner version of the dataset, suggesting that the original version might contain some tokenization-related (semantic) noises that might be preventing the models from performing well.}
\label{tab:AmazonOriginalZeroShot}
\end{table*}

\newpage

\begin{table*}[!h]
\small 
\centering
\scalebox{0.85}{
\begin{tabular}{c | >{\rowmac}c | >{\rowmac}c  >{\rowmac}c  >{\rowmac}c  >{\rowmac}c | >{\rowmac}c  >{\rowmac}c  >{\rowmac}c  >{\rowmac}c <{\clearrow}}
\toprule
&  & \multicolumn{4}{c|}{ \bf{Positive $\to$ Negative} } & \multicolumn{4}{c}{ \bf{Negative $\to$ Positive} } \\
\bf{Model} &  \bf{Delimiter-Pair} & \bf{Acc} & \bf{\emph{r}-sBLEU} & \bf{\emph{s}-sBLEU} & \bf{PPL} & \bf{Acc} & \bf{\emph{r}-sBLEU} & \bf{\emph{s}-sBLEU} & \bf{PPL} \\ 
\midrule 
    \multirowcell{10}{\bf{GPT-2-Small} \\ (117M)} & $\langle \cdot \rangle$ & 0.34 & 17.9 & 33.4 & 47 & 0.19 & 13.7 & 30.5 & 42 \\
    & $\text{* "} \cdot  \text{"}$ & 0.43 & 8.1 & 14.8 & 38 & 0.26 & 7.9 & 16.3 & 37 \\
    & $\rangle \text{ "} \cdot \text{"}$ & 0.45 & 6.3 & 11.6 & 29 & 0.25 & 7.7 & 15.4 & 32 \\
    & $\{ \cdot \}$ & 0.31 & 18.9 & 34.9 & 49 & 0.18 & 17.7 & 38.1 & 48 \\
    & $\text{--} \cdot  \text{--}$ & 0.42 & 6.6 & 12.0 & 24 & 0.21 & 6.6 & 13.7 & 26 \\
    & $\{ \{ \cdot \} \}$ & 0.28 & 30.4 & 56.7 & 90 & 0.19 & 28.4 & 60.9 & 81 \\
    & $( \cdot )$ & 0.34 & 22.2 & 39.1 & 57 & 0.21 & 18.7 & 40.1 & 48 \\
    & $\text{"} \cdot \text{"}$ & 0.46 & 8.7 & 15.9 & 35 & 0.27 & 7.3 & 14.9 & 34 \\
    & $[ \cdot ]$ & 0.32 & 19.9 & 35.4 & 59 & 0.20 & 18.4 & 39.7 & 55 \\
    & $\langle\langle\langle \cdot \rangle\rangle\rangle$ & 0.39 & 9.3 & 17.0 & 31 & 0.20 & 9.0 & 18.8 & 28 \\
\midrule 
    \multirowcell{10}{\bf{GPT-2-Medium} \\ (345M)} & $\langle \cdot \rangle$ & 0.43 & 19.3 & 33.7 & 49 & 0.28 & 16.8 & 33.4 & 47 \\
  & $\text{* "} \cdot  \text{"}$ & 0.52 & 10.4 & 17.2 & 31 & 0.33 & 7.6 & 15.1 & 29 \\
  & $\rangle \text{ "} \cdot \text{"}$ & 0.46 & 10.4 & 17.6 & 30 & 0.32 & 6.3 & 12.9 & 25 \\
  & $\{ \cdot \}$ & 0.48 & 21.9 & 36.7 & 68 & 0.32 & 20.1 & 38.0 & 57 \\
  & $\text{--} \cdot  \text{--}$ & 0.57 & 3.9 & 6.8 & 20 & 0.29 & 3.1 & 5.9 & 18 \\
  & $\{ \{ \cdot \} \}$ & 0.45 & 23.3 & 40.1 & 62 & 0.31 & 21.2 & 41.1 & 64 \\
  & $( \cdot )$ & 0.45 & 19.9 & 33.1 & 50 & 0.32 & 17.6 & 35.3 & 58 \\
  & $\text{"} \cdot \text{"}$ & 0.49 & 9.8 & 16.1 & 31 & 0.35 & 7.0 & 12.8 & 25 \\
  & $[ \cdot ]$ & 0.50 & 15.7 & 25.9 & 42 & 0.30 & 13.2 & 26.1 & 46 \\
  & $\langle\langle\langle \cdot \rangle\rangle\rangle$ & 0.48 & 8.8 & 14.8 & 26 & 0.31 & 10.1 & 19.2 & 30 \\
\midrule 
    \multirowcell{10}{\bf{GPT-2-Large} \\ (774M)} & $\langle \cdot \rangle$ & 0.40 & 26.3 & 44.6 & 68 & 0.29 & 22.9 & 43.4 & 52 \\
    & $\text{* "} \cdot  \text{"}$ & 0.44 & 23.5 & 40.9 & 54 & 0.27 & 18.8 & 36.9 & 42 \\
    & $\rangle \text{ "} \cdot \text{"}$ & 0.43 & 20.8 & 36.0 & 44 & 0.29 & 19.4 & 39.2 & 42 \\
    & $\{ \cdot \}$ & 0.44 & 28.6 & 49.1 & 70 & 0.28 & 26.0 & 51.2 & 55 \\
    & $\text{--} \cdot  \text{--}$ & 0.40 & 7.3 & 12.0 & 17 & 0.36 & 7.4 & 13.8 & 19 \\
    & $\{ \{ \cdot \} \}$ & 0.43 & 31.0 & 53.1 & 90 & 0.29 & 31.3 & 60.0 & 79 \\
    & $( \cdot )$ & 0.39 & 23.0 & 39.1 & 51 & 0.26 & 23.3 & 45.3 & 61 \\
    & $\text{"} \cdot \text{"}$ & 0.47 & 18.1 & 29.8 & 47 & 0.31 & 16.8 & 32.7 & 48 \\
    & $[ \cdot ]$ & 0.47 & 20.2 & 34.3 & 50 & 0.26 & 17.5 & 34.1 & 42 \\
    & $\langle\langle\langle \cdot \rangle\rangle\rangle$ & 0.49 & 9.9 & 16.2 & 21 & 0.27 & 9.4 & 18.2 & 22 \\
\midrule 
    \multirowcell{10}{\bf{GPT-2-XL} \\ (1558M)} & $\langle \cdot \rangle$ & 0.40 & 25.6 & 42.0 & 68 & 0.29 & 23.1 & 43.0 & 65 \\
    & $\text{* "} \cdot  \text{"}$ & 0.36 & 22.5 & 39.4 & 48 & 0.31 & 18.7 & 37.5 & 47 \\
    & $\rangle \text{ "} \cdot \text{"}$ & 0.40 & 18.8 & 31.5 & 43 & 0.27 & 19.2 & 37.8 & 46 \\
    & $\{ \cdot \}$ & 0.46 & 21.5 & 35.4 & 59 & 0.32 & 22.3 & 41.4 & 70 \\
    & $\text{--} \cdot  \text{--}$ & 0.53 & 7.2 & 11.7 & 23 & 0.32 & 6.9 & 11.8 & 21 \\
    & $\{ \{ \cdot \} \}$ & 0.45 & 25.7 & 43.2 & 81 & 0.31 & 24.8 & 45.4 & 72 \\
    & $( \cdot )$ & 0.48 & 19.3 & 30.7 & 52 & 0.30 & 17.8 & 33.4 & 53 \\
    & $\text{"} \cdot \text{"}$ & 0.45 & 20.5 & 33.6 & 49 & 0.31 & 17.9 & 33.4 & 51 \\
    & $[ \cdot ]$ & 0.47 & 21.1 & 33.4 & 55 & 0.32 & 19.2 & 34.7 & 55 \\
    & $\langle\langle\langle \cdot \rangle\rangle\rangle$ & 0.47 & 7.8 & 13.1 & 24 & 0.38 & 6.8 & 12.5 & 25 \\
\midrule 
    \multirowcell{10}{\bf{GPT-Neo-1.3B} \\ (1.3B)} & $\langle \cdot \rangle$ & 0.50 & 11.4 & 20.1 & 38 & 0.30 & 10.6 & 19.7 & 34 \\
    & $\text{* "} \cdot  \text{"}$ & 0.38 & 15.0 & 25.0 & 37 & 0.26 & 11.9 & 22.4 & 31 \\
    & $\rangle \text{ "} \cdot \text{"}$ & 0.40 & 12.6 & 22.1 & 35 & 0.26 & 11.3 & 22.2 & 29 \\
    & $\{ \cdot \}$ & 0.49 & 11.8 & 19.9 & 34 & 0.31 & 10.9 & 20.5 & 35 \\
    & $\text{--} \cdot  \text{--}$ & 0.50 & 4.1 & 6.8 & 18 & 0.25 & 4.4 & 8.5 & 18 \\
    & $\{ \{ \cdot \} \}$ & 0.48 & 13.9 & 23.9 & 41 & 0.35 & 12.4 & 22.9 & 38 \\
    & $( \cdot )$ & 0.42 & 16.6 & 27.8 & 53 & 0.28 & 13.1 & 25.8 & 42 \\
    & $\text{"} \cdot \text{"}$ & 0.45 & 13.8 & 24.8 & 36 & 0.30 & 12.4 & 24.7 & 32 \\
    & $[ \cdot ]$ & 0.46 & 16.7 & 28.1 & 45 & 0.26 & 14.7 & 28.5 & 43 \\
    & $\langle\langle\langle \cdot \rangle\rangle\rangle$ & 0.57 & 3.4 & 5.8 & 20 & 0.36 & 3.1 & 5.5 & 18 \\
\midrule 
    \multirowcell{10}{\bf{GPT-Neo-2.7B} \\ (2.7B)} & $\langle \cdot \rangle$ & 0.44 & 20.1 & 34.8 & 51 & 0.29 & 19.5 & 38.3 & 46 \\
    & $\text{* "} \cdot  \text{"}$ & 0.40 & 27.2 & 47.9 & 61 & 0.22 & 28.9 & 57.6 & 58 \\
    & $\rangle \text{ "} \cdot \text{"}$ & 0.37 & 21.8 & 39.4 & 45 & 0.21 & 22.5 & 45.6 & 41 \\
    & $\{ \cdot \}$ & 0.48 & 21.4 & 36.5 & 55 & 0.28 & 23.7 & 45.9 & 57 \\
    & $\text{--} \cdot  \text{--}$ & 0.56 & 3.9 & 6.7 & 14 & 0.26 & 3.8 & 7.4 & 13 \\
    & $\{ \{ \cdot \} \}$ & 0.43 & 21.2 & 36.2 & 44 & 0.27 & 28.0 & 55.7 & 56 \\
    & $( \cdot )$ & 0.48 & 17.0 & 28.7 & 42 & 0.32 & 19.5 & 36.8 & 52 \\
    & $\text{"} \cdot \text{"}$ & 0.38 & 25.6 & 44.5 & 58 & 0.22 & 28.6 & 58.2 & 59 \\
    & $[ \cdot ]$ & 0.48 & 18.7 & 32.1 & 47 & 0.26 & 23.3 & 46.2 & 50 \\
    & $\langle\langle\langle \cdot \rangle\rangle\rangle$ & 0.49 & 14.8 & 24.9 & 33 & 0.32 & 18.7 & 37.3 & 37 \\
\midrule 
    \multirowcell{10}{\bf{GPT-J-6B} \\ (6B)} & $\langle \cdot \rangle$ & 0.40 & 25.0 & 43.4 & 66 & 0.35 & 20.1 & 35.7 & 56 \\
    & $\text{* "} \cdot  \text{"}$ & 0.42 & 26.8 & 44.8 & 60 & 0.33 & 23.5 & 41.9 & 56 \\
    & $\rangle \text{ "} \cdot \text{"}$ & 0.39 & 29.3 & 50.3 & 65 & 0.31 & 24.2 & 44.8 & 64 \\
    & $\{ \cdot \}$ & 0.41 & 26.1 & 44.6 & 57 & 0.33 & 27.1 & 47.7 & 72 \\
    & $\text{--} \cdot  \text{--}$ & 0.46 & 5.9 & 9.7 & 17 & 0.23 & 4.6 & 8.6 & 17 \\
    & $\{ \{ \cdot \} \}$ & 0.30 & 29.9 & 53.8 & 56 & 0.30 & 27.5 & 52.3 & 73 \\
    & $( \cdot )$ & 0.44 & 21.0 & 34.8 & 53 & 0.37 & 19.5 & 37.7 & 64 \\
    & $\text{"} \cdot \text{"}$ & 0.41 & 27.8 & 45.7 & 63 & 0.34 & 25.3 & 45.2 & 62 \\
    & $[ \cdot ]$ & 0.45 & 20.1 & 34.0 & 53 & 0.36 & 20.8 & 37.7 & 65 \\
    & $\langle\langle\langle \cdot \rangle\rangle\rangle$ & 0.39 & 28.3 & 47.4 & 66 & 0.32 & 25.6 & 47.2 & 66 \\
\bottomrule
\end{tabular}
}
\caption{Zero-shot performances of the off-the-shelf ``small'' language models on the clean version of the \amazon dataset (\amazon-clean, in short). As before, none of the models could go beyond the $60\%$ accuracy level, but most of them seem to have achieved slightly better perplexity scores in the clean version of the dataset than in the original version.}
\label{tab:AmazonCleanZeroShot}
\end{table*}

\newpage

\begin{table*}[!h]
\small 
\centering
\scalebox{0.85}{
\begin{tabular}{c | >{\rowmac}c | >{\rowmac}c  >{\rowmac}c  >{\rowmac}c  >{\rowmac}c | >{\rowmac}c  >{\rowmac}c  >{\rowmac}c  >{\rowmac}c <{\clearrow}}
\toprule
&  & \multicolumn{4}{c|}{ \bf{Positive $\to$ Negative} } & \multicolumn{4}{c}{ \bf{Negative $\to$ Positive} } \\
\bf{Model} &  \bf{Delimiter-Pair} & \bf{Acc} & \bf{\emph{r}-sBLEU} & \bf{\emph{s}-sBLEU} & \bf{PPL} & \bf{Acc} & \bf{\emph{r}-sBLEU} & \bf{\emph{s}-sBLEU} & \bf{PPL} \\ 
\midrule
    \multirowcell{10}{\bf{GPT-2-Small} \\ (117M)} & $\langle \cdot \rangle$ & 0.38 & 10.1 & 29.7 & 49 & 0.11 & 13.8 & 40.6 & 47 \\
    & $\text{* "} \cdot  \text{"}$ & 0.41 & 3.1 & 10.6 & 31 & 0.21 & 4.4 & 14.0 & 33 \\
    & $\rangle \text{ "} \cdot \text{"}$ & 0.33 & 4.8 & 13.4 & 31 & 0.15 & 6.5 & 18.6 & 31 \\
    & $\{ \cdot \}$ & 0.36 & 8.4 & 23.3 & 35 & 0.15 & 7.8 & 23.5 & 34 \\
    & $\text{--} \cdot  \text{--}$ & 0.45 & 4.5 & 12.0 & 28 & 0.14 & 6.7 & 19.3 & 33 \\
    & $\{ \{ \cdot \} \}$ & 0.37 & 13.4 & 38.9 & 62 & 0.11 & 16.6 & 49.1 & 54 \\
    & $( \cdot )$ & 0.36 & 8.0 & 25.2 & 42 & 0.13 & 10.9 & 32.7 & 46 \\
    & $\text{"} \cdot \text{"}$ & 0.37 & 4.9 & 13.7 & 30 & 0.18 & 6.7 & 20.3 & 35 \\
    & $[ \cdot ]$ & 0.36 & 6.8 & 20.7 & 38 & 0.12 & 8.7 & 25.2 & 31 \\
    & $\langle\langle\langle \cdot \rangle\rangle\rangle$ & 0.43 & 2.1 & 5.8 & 17 & 0.09 & 2.1 & 6.4 & 17 \\
\midrule 
    \multirowcell{10}{\bf{GPT-2-Medium} \\ (345M)} & $\langle \cdot \rangle$ & 0.55 & 9.9 & 27.0 & 44 & 0.31 & 10.3 & 28.2 & 38 \\
    & $\text{* "} \cdot  \text{"}$ & 0.64 & 7.4 & 17.5 & 40 & 0.38 & 7.2 & 16.7 & 33 \\
    & $\rangle \text{ "} \cdot \text{"}$ & 0.52 & 5.5 & 15.3 & 26 & 0.31 & 5.8 & 14.8 & 26 \\
    & $\{ \cdot \}$ & 0.66 & 7.6 & 17.6 & 34 & 0.35 & 8.7 & 24.1 & 34 \\
    & $\text{--} \cdot  \text{--}$ & 0.69 & 4.6 & 11.3 & 30 & 0.36 & 5.0 & 12.3 & 27 \\
    & $\{ \{ \cdot \} \}$ & 0.68 & 12.2 & 30.3 & 52 & 0.32 & 14.8 & 36.6 & 58 \\
    & $( \cdot )$ & 0.63 & 8.5 & 22.1 & 46 & 0.32 & 10.5 & 26.8 & 48 \\
    & $\text{"} \cdot \text{"}$ & 0.66 & 5.1 & 13.1 & 29 & 0.41 & 6.3 & 15.2 & 30 \\
    & $[ \cdot ]$ & 0.66 & 8.4 & 22.0 & 36 & 0.32 & 8.1 & 21.0 & 33 \\
    & $\langle\langle\langle \cdot \rangle\rangle\rangle$ & 0.64 & 1.8 & 4.7 & 16 & 0.24 & 2.2 & 5.9 & 16 \\
\midrule 
    \multirowcell{10}{\bf{GPT-2-Large} \\ (774M)} & $\langle \cdot \rangle$ & 0.65 & 14.5 & 36.8 & 54 & 0.22 & 17.3 & 46.7 & 38 \\
    & $\text{* "} \cdot  \text{"}$ & 0.61 & 13.8 & 33.7 & 43 & 0.27 & 15.8 & 44.4 & 45 \\
    & $\rangle \text{ "} \cdot \text{"}$ & 0.57 & 16.2 & 44.0 & 59 & 0.27 & 18.7 & 51.4 & 53 \\
    & $\{ \cdot \}$ & 0.68 & 12.8 & 30.6 & 41 & 0.26 & 13.5 & 37.5 & 38 \\
    & $\text{--} \cdot  \text{--}$ & 0.64 & 8.9 & 22.3 & 31 & 0.24 & 9.7 & 26.3 & 25 \\
    & $\{ \{ \cdot \} \}$ & 0.69 & 18.2 & 45.9 & 75 & 0.24 & 20.6 & 58.2 & 55 \\
    & $( \cdot )$ & 0.68 & 10.6 & 26.0 & 40 & 0.28 & 15.4 & 40.6 & 46 \\
    & $\text{"} \cdot \text{"}$ & 0.74 & 12.0 & 25.9 & 44 & 0.34 & 14.3 & 34.7 & 42 \\
    & $[ \cdot ]$ & 0.70 & 8.3 & 20.2 & 31 & 0.28 & 9.7 & 26.2 & 32 \\
    & $\langle\langle\langle \cdot \rangle\rangle\rangle$ & 0.73 & 6.1 & 14.8 & 21 & 0.27 & 6.9 & 17.7 & 19 \\
\midrule 
    \multirowcell{10}{\bf{GPT-2-XL} \\ (1558M)} & $\langle \cdot \rangle$ & 0.67 & 15.0 & 35.3 & 59 & 0.41 & 16.4 & 40.4 & 54 \\
    & $\text{* "} \cdot  \text{"}$ & 0.67 & 10.0 & 25.0 & 35 & 0.37 & 13.0 & 31.6 & 36 \\
    & $\rangle \text{ "} \cdot \text{"}$ & 0.66 & 10.4 & 25.8 & 41 & 0.34 & 12.5 & 30.2 & 39 \\
    & $\{ \cdot \}$ & 0.78 & 9.7 & 21.1 & 41 & 0.41 & 12.1 & 30.1 & 40 \\
    & $\text{--} \cdot  \text{--}$ & 0.74 & 6.4 & 13.9 & 25 & 0.37 & 6.3 & 14.2 & 20 \\
    & $\{ \{ \cdot \} \}$ & 0.67 & 17.2 & 38.9 & 61 & 0.35 & 18.8 & 49.2 & 66 \\
    & $( \cdot )$ & 0.72 & 8.6 & 18.5 & 35 & 0.40 & 12.4 & 28.3 & 42 \\
    & $\text{"} \cdot \text{"}$ & 0.72 & 9.7 & 23.3 & 41 & 0.38 & 10.3 & 24.9 & 34 \\
    & $[ \cdot ]$ & 0.72 & 9.2 & 22.0 & 35 & 0.41 & 10.1 & 23.5 & 31 \\
    & $\langle\langle\langle \cdot \rangle\rangle\rangle$ & 0.70 & 4.0 & 9.5 & 18 & 0.39 & 4.6 & 11.0 & 17 \\
\midrule 
    \multirowcell{10}{\bf{GPT-Neo-1.3B} \\ (1.3B)} & $\langle \cdot \rangle$ & 0.61 & 6.5 & 16.0 & 28 & 0.38 & 6.8 & 15.7 & 26 \\
    & $\text{* "} \cdot  \text{"}$ & 0.31 & 13.3 & 38.7 & 33 & 0.24 & 13.5 & 35.4 & 37 \\
    & $\rangle \text{ "} \cdot \text{"}$ & 0.24 & 16.9 & 52.1 & 54 & 0.21 & 15.7 & 45.8 & 44 \\
    & $\{ \cdot \}$ & 0.66 & 3.2 & 8.4 & 19 & 0.38 & 5.3 & 12.2 & 21 \\
    & $\text{--} \cdot  \text{--}$ & 0.52 & 2.9 & 8.4 & 17 & 0.30 & 4.4 & 11.2 & 20 \\
    & $\{ \{ \cdot \} \}$ & 0.60 & 9.1 & 23.6 & 35 & 0.39 & 8.5 & 21.2 & 30 \\
    & $( \cdot )$ & 0.59 & 6.8 & 18.6 & 34 & 0.27 & 11.1 & 31.1 & 47 \\
    & $\text{"} \cdot \text{"}$ & 0.46 & 14.9 & 40.2 & 54 & 0.23 & 14.9 & 40.1 & 47 \\
    & $[ \cdot ]$ & 0.57 & 8.1 & 20.8 & 38 & 0.36 & 8.4 & 22.1 & 33 \\
    & $\langle\langle\langle \cdot \rangle\rangle\rangle$ & 0.68 & 1.3 & 3.8 & 17 & 0.38 & 1.9 & 4.1 & 16 \\
\midrule 
    \multirowcell{10}{\bf{GPT-Neo-2.7B} \\ (2.7B)} & $\langle \cdot \rangle$ & 0.68 & 8.3 & 21.8 & 28 & 0.28 & 12.5 & 33.1 & 31 \\
    & $\text{* "} \cdot  \text{"}$ & 0.56 & 14.1 & 39.1 & 54 & 0.18 & 18.1 & 51.9 & 51 \\
    & $\rangle \text{ "} \cdot \text{"}$ & 0.54 & 12.1 & 34.3 & 43 & 0.21 & 15.9 & 46.4 & 42 \\
    & $\{ \cdot \}$ & 0.63 & 7.5 & 19.5 & 27 & 0.26 & 12.4 & 32.8 & 32 \\
    & $\text{--} \cdot  \text{--}$ & 0.63 & 3.4 & 8.5 & 16 & 0.26 & 3.7 & 9.8 & 14 \\
    & $\{ \{ \cdot \} \}$ & 0.55 & 12.0 & 32.9 & 40 & 0.20 & 16.2 & 48.1 & 42 \\
    & $( \cdot )$ & 0.73 & 7.3 & 18.0 & 38 & 0.34 & 13.6 & 34.2 & 47 \\
    & $\text{"} \cdot \text{"}$ & 0.55 & 12.6 & 33.2 & 42 & 0.17 & 15.4 & 45.9 & 42 \\
    & $[ \cdot ]$ & 0.62 & 8.0 & 21.3 & 33 & 0.27 & 13.7 & 37.0 & 39 \\
    & $\langle\langle\langle \cdot \rangle\rangle\rangle$ & 0.64 & 4.6 & 12.6 & 21 & 0.27 & 7.1 & 18.5 & 20 \\
\midrule 
    \multirowcell{10}{\bf{GPT-J-6B} \\ (6B)} & $\langle \cdot \rangle$ & 0.60 & 14.9 & 37.6 & 52 & 0.44 & 14.4 & 33.0 & 51 \\
    & $\text{* "} \cdot  \text{"}$ & 0.57 & 16.2 & 41.0 & 57 & 0.36 & 16.1 & 37.1 & 49 \\
    & $\rangle \text{ "} \cdot \text{"}$ & 0.52 & 16.4 & 43.3 & 62 & 0.32 & 17.2 & 42.6 & 56 \\
    & $\{ \cdot \}$ & 0.60 & 13.6 & 36.1 & 51 & 0.46 & 14.3 & 32.0 & 46 \\
    & $\text{--} \cdot  \text{--}$ & 0.60 & 4.4 & 11.4 & 20 & 0.27 & 3.7 & 8.9 & 17 \\
    & $\{ \{ \cdot \} \}$ & 0.44 & 16.2 & 44.4 & 50 & 0.34 & 17.8 & 43.0 & 56 \\
    & $( \cdot )$ & 0.64 & 11.3 & 29.2 & 46 & 0.50 & 15.8 & 34.1 & 57 \\
    & $\text{"} \cdot \text{"}$ & 0.57 & 12.7 & 34.0 & 52 & 0.37 & 14.8 & 34.5 & 43 \\
    & $[ \cdot ]$ & 0.58 & 13.6 & 35.3 & 56 & 0.51 & 13.1 & 29.1 & 44 \\
    & $\langle\langle\langle \cdot \rangle\rangle\rangle$ & 0.47 & 16.1 & 45.3 & 58 & 0.40 & 13.5 & 31.7 & 43 \\
\bottomrule
\end{tabular}
}
\caption{Zero-shot performances of the off-the-shelf ``small'' language models on the original version of the \yelp dataset. Amongst all the model architectures, GPT-J-6B had the finest results, both quantitatively and qualitatively. We also note the performance differences between the positive to negative direction and the negative to positive direction across all the experiments. It appears that the former direction is easier for all the models than the latter direction. Furthermore, as in the case of \amazon, Table~\ref{tab:YelpCleanZeroShot} illustrates that most models performed slightly better in the the clean version of \yelp than in the original version.} 
\label{tab:YelpOriginalZeroShot}
\end{table*}

\newpage

\begin{table*}[!h]
\small 
\centering
\scalebox{0.85}{
\begin{tabular}{c | >{\rowmac}c | >{\rowmac}c  >{\rowmac}c  >{\rowmac}c  >{\rowmac}c | >{\rowmac}c  >{\rowmac}c  >{\rowmac}c  >{\rowmac}c <{\clearrow}}
\toprule
&  & \multicolumn{4}{c|}{ \bf{Positive $\to$ Negative} } & \multicolumn{4}{c}{ \bf{Negative $\to$ Positive} } \\
\bf{Model} &  \bf{Delimiter-Pair} & \bf{Acc} & \bf{\emph{r}-sBLEU} & \bf{\emph{s}-sBLEU} & \bf{PPL} & \bf{Acc} & \bf{\emph{r}-sBLEU} & \bf{\emph{s}-sBLEU} & \bf{PPL} \\ 
\midrule
    \multirowcell{10}{\bf{GPT-2-Small} \\ (117M)} & $\langle \cdot \rangle$ & 0.34 & 13.1 & 37.9 & 52 & 0.14 & 12.9 & 38.2 & 50 \\
    & $\text{* "} \cdot  \text{"}$ & 0.38 & 3.2 & 9.2 & 30 & 0.23 & 3.7 & 11.3 & 29 \\
    & $\rangle \text{ "} \cdot \text{"}$ & 0.38 & 3.1 & 10.4 & 25 & 0.16 & 5.3 & 14.9 & 27 \\
    & $\{ \cdot \}$ & 0.36 & 6.6 & 19.3 & 28 & 0.12 & 8.2 & 25.7 & 29 \\
    & $\text{--} \cdot  \text{--}$ & 0.43 & 4.3 & 12.4 & 24 & 0.12 & 5.3 & 16.7 & 27 \\
    & $\{ \{ \cdot \} \}$ & 0.37 & 16.0 & 45.0 & 67 & 0.12 & 17.2 & 54.0 & 59 \\
    & $( \cdot )$ & 0.42 & 8.2 & 22.5 & 34 & 0.13 & 12.1 & 37.9 & 37 \\
    & $\text{"} \cdot \text{"}$ & 0.35 & 4.9 & 15.4 & 33 & 0.21 & 6.1 & 18.5 & 30 \\
    & $[ \cdot ]$ & 0.42 & 8.3 & 23.0 & 41 & 0.13 & 11.3 & 35.9 & 39 \\
    & $\langle\langle\langle \cdot \rangle\rangle\rangle$ & 0.50 & 1.5 & 4.0 & 14 & 0.11 & 1.9 & 5.9 & 15 \\
\midrule 
    \multirowcell{10}{\bf{GPT-2-Medium} \\ (345M)} & $\langle \cdot \rangle$ & 0.57 & 8.9 & 23.7 & 37 & 0.31 & 9.4 & 27.3 & 39 \\
    & $\text{* "} \cdot  \text{"}$ & 0.64 & 5.8 & 14.5 & 31 & 0.41 & 5.4 & 14.7 & 29 \\
    & $\rangle \text{ "} \cdot \text{"}$ & 0.52 & 5.7 & 16.7 & 28 & 0.30 & 5.6 & 15.0 & 26 \\
    & $\{ \cdot \}$ & 0.65 & 7.3 & 19.9 & 37 & 0.33 & 10.5 & 28.6 & 41 \\
    & $\text{--} \cdot  \text{--}$ & 0.66 & 3.9 & 9.7 & 23 & 0.34 & 3.0 & 7.4 & 20 \\
    & $\{ \{ \cdot \} \}$ & 0.63 & 13.1 & 33.5 & 52 & 0.31 & 12.7 & 35.5 & 48 \\
    & $( \cdot )$ & 0.64 & 9.4 & 25.3 & 44 & 0.29 & 11.8 & 33.8 & 44 \\
    & $\text{"} \cdot \text{"}$ & 0.63 & 5.2 & 14.2 & 29 & 0.42 & 6.1 & 15.9 & 27 \\
    & $[ \cdot ]$ & 0.64 & 7.0 & 18.4 & 35 & 0.33 & 8.2 & 22.4 & 33 \\
    & $\langle\langle\langle \cdot \rangle\rangle\rangle$ & 0.62 & 1.9 & 5.0 & 15 & 0.24 & 1.7 & 4.8 & 14 \\
\midrule 
    \multirowcell{10}{\bf{GPT-2-Large} \\ (774M)} & $\langle \cdot \rangle$ & 0.63 & 14.3 & 36.3 & 46 & 0.27 & 17.9 & 48.2 & 44 \\
    & $\text{* "} \cdot  \text{"}$ & 0.65 & 13.5 & 33.3 & 47 & 0.35 & 12.5 & 34.6 & 36 \\
    & $\rangle \text{ "} \cdot \text{"}$ & 0.61 & 13.9 & 35.9 & 47 & 0.32 & 15.3 & 42.9 & 44 \\
    & $\{ \cdot \}$ & 0.67 & 12.0 & 28.8 & 40 & 0.30 & 12.5 & 33.8 & 30 \\
    & $\text{--} \cdot  \text{--}$ & 0.65 & 5.0 & 13.7 & 18 & 0.26 & 7.3 & 19.8 & 20 \\
    & $\{ \{ \cdot \} \}$ & 0.75 & 17.2 & 39.9 & 59 & 0.31 & 21.3 & 58.1 & 62 \\
    & $( \cdot )$ & 0.69 & 12.2 & 29.2 & 47 & 0.31 & 14.6 & 40.7 & 46 \\
    & $\text{"} \cdot \text{"}$ & 0.77 & 11.8 & 27.3 & 41 & 0.37 & 11.7 & 29.6 & 34 \\
    & $[ \cdot ]$ & 0.75 & 10.3 & 24.7 & 40 & 0.38 & 12.9 & 32.9 & 38 \\
    & $\langle\langle\langle \cdot \rangle\rangle\rangle$ & 0.72 & 3.6 & 9.1 & 16 & 0.31 & 4.2 & 10.7 & 15 \\
\midrule 
    \multirowcell{10}{\bf{GPT-2-XL} \\ (1558M)} & $\langle \cdot \rangle$ & 0.64 & 17.4 & 40.1 & 58 & 0.35 & 17.3 & 41.5 & 53 \\
    & $\text{* "} \cdot  \text{"}$ & 0.69 & 11.3 & 28.2 & 40 & 0.41 & 12.6 & 31.3 & 33 \\
    & $\rangle \text{ "} \cdot \text{"}$ & 0.71 & 9.7 & 22.1 & 36 & 0.35 & 11.6 & 28.6 & 34 \\
    & $\{ \cdot \}$ & 0.73 & 8.6 & 21.3 & 35 & 0.46 & 11.4 & 25.9 & 35 \\
    & $\text{--} \cdot  \text{--}$ & 0.70 & 6.0 & 15.4 & 23 & 0.39 & 6.8 & 17.5 & 25 \\
    & $\{ \{ \cdot \} \}$ & 0.63 & 17.4 & 40.9 & 70 & 0.38 & 19.1 & 46.8 & 59 \\
    & $( \cdot )$ & 0.72 & 10.8 & 25.0 & 45 & 0.39 & 14.0 & 31.1 & 41 \\
    & $\text{"} \cdot \text{"}$ & 0.77 & 7.6 & 17.6 & 31 & 0.44 & 9.9 & 23.1 & 30 \\
    & $[ \cdot ]$ & 0.75 & 10.8 & 24.9 & 38 & 0.41 & 12.0 & 29.9 & 43 \\
    & $\langle\langle\langle \cdot \rangle\rangle\rangle$ & 0.68 & 2.2 & 5.4 & 14 & 0.32 & 2.0 & 5.1 & 13 \\
\midrule 
    \multirowcell{10}{\bf{GPT-Neo-1.3B} \\ (1.3B)} & $\langle \cdot \rangle$ & 0.68 & 6.5 & 16.7 & 27 & 0.42 & 6.9 & 17.6 & 29 \\
    & $\text{* "} \cdot  \text{"}$ & 0.38 & 12.5 & 37.0 & 37 & 0.22 & 12.5 & 36.2 & 33 \\
    & $\rangle \text{ "} \cdot \text{"}$ & 0.32 & 13.7 & 42.3 & 41 & 0.19 & 16.1 & 47.9 & 40 \\
    & $\{ \cdot \}$ & 0.69 & 4.6 & 10.5 & 22 & 0.37 & 6.3 & 15.3 & 23 \\
    & $\text{--} \cdot  \text{--}$ & 0.58 & 3.1 & 8.1 & 18 & 0.33 & 4.2 & 11.1 & 17 \\
    & $\{ \{ \cdot \} \}$ & 0.69 & 7.2 & 17.0 & 30 & 0.40 & 8.9 & 20.6 & 27 \\
    & $( \cdot )$ & 0.63 & 8.6 & 21.3 & 39 & 0.28 & 8.3 & 23.3 & 28 \\
    & $\text{"} \cdot \text{"}$ & 0.47 & 12.6 & 35.2 & 43 & 0.30 & 13.8 & 35.6 & 36 \\
    & $[ \cdot ]$ & 0.68 & 8.3 & 21.4 & 40 & 0.34 & 8.6 & 23.9 & 30 \\
    & $\langle\langle\langle \cdot \rangle\rangle\rangle$ & 0.72 & 1.2 & 2.9 & 15 & 0.38 & 1.7 & 3.6 & 15 \\
\midrule 
    \multirowcell{10}{\bf{GPT-Neo-2.7B} \\ (2.7B)} & $\langle \cdot \rangle$ & 0.66 & 8.8 & 23.4 & 31 & 0.32 & 13.9 & 35.3 & 36 \\
    & $\text{* "} \cdot  \text{"}$ & 0.58 & 14.5 & 36.9 & 42 & 0.17 & 17.4 & 51.1 & 42 \\
    & $\rangle \text{ "} \cdot \text{"}$ & 0.54 & 13.8 & 38.3 & 43 & 0.21 & 13.2 & 39.9 & 32 \\
    & $\{ \cdot \}$ & 0.64 & 7.0 & 19.0 & 24 & 0.28 & 11.0 & 32.6 & 29 \\
    & $\text{--} \cdot  \text{--}$ & 0.68 & 3.4 & 9.1 & 17 & 0.26 & 5.0 & 14.3 & 16 \\
    & $\{ \{ \cdot \} \}$ & 0.57 & 11.0 & 29.3 & 31 & 0.24 & 15.5 & 46.1 & 35 \\
    & $( \cdot )$ & 0.76 & 10.3 & 23.1 & 44 & 0.41 & 14.4 & 33.6 & 43 \\
    & $\text{"} \cdot \text{"}$ & 0.59 & 12.5 & 33.6 & 42 & 0.21 & 14.9 & 43.4 & 38 \\
    & $[ \cdot ]$ & 0.66 & 9.6 & 23.7 & 32 & 0.29 & 14.9 & 42.4 & 43 \\
    & $\langle\langle\langle \cdot \rangle\rangle\rangle$ & 0.64 & 5.4 & 14.4 & 21 & 0.27 & 8.1 & 21.7 & 23 \\
\midrule 
    \multirowcell{10}{\bf{GPT-J-6B} \\ (6B)} & $\langle \cdot \rangle$ & 0.62 & 14.1 & 35.3 & 50 & 0.47 & 14.7 & 33.4 & 44 \\
    & $\text{* "} \cdot  \text{"}$ & 0.55 & 17.1 & 43.9 & 65 & 0.40 & 13.2 & 31.8 & 41 \\
    & $\rangle \text{ "} \cdot \text{"}$ & 0.61 & 16.9 & 41.6 & 56 & 0.38 & 13.3 & 30.5 & 37 \\
    & $\{ \cdot \}$ & 0.61 & 14.3 & 34.7 & 49 & 0.48 & 13.5 & 30.6 & 43 \\
    & $\text{--} \cdot  \text{--}$ & 0.54 & 5.4 & 14.7 & 22 & 0.36 & 4.8 & 10.9 & 19 \\
    & $\{ \{ \cdot \} \}$ & 0.42 & 14.7 & 42.3 & 38 & 0.33 & 17.5 & 45.0 & 53 \\
    & $( \cdot )$ & 0.66 & 12.9 & 30.7 & 50 & 0.51 & 11.5 & 23.4 & 44 \\
    & $\text{"} \cdot \text{"}$ & 0.66 & 15.7 & 36.1 & 55 & 0.40 & 16.6 & 36.2 & 45 \\
    & $[ \cdot ]$ & 0.69 & 11.8 & 28.7 & 45 & 0.53 & 13.3 & 27.3 & 43 \\
    & $\langle\langle\langle \cdot \rangle\rangle\rangle$ & 0.53 & 10.7 & 29.8 & 35 & 0.43 & 11.4 & 25.4 & 30 \\
\bottomrule
\end{tabular}
}
\caption{Zero-shot performances of the off-the-shelf ``small'' language models on the clean version of the \yelp dataset (\yelp-clean, in short). In contrast to Table~\ref{tab:YelpOriginalZeroShot}, we note that models, overall, achieved better results in \yelp-clean than in \yelp-original. Some models even could go beyond the $75\%$ accuracy level in the positive to negative direction. Consistent with the previous findings, these results also indicate that curly brackets $\{ \cdot \}$, square brackets $[ \cdot ]$, parentheses $( \cdot )$, and quotes $\text{"} \cdot \text{"}$ are favourable delimiter-pairs, leading to better outcomes than many other delimiter-pairs.}
\label{tab:YelpCleanZeroShot}
\end{table*}

\newpage

\begin{table*}[!h]
\small 
\centering
\scalebox{0.95}{
\begin{tabular}{c | >{\rowmac}c | >{\rowmac}c  >{\rowmac}c  >{\rowmac}c  >{\rowmac}c <{\clearrow}}
\toprule
&  & \multicolumn{4}{c}{ \bf{Shakespearean $\to$ Modern English} } \\
\bf{Model} &  \bf{Setting} & \bf{Acc} & \bf{\emph{r}-sBLEU} & \bf{\emph{s}-sBLEU} & \bf{PPL}  \\ 
\toprule
    {\bf{GPT-2-Small}} (117M) & 4-Shot (Top Choice) & 0.35 & 17.1 & 42.4 & 65 \\
    {\bf{GPT-2-Medium}} (345M) & 4-Shot (Top Choice) & 0.50 & 7.1 & 13.9 & 65 \\
    {\bf{GPT-2-Large}} (774M) & 4-Shot (Top Choice) & 0.38 & 14.1 & 30.9 & 134 \\
    {\bf{GPT-2-XL}} (1558M) & 4-Shot (Top Choice) & 0.39 & 18.9 & 38.4 & 90 \\
    {\bf{GPT-Neo-1.3B}} (1.3B) & 4-Shot (Top Choice) & 0.39 & 17.2 & 37.0 & 63 \\
    {\bf{GPT-Neo-2.7B}} (2.7B) & 4-Shot (Top Choice) & 0.62 & 23.9 & 41.4 & 106 \\
    {\bf{GPT-J-6B}} (6B) & \setrow{\bfseries} 4-Shot (Top Choice) & 0.78 & 21.9 & 31.8 & 81 \\
\bottomrule
\end{tabular}
}
\caption{Four-shot performances of the off-the-shelf ``small'' language models on the clean version of the \shakespeare corpus. In this few-shot setup, we included a simple natural-language task description and four illustrative examples in the prompt. We note that GPT-J-6B was able to ``translate'' sentences written in Elizabethan English to Modern English successfully, achieving a transfer accuracy score of $78\%$, reference BLEU score of 21.9, and perplexity value of 81.
}
\label{tab:ShakespeareCleanFewShotFullResults}
\end{table*}

\begin{table*}[t]
\small 
\centering
\scalebox{0.90}{
\begin{tabular}{c | >{\rowmac}c | >{\rowmac}c  >{\rowmac}c  >{\rowmac}c  >{\rowmac}c | >{\rowmac}c  >{\rowmac}c  >{\rowmac}c  >{\rowmac}c <{\clearrow}}
\toprule
&  & \multicolumn{4}{c|}{\bf{Positive $\to$ Negative}} & \multicolumn{4}{c}{\bf{Negative $\to$ Positive}} \\
\bf{Model} &  \bf{Setting} & \bf{Acc} & \bf{\emph{r}-sBLEU} & \bf{\emph{s}-sBLEU} & \bf{PPL} & \bf{Acc} & \bf{\emph{r}-sBLEU} & \bf{\emph{s}-sBLEU} & \bf{PPL} \\ 
\toprule
    \bf{Style-Embedding} & \multirowcell{4}{\textbf{\citet{li2018delete}}} & 0.33 & 16.2 & 33.2 & 265 & 0.47 & 13.1 & 29.0 & 287 \\
    \bf{CrossAligned} & & 0.66 & 2.2 & 3.0 & 93 & 0.74 & 1.7 & 2.4 & 96 \\
    \bf{DeleteAndRetrieve} & & 0.49 & 33.3 & 60.3 & 120 & 0.51 & 26.7 & 53.5 & 113 \\
    \bf{TemplateBased} & & 0.65 & 38.1 & 70.5 & 243 & 0.56 & 31.0 & 65.7 & 200 \\
\midrule
\multirowcell{4}{\bf{GPT-2-Small} \\ (117M)} & 0-Shot (Top Choice) & 0.31 & 18.9 & 34.9 & 49 & 0.18 & 17.7 & 38.1 & 48 \\
    & 4-Shot (Top Choice) & 0.32 & 23.8 & 46.1 & 94 & 0.25 & 27.7 & 58.2 & 67 \\ 
    & \setrow{\bfseries} 4-Shot (RC: 3, IF) & 0.42 & 21.1 & 39.2 & 68 & 0.32 & 24.6 & 51.0 & 69 \\ 
    & 4-Shot (RC: 3, FS) & 0.38 & 23.0 & 43.5 & 73 & 0.30 & 27.2 & 52.4 & 77 \\ 
\midrule
\multirowcell{4}{\bf{GPT-2-Medium} \\ (345M)}  &0-Shot (Top Choice) & 0.48 & 21.9 & 36.7 & 68 & 0.32 & 20.1 & 38.0 & 57 \\
    & 4-Shot (Top Choice) & 0.44 & 12.3 & 17.8 & 78 & 0.42 & 11.5 & 17.8 & 72 \\
    & \setrow{\bfseries} 4-Shot (RC: 3, IF) & 0.58 & 12.6 & 18.6 & 66 & 0.55 & 10.2 & 15.0 & 53 \\
    & 4-Shot (RC: 3, FS) & 0.52 & 7.2 & 10.0 & 59 & 0.50 & 5.6 & 9.0 & 56 \\
\midrule
\multirowcell{4}{\bf{GPT-2-Large} \\ (774M)} &0-Shot (Top Choice) & 0.44 & 28.6 & 49.1 & 70 & 0.28 & 26.0 & 51.2 & 55 \\
    & 4-Shot (Top Choice) & 0.47 & 17.1 & 27.7 & 54 & 0.32 & 15.0 & 27.4 & 101 \\
    & \setrow{\bfseries} 4-Shot (RC: 3, IF) & 0.60 & 15.4 & 24.7 & 62 & 0.43 & 15.6 & 27.7 & 59 \\
    & 4-Shot (RC: 3, FS) & 0.55 & 21.6 & 33.1 & 55 & 0.35 & 20.0 & 34.7 & 53 \\
\midrule
\multirowcell{4}{\bf{GPT-2-XL} \\ (1558M)} & 0-Shot (Top Choice) & 0.46 & 21.5 & 35.4 & 59 & 0.32 & 22.3 & 41.4 & 70 \\
    & 4-Shot (Top Choice) & 0.63 & 13.7 & 20.3 & 65 & 0.44 & 14.5 & 22.3 & 60 \\
    & \setrow{\bfseries} 4-Shot (RC: 3, IF) & 0.70 & 11.5 & 17.2 & 77 & 0.56 & 13.2 & 19.9 & 50 \\
    & 4-Shot (RC: 3, FS) & 0.66 & 11.5 & 16.6 & 54 & 0.50 & 14.8 & 19.7 & 58 \\
\midrule
\multirowcell{4}{\bf{GPT-Neo-1.3B} \\ (1.3B)} & 0-Shot (Top Choice) & 0.49 & 11.8 & 19.9 & 34 & 0.31 & 10.9 & 20.5 & 35 \\
    & 4-Shot (Top Choice) & 0.53 & 22.1 & 35.8 & 68 & 0.34 & 22.0 & 39.6 & 67 \\
    & \setrow{\bfseries} 4-Shot (RC: 3, IF) & 0.60 & 21.2 & 33.4 & 66 & 0.39 & 21.6 & 36.2 & 65 \\
    & 4-Shot (RC: 3, FS) & 0.56 & 22.2 & 33.1 & 67 & 0.32 & 20.9 & 32.6 & 63 \\
\midrule
\multirowcell{4}{\bf{GPT-Neo-2.7B} \\ (2.7B)} & 0-Shot (Top Choice) & 0.48 & 21.4 & 36.5 & 55 & 0.28 & 23.7 & 45.9 & 57 \\
    & 4-Shot (Top Choice) & 0.52 & 22.3 & 33.7 & 74 & 0.35 & 22.3 & 39.5 & 74 \\
    & \setrow{\bfseries} 4-Shot (RC: 3, IF) & 0.60 & 21.7 & 32.3 & 69 & 0.42 & 20.6 & 34.9 & 66 \\
    & 4-Shot (RC: 3, FS) & 0.55 & 21.2 & 30.2 & 65 & 0.40 & 19.2 & 29.7 & 60 \\
\midrule
\multirowcell{4}{\bf{GPT-J-6B} \\ (6B)} & 0-Shot (Top Choice) & 0.41 & 26.1 & 44.6 & 57 & 0.33 & 27.1 & 47.7 & 72 \\
& 4-Shot (Top Choice)  & 0.59 & 20.5 & 31.9 & 69 & 0.46 & 18.1 & 28.8 & 60 \\
    & \setrow{\bfseries} 4-Shot (RC: 3, IF) & 0.65 & 21.5 & 31.4 & 70 & 0.52 & 19.3 & 29.3 & 58 \\
    & 4-Shot (RC: 3, FS)  & 0.64 & 17.0 & 24.7 & 61 & 0.50 & 18.6 & 25.7 & 59 \\
\bottomrule
\end{tabular}
}
\caption{Four-shot results on \amazon-clean. We show the average results under (i) the zero-shot setting (\emph{0-Shot (Top Choice)}); (ii) the four-shot setting in which we chose the top beam search result (\emph{4-Shot (Top Choice)}; (iii) the four-shot setting in which we generated three outputs from the model, re-scored and re-ranked them according to the textual similarity and style factors---ignoring the fluency aspect---(\emph{4-Shot (RC: 3, IF)}); and (iv) the four-shot setting in which we generated three outputs from the model, re-scored and re-ranked them according to the textual similarity, style, and fluency factors  (\emph{4-Shot (RC: 3, FS)}. (Here, ``RC'' denotes to the re-ranking-and-choosing method, ``IF'' ignoring fluency, and ``FS'' full set (meaning that we consider all the textual similarity, transfer accuracy, and fluency criteria). First, we stress that proving few-shot examples in the input resulted in 10-15\% improvements in the accuracy scores in most of our models (see 0-shot results vs. 4-shot results). Second, we highlight that some of our off-the-shelf models (e.g., GPT-2-XL and GPT-J-6B) performed on par with, and even succeeded the performances of, the specially-tailored models of \citet{li2018delete} along certain metrics. (For instance, our off-the-shelf models achieve significantly lower perplexity rates than theirs.) Third, we note that the Prompt-and-Rerank method (described in \S\ref{subsec:reranking_alg}) seems to boost the models' performances in almost all the cases. Fourth, we note that 4-Shot (RC: 3, IF) often performs noticeably better than 4-Shot (RC: 3, FS) across all the models, suggesting that we may not need to include the fluency factor in our re-scoring calculations after all.
}
\label{tab:AmazonCleanFewShotFullResults}
\end{table*}

\newpage

\begin{table*}[!h]
\small 
\centering
\scalebox{0.90}{
\begin{tabular}{c | >{\rowmac}c | >{\rowmac}c  >{\rowmac}c  >{\rowmac}c  >{\rowmac}c | >{\rowmac}c  >{\rowmac}c  >{\rowmac}c  >{\rowmac}c <{\clearrow}}
\toprule
&  & \multicolumn{4}{c|}{\bf{Positive $\to$ Negative}} & \multicolumn{4}{c}{\bf{Negative $\to$ Positive}} \\
\bf{Model} &  \bf{Setting} & \bf{Acc} & \bf{\emph{r}-sBLEU} & \bf{\emph{s}-sBLEU} & \bf{PPL} & \bf{Acc} & \bf{\emph{r}-sBLEU} & \bf{\emph{s}-sBLEU} & \bf{PPL} \\ 
\toprule
\multicolumn{1}{c|}{\textbf{BackTranslation}} & \multicolumn{1}{c|}{\textbf{\citep{prabhumoye2018style}}}&  0.90 &  2.0  &  2.7  &  120 &  0.99 &  1.9  &  2.6  &  64 \\ \midrule

\multicolumn{1}{c|}{\textbf{UnpairedRL}} & \multicolumn{1}{c|}{\textbf{\citep{xu2018unpaired}}}             &  0.42 &  16.1 &  46.0 &  408 &  0.56 &  17.5 &  45.3 &  362 \\ 
\midrule
\multicolumn{1}{c|}{\textbf{CrossAlignment}} & \multicolumn{1}{c|}{\textbf{\citep{shen2017style}}}          & 0.72 &  7.3  &  19.3 &  244 &  0.74 &  8.3  &  19.3 &  190 \\ \midrule
\textbf{Multidecoder} & \multirowcell{2}{\textbf{\citep{fu2018style}}}  & 0.42 &  13.4 &  43.2 &  376 &  0.49 &  12.6 &  35.5 &  369 \\
\textbf{StyleEmbedding} &  & 0.08 &  19.7 &  71.3 &  154 &  0.10 &  18.9 &  62.7 &  197 \\ \midrule
\bf{Style-Embedding} & \multirowcell{6}{{\textbf{\citet{li2018delete}}}} & 0.08 & 19.7 & 71.3 & 154 & 0.10 & 18.9 & 62.7 & 197 \\
    \bf{Delete-Only} & & 0.89 & 12.7 & 33.1 & 195 & 0.81 & 14.0 & 34.7 & 169 \\
    \bf{Retrieve-Only} &  & 1.00 &  1.1  &  2.1  &  93  &  0.98 &  1.8  &  2.8  &  86 \\
    \bf{CrossAligned} &  & 0.72 & 7.3 & 19.3 & 244 & 0.74 & 8.3 & 19.3 & 190 \\
    \setrow{\bfseries} DeleteAndRetrieve &  & 0.90 & 14.5 & 36.8 & 279 & 0.89 & 14.8 & 35.9 & 100 \\
    \bf{TemplateBased} &  & 0.84 & 21.2 & 55.2 & 289 & 0.83 & 20.9 & 55.7 & 190 \\
\midrule
\multicolumn{1}{c|}{\textbf{DualR}} & \multicolumn{1}{c|}{\textbf{\citep{luo2019dual}}}                     & 0.91 &  26.5 &  58.7 &  125 &  0.85 &  25.3 &  58.8 &  141 \\ \midrule
\multicolumn{1}{c|}{\textbf{B-GST}} & \multicolumn{1}{c|}{\textbf{\citep{sudhakar2019transforming}}}  & 0.83 &  19.8 &  46.8 &  153 &  0.79 &  23.4 &  46.1 &  163 \\ 
\midrule
\bf{Multi-Class} & \multirowcell{2}{(\textbf{StyleTransformer}, \\ \textbf{\citet{dai2019style}})} & 0.94 & 26.3 & 61.0 & 177 & 0.77 & 26.5 & 65.0 & 173 \\
\setrow{\bfseries} Conditional & & 0.95 & 22.6 & 52.6 & 211 & 0.87 & 23.1 & 53.0 & 234 \\
\midrule
\multirowcell{4}{\bf{GPT-2-Small} \\ (117M)} & \setrow{\bfseries} 0-Shot (Top Choice) & 0.36 & 6.6 & 19.3 & 28 & 0.12 & 8.2 & 25.7 & 29 \\
    & 4-Shot (Top Choice) & 0.08 & 24.8 & 75.2 & 94 & 0.10 & 23.1 & 74.1 & 81 \\
    & 4-Shot (RC: 3, IF) & 0.14 & 21.5 & 71.0 & 73 & 0.17 & 18.5 & 58.5 & 99 \\
    & 4-Shot (RC: 3, FS) & 0.06 & 26.7 & 84.6 & 72 & 0.09 & 27.9 & 85.4 & 68 \\
\midrule
\multirowcell{4}{\bf{GPT-2-Medium} \\ (345M)}  &0-Shot (Top Choice) & 0.65 & 7.3 & 19.9 & 37 & 0.33 & 10.5 & 28.6 & 41 \\
    & 4-Shot (Top Choice) & 0.49 & 14.5 & 34.1 & 72 & 0.35 & 13.3 & 35.2 & 56 \\
    & \setrow{\bfseries} 4-Shot (RC: 3, IF) & 0.68 & 15.0 & 35.1 & 69 & 0.53 & 12.6 & 29.8 & 45 \\
    & 4-Shot (RC: 3, FS) & 0.43 & 20.4 & 46.4 & 74 & 0.40 & 17.7 & 43.5 & 48 \\
\midrule
\multirowcell{4}{\bf{GPT-2-Large} \\ (774M)} &0-Shot (Top Choice) & 0.67 & 12.0 & 28.8 & 40 & 0.30 & 12.5 & 33.8 & 30 \\
    & 4-Shot (Top Choice) & 0.79 & 16.6 & 32.8 & 84 & 0.57 & 14.5 & 31.0 & 74 \\
    & \setrow{\bfseries} 4-Shot (RC: 3, IF) & 0.79 & 10.6 & 24.7 & 79 & 0.58 & 12.1 & 30.3 & 53 \\
    & 4-Shot (RC: 3, FS) & 0.58 & 23.0 & 56.8 & 76 & 0.45 & 22.1 & 53.5 & 64 \\
\midrule
\multirowcell{4}{\bf{GPT-2-XL} \\ (1558M)} & 0-Shot (Top Choice) & 0.73 & 8.6 & 21.3 & 35 & 0.46 & 11.4 & 25.9 & 35 \\
    & 4-Shot (Top Choice) & 0.63 & 13.7 & 20.3 & 65 & 0.44 & 14.5 & 22.3 & 60 \\
    & \setrow{\bfseries} 4-Shot (RC: 3, IF) & 0.87 & 14.8 & 28.7 & 65 & 0.72 & 12.0 & 25.3 & 55 \\
    & 4-Shot (RC: 3, FS) & 0.77 & 21.1 & 38.7 & 85 & 0.62 & 18.0 & 35.2 & 70 \\
\midrule
\multirowcell{4}{\bf{GPT-Neo-1.3B} \\ (1.3B)} & 0-Shot (Top Choice) & 0.69 & 4.6 & 10.5 & 22 & 0.37 & 6.3 & 15.3 & 23 \\
    & 4-Shot (Top Choice) & 0.78 & 14.8 & 30.2 & 58 & 0.45 & 14.5 & 32.0 & 56 \\
    & \setrow{\bfseries} 4-Shot (RC: 3, IF) & 0.85 & 14.6 & 30.1 & 59 & 0.61 & 13.1 & 28.3 & 42 \\
    & 4-Shot (RC: 3, FS) & 0.77 & 22.5 & 46.1 & 87 & 0.49 & 23.5 & 46.1 & 72 \\
\midrule
\multirowcell{4}{\bf{GPT-Neo-2.7B} \\ (2.7B)} & 0-Shot (Top Choice) & 0.64 & 7.0 & 19.0 & 24 & 0.28 & 11.0 & 32.6 & 29 \\
    & 4-Shot (Top Choice) & 0.83 & 22.8 & 42.7 & 89 & 0.42 & 21.8 & 47.1 & 89 \\
    & \setrow{\bfseries} 4-Shot (RC: 3, IF) & 0.88 & 23.5 & 45.8 & 96 & 0.52 & 22.0 & 48.0 & 69 \\
    & 4-Shot (RC: 3, FS) & 0.80 & 24.5 & 44.5 & 87 & 0.48 & 23.9 & 48.4 & 68 \\
\midrule
\multirowcell{4}{\bf{GPT-J-6B} \\ (6B)} & 0-Shot (Top Choice) & 0.61 & 14.3 & 34.7 & 49 & 0.48 & 13.5 & 30.6 & 43 \\
& 4-Shot (Top Choice) & 0.81 & 25.3 & 50.5 & 107 & 0.52 & 21.7 & 48.7 & 82 \\
    & \setrow{\bfseries} 4-Shot (RC: 3, IF) & 0.87 & 23.0 & 47.7 & 80 & 0.65 & 20.2 & 44.6 & 58 \\
    & 4-Shot (RC: 3, FS) & 0.79 & 25.9 & 51.5 & 78 & 0.55 & 26.3 & 50.0 & 67 \\
\bottomrule
\end{tabular}
}
\caption{Four-shot results on \yelp-clean. As before, we detail the average results under different zero- and few-shot settings: Overall, our few-shot results on \yelp-clean are consistent with those on \amazon-clean, as reported in Table~\ref{tab:AmazonCleanFewShotFullResults}. GPT-2-XL and GPT-J-6B models, amongst all the models, have achieved the most successful performances, leveling themselves almost with the custom-made (trained) state-of-the-art models. We present some of the generated examples from these models in Table~\ref{tab:generations_yelp}.}
\label{tab:YelpCleanFewShotFullResults}
\end{table*}

\newpage

\begin{table*}[!h]
\small 
\centering
\scalebox{0.95}{
\begin{tabular}{c | >{\rowmac}c | >{\rowmac}c  >{\rowmac}c  >{\rowmac}c  >{\rowmac}c <{\clearrow}}
\toprule
&  & \multicolumn{4}{c}{ \bf{Ungrammatical $\to$ Grammatical} } \\
\bf{Model} &  \bf{Setup} & \bf{GLEU} & \bf{\emph{r}-sBLEU} & \bf{\emph{s}-sBLEU} & \bf{PPL} \\
\toprule 
    {\bf{GPT-2-Small}} (117M) & 4-Shot (Top Choice) & 35.9 & 74.8 & 91.5 & 76 \\
    {\bf{GPT-2-Medium}} (345M) & 4-Shot (Top Choice) & 19.9 & 38.0 & 40.6 & 63 \\
    {\bf{GPT-2-Large}} (774M) & 4-Shot (Top Choice) &  30.0 & 56.8 & 64.1 & 55 \\
    {\bf{GPT-2-XL}} (1558M) & 4-Shot (Top Choice) & 24.8 & 46.2 & 47.0 & 57 \\
    {\bf{GPT-Neo-1.3B}} (1.3B) & 4-Shot (Top Choice) & 26.6 & 48.4 & 49.4 & 54 \\
    {\bf{GPT-Neo-2.7B}} (2.7B) & 4-Shot (Top Choice) & 34.5 & 57.4 & 54.1 & 40 \\
    \setrow{\bfseries} {\bf{GPT-J-6B}} (6B) & 4-Shot (Top Choice) & 40.0 & 64.8 & 59.1 & 48 \\
\bottomrule
\end{tabular}
}
\caption{Four-shot performances of the off-the-shelf ``small'' language models on the clean version of the \jfleg corpus. In this task, as a baseline, we consider the model which directly copies its input---we call this model ``copy-input'' model; this model achieves a GLEU score score 37.7. All but GPT-J-6B fail to beat the performance of the baseline ``copy-input'' model. GPT-J-6B, on the other hand, achieves a GLEU score of 40.0. Small language models fail, in fact rather miserably, at this grammatical error correction task. There is therefore an open room for improvement. We hope that our results will encourage researchers to come up with more effective ways to utilize pre-trained language models to solve this challenging problem.}
\label{tab:JFLEG_clean_few_shot_results}
\end{table*}

\newpage

\begin{table*}[!h]
\small 
\centering
\scalebox{0.95}{
\begin{tabular}{c | >{\rowmac}c | >{\rowmac}c  >{\rowmac}c  >{\rowmac}c  >{\rowmac}c <{\clearrow}}
\toprule
&  & \multicolumn{4}{c}{ \bf{Informal $\to$ Formal} } \\
\bf{Model} &  \bf{Setup} & \bf{Accuracy} & \bf{\emph{r}-sBLEU} & \bf{\emph{s}-sBLEU} & \bf{PPL} \\
\toprule 
    {\bf{GPT-2-Small}} (117M) & 4-Shot (Top Choice) & 0.85 & 6.1 & 8.7 & 41 \\
    {\bf{GPT-2-Medium}} (345M) & 4-Shot (Top Choice) & 0.76 & 12.9 & 16.2 & 39 \\
    {\bf{GPT-2-Large}} (774M) & 4-Shot (Top Choice) & 0.78 & 23.2 & 31.3 & 33 \\
    {\bf{GPT-2-XL}} (1558M) & 4-Shot (Top Choice) & 0.82 & 32.7 & 41.9 & 58 \\
    \setrow{\bfseries} {\bf{GPT-Neo-1.3B}} (1.3B) & 4-Shot (Top Choice) & 0.85 & 36.4 & 49.6 & 68 \\
    {\bf{GPT-Neo-2.7B}} (2.7B) & 4-Shot (Top Choice) & 0.81 & 50.0 & 61.2 & 64 \\
    {\bf{GPT-J-6B}} (6B) & 4-Shot (Top Choice) & 0.69 & 47.9 & 52.3 & 49 \\
\bottomrule
\end{tabular}
}
\caption{Four-shot results on \gyafc-clean. We highlight that most of the off-the-shelf ``small'' language models could obtain at least 80\% accuracy in the informal to formal direction. Amongst all the models, GPT-2-XL, GPT-Neo-1.3B, and GPT-Neo-2.7B appeared to be most successful, achieving not only high accuracy scores but also high BLEU scores and relatively low perplexity rates. 
}
\label{tab:GYAFC_clean_few_shot_results}
\end{table*}

\begin{table*}[!h]
\small 
\centering
\scalebox{0.95}{
\begin{tabular}{>{\rowmac}c  | >{\rowmac}c  >{\rowmac}c  >{\rowmac}c  >{\rowmac}c}
\toprule
\bf{Model} & \bf{Correct-Class Accuracy} & \bf{Opposite-Class Accuracy} &\bf{\emph{reference}-sBLEU} \\ \midrule
GPT-2-Small & 0.42 & 0.46 & 51.9 \\
GPT-2-Medium & 0.46 & 0.46 & 60.3 \\
GPT-2-Large & 0.53 & 0.35 &65.6 \\
GPT-2-XL & 0.56 & 0.38 &68.5 \\
GPT-Neo-1.3B & 0.55 & 0.37 &67.3 \\
GPT-Neo-2.7B & 0.57 & 0.38 &69.6 \\
\setrow{\bfseries} GPT-J-6B & 0.74 & 0.21 &81.9 \\
\bottomrule
\end{tabular}
}
\vspace{-0.5em}
\caption{Four-shot performances of the off-the-shelf ``small'' language models on the symbolic manipulation task (\symb) defined in \S\ref{subsec:datasets_and_tasks}. Correct-Class accuracy refers to the accuracy of the model under exact-string matching, whereas Opposite-Class accuracy refers to the fraction of the cases for which the model copied and placed the right input words in the output but verbalized the incorrect (opposite) inequality symbol, that is writing ``less than'' instead of ``greater than'' or vice versa in between the expressions (for instance, the ground-truth might be ``olive is greater than cat'', but the model might have generated ``olive is less than cat.'') It was surprising to discover that most models failed to go beyond 60\% accuracy on this small dataset. GPT-J-6B, on the other hand, outperformed all the other models, achieving an accuracy score of 74\% on this task. We also remark that of the cases for which the models failed to generate the correct output, they often were able to copy the appropriate words from the input but failed to write the correct inequality symbol at the end.}
\label{tab:SymbolicManipulation_few_shot_results}
\end{table*}

\newpage
\begin{table*}[!h]
\small 
\centering
\scalebox{0.90}{
\begin{tabular}{c | >{\rowmac}c | >{\rowmac}c  >{\rowmac}c  >{\rowmac}c  >{\rowmac}c | >{\rowmac}c  >{\rowmac}c  >{\rowmac}c  >{\rowmac}c <{\clearrow}}
\toprule
&  & \multicolumn{4}{c|}{\bf{Positive $\to$ Negative}} & \multicolumn{4}{c}{\bf{Negative $\to$ Positive}} \\
\bf{Model} &  \bf{Setting} & \bf{Acc} & \bf{\emph{r}-sBLEU} & \bf{\emph{s}-sBLEU} & \bf{PPL} & \bf{Acc} & \bf{\emph{r}-sBLEU} & \bf{\emph{s}-sBLEU} & \bf{PPL} \\ 
\toprule
\multirowcell{3}{{\bf{GPT-2-Small}} \\ (117M)} & 4-Shot, Vanilla & 0.14 & 25.3 & 82.1 & 79 & 0.13 & 24.9 & 80.6 & 75 \\
    & 4-Shot, Contrastive & 0.14 & 21.5 & 71.   0 & 73 & 0.17 & 18.5 & 58.5 & 99 \\
    & \setrow{\bfseries} 4-Shot, Negation-v1 & 0.05 & 25.2 & 83.7 & 72 & 0.07 & 23.1 & 75.3 & 68 \\
    & 4-Shot, Negation-v2 & 0.06 & 25.4 & 84.5 & 75 & 0.08 & 25.3 & 83.0 & 74 \\
\midrule
\multirowcell{4}{{\bf{GPT-2-Medium}} \\(345M)} & 4-Shot, Vanilla & 0.63 & 19.7 & 49.4 & 75 & 0.38 & 17.7 & 49.7 & 62 \\
    & \setrow{\bfseries} 4-Shot, Contrastive & 0.68 & 15.0 & 35.1 & 69 & 0.53 & 12.6 & 29.8 & 45 \\
    & 4-Shot, Negation-v1 & 0.34 & 15.7 & 41.6 & 63 & 0.22 & 15.4 & 42.3 & 55 \\
    & 4-Shot, Negation-v2 & 0.36 & 14.1 & 39.2 & 63 & 0.30 & 12.6 & 36.2 & 51 \\
\midrule
\multirowcell{4}{{\bf{GPT-2-Large}} \\ (774M)} & 4-Shot, Vanilla & 0.75 & 16.2 & 38.9 & 60 & 0.52 & 17.2 & 45.6 & 55 \\
    & \setrow{\bfseries} 4-Shot, Contrastive & 0.79 & 10.6 & 24.7 & 79 & 0.58 & 12.1 & 30.3 & 53 \\
    & 4-Shot, Negation-v1 & 0.41 & 10.8 & 30.1 & 79 & 0.27 & 13.4 & 36.7 & 59 \\
    & 4-Shot, Negation-v2 & 0.14 & 16.9 & 52.1 & 57 & 0.22 & 12.6 & 36.3 & 59 \\
\midrule
\multirowcell{4}{{\bf{GPT-2-XL}} \\ (1558M)} & 4-Shot, Vanilla & 0.86 & 15.6 & 32.0 & 59 & 0.70 & 13.8 & 29.9 & 58 \\
    & \setrow{\bfseries} 4-Shot, Contrastive & 0.87 & 14.8 & 28.7 & 65 & 0.72 & 12.0 & 25.3 & 55 \\
    & 4-Shot, Negation-v1 & 0.83 & 11.8 & 24.1 & 81 & 0.50 & 14.9 & 32.0 & 54 \\
    & 4-Shot, Negation-v2 & 0.53 & 19.0 & 43.5 & 77 & 0.51 & 16.9 & 37.7 & 61 \\
\midrule
\multirowcell{4}{{\bf{GPT-Neo-1.3B}} \\ (1.3B)} & 4-Shot, Vanilla & 0.80 & 17.2 & 38.5 & 80 & 0.52 & 14.5 & 35.6 & 50 \\
    & \setrow{\bfseries} 4-Shot, Contrastive & 0.85 & 14.6 & 30.1 & 59 & 0.61 & 13.1 & 28.3 & 42 \\
    & 4-Shot, Negation-v1 & 0.79 & 16.1 & 34.7 & 72 & 0.00 & 0.0 & 0.0 & 0 \\
    & 4-Shot, Negation-v2 & 0.57 & 16.5 & 40.6 & 67 & 0.00 & 0.0 & 0.0 & 0 \\
\midrule
\multirowcell{4}{{\bf{GPT-Neo-2.7B}} \\ (2.7B)} & 4-Shot, Vanilla & 0.86 & 24.7 & 51.2 & 104 & 0.43 & 24.3 & 54.9 & 74 \\
    & \setrow{\bfseries} 4-Shot, Contrastive & 0.88 & 23.5 & 45.8 & 96 & 0.52 & 22.0 & 48.0 & 69 \\
    & 4-Shot, Negation-v1 & 0.80 & 22.5 & 47.8 & 79 & 0.00 & 0.0 & 0.0 & 0 \\
    & 4-Shot, Negation-v2 & 0.76 & 22.0 & 48.2 & 85 & 0.00 & 0.0 & 0.0 & 0 \\
\midrule
\multirowcell{4}{{\bf{GPT-J-6B}}\\ (6B)} & \setrow{\bfseries} 4-Shot, Vanilla & 0.90 & 23.5 & 51.0 & 85 & 0.62 & 22.8 & 49.9 & 63 \\
    & 4-Shot, Contrastive & 0.87 & 23.0 & 47.7 & 80 & 0.65 & 20.2 & 44.6 & 58 \\
    & 4-Shot, Negation-v1 & 0.82 & 23.6 & 50.9 & 85 & 0.53 & 24.3 & 54.7 & 65 \\
    & 4-Shot, Negation-v2 & 0.73 & 25.1 & 59.1 & 95 & 0.48 & 25.2 & 58.1 & 66 \\
\bottomrule
\end{tabular}
}
\caption{Four-shot performances of the off-the-shelf ``small'' language models on the \yelp-clean dataset under different prompting protocols. Rows are bolded according to best accuracy scores. Across all models, the vanilla and contrastive prompting protocols appear to yield the most favourable results, though the contrastive prompting protocol typically performs better than the vanilla prompting protocol along the accuracy dimension, but not necessarily along the \emph{r}-sBLEU dimension. We, however, note that in the case of GPT-J-6B, vanilla prompting performs on par with contrastive prompting overall. }
\label{tab:yelp_clean_different_phrasing}
\end{table*}
\newpage
\begin{table*}[!h]
\small 
\centering
\scalebox{1.0}{
\begin{tabular}{l | p{0.75\textwidth}}
\toprule
\bf{\quad Model} & \bf{\quad \quad\quad\quad\quad \quad\quad\quad\quad \quad \quad \quad\quad Text} \\ 
\toprule
\textcolor{gray}{\bf{Input \emph{(P$\to$N)}}} & \textcolor{gray}{the staff is \textcolor{cyan}{awesome} and location is \textcolor{cyan}{right in the heart} of old town!} \\
\cdashlinelr{1-2}
GPT-2-Small & the staff is awesome and location is right in the heart of old town! \\  
GPT-2-Medium & the staff was great \textcolor{purple}{but} the location was an even higher priority. \\  
GPT-2-Large & the staff is \textcolor{purple}{awful} and location is \textcolor{purple}{not in the center} of old town! \\  
GPT-2-XL & the staff was \textcolor{purple}{kind of awful}, location was \textcolor{purple}{kind of a shithole} \\  
GPT-Neo-1.3B & the staff is \textcolor{purple}{awful} and location is \textcolor{purple}{just a couple minutes away from the city center!} \\  
GPT-Neo-2.7B & the staff is \textcolor{purple}{terrible} and location is \textcolor{purple}{just outside of} old town! \\  
GPT-J-6B & the staff is \textcolor{purple}{terrible} and location is \textcolor{purple}{in the middle of nowhere!} \\  
\midrule
\textcolor{gray}{\bf{Input \emph{(P$\to$N)}}} & \textcolor{gray}{he always looks \textcolor{cyan}{gorgeous} and is \textcolor{cyan}{so happy} when i pick him up!} \\  
\cdashlinelr{1-2}
GPT-2-Small & he always looks gorgeous and is so happy when i pick him up! \\  
GPT-2-Medium & he always looks \textcolor{purple}{so sad and upset} by the fact that i pick him up! \\  
GPT-2-Large & he always looks \textcolor{purple}{ugly} and is \textcolor{purple}{so sad} when i pick him up! \\  
GPT-2-XL & he \textcolor{purple}{never looks gorgeous} and is \textcolor{purple}{so sad} when i pick him up! \\  
GPT-Neo-1.3B & he always looks \textcolor{purple}{awful} and is \textcolor{purple}{so sad} when i pick him up! \\  
GPT-Neo-2.7B & he always looks \textcolor{purple}{horrible} and is \textcolor{purple}{so unhappy} when i pick him up! \\  
GPT-J-6B & he always looks \textcolor{purple}{ugly} and is \textcolor{purple}{so sad }when i pick him up! \\ 
\midrule
\textcolor{gray}{\bf{Input \emph{(N$\to$P)}}} & \textcolor{gray}{the food was \textcolor{cyan}{low quality} and thrown together quickly with \textcolor{cyan}{little} flavor.} \\
\cdashlinelr{1-2}
GPT-2-Small & the food was low quality and thrown together quickly with little flavor. \\  
GPT-2-Medium & the food was \textcolor{purple}{high quality} and thrown together quickly with a \textcolor{purple}{great} flavor. \\  
GPT-2-Large & the food was \textcolor{purple}{excellent} and \textcolor{purple}{was cooked to perfection by very skilled and professional server-chefs.} \\  
GPT-2-XL & the food was \textcolor{purple}{high quality} and \textcolor{purple}{tasted really good.} \\  
GPT-Neo-1.3B & the food was \textcolor{purple}{high quality} and \textcolor{purple}{tasted amazing!} \\  
GPT-Neo-2.7B & the food was \textcolor{purple}{fantastic} and \textcolor{purple}{presented exceptionally with excellent quality and excellent flavors!} \\  
GPT-J-6B & the food was \textcolor{purple}{delicious} and \textcolor{purple}{prepared with a lot of care!} \\ 
\midrule
\textcolor{gray}{\bf{Input \emph{(N$\to$P)}}} & \textcolor{gray}{we went there on a \textcolor{cyan}{busy} holiday weekend and the service was \textcolor{cyan}{extremely slow}.} \\
\cdashlinelr{1-2}
GPT-2-Small & we went there on a busy holiday weekend and the service was extremely slow. \\
GPT-2-Medium & we went there on a busy holiday weekend and the service was \textcolor{purple}{exceptional!} \\
GPT-2-Large & we did go there on a busy holiday weekend--it was \textcolor{purple}{fast!} \\
GPT-2-XL & we went there on a \textcolor{purple}{quiet} weekend and the service was \textcolor{purple}{very fast}. \\
GPT-Neo-1.3B & we went there on a busy holiday weekend and the service was \textcolor{purple}{exceptional!} \\
GPT-Neo-2.7B & we went there on a \textcolor{purple}{slow} holiday weekend and the service was \textcolor{purple}{extremely fast} \\
GPT-J-6B & we went there on a busy holiday weekend and the service was \textcolor{purple}{extremely fast}. \\
\midrule
\textcolor{gray}{\bf{Input \emph{(P$\to$N)}}} & \textcolor{gray}{i \textcolor{cyan}{love} the outside seating and the reverse happy hour is \textcolor{cyan}{incredibly affordable}.} \\
\cdashlinelr{1-2}
GPT-2-Small & i love the outside seating and the reverse happy hour is unbelievably affordable. \\
\multirowcell{2}{GPT-2-Medium} & i love everything about this place--the music is great, everyone is super nice, and they have amazing desserts that are affordable. \\
GPT-2-Large & i loved the outside seating and the reverse happy hour is \textcolor{purple}{insanely expensive}. \\
GPT-2-XL & i \textcolor{purple}{hate} the outside seating and the reverse happy hour. \\
GPT-Neo-1.3B & i \textcolor{purple}{hate} the outside seating and the reverse happy hour is \textcolor{purple}{incredibly costly}. \\
GPT-Neo-2.7B & i \textcolor{purple}{hate} the outside seating and the reverse happy hour is \textcolor{purple}{outrageously expensive}. \\
GPT-J-6B & i \textcolor{purple}{hate} the outside seating and the reverse happy hour is \textcolor{purple}{incredibly expensive}. \\
\midrule
\textcolor{gray}{\bf{Input \emph{(N$\to$P)}}} & \textcolor{gray}{the service is \textcolor{cyan}{mediocre} and the restaurant is \textcolor{cyan}{just average}.} \\
\cdashlinelr{1-2}
GPT-2-Small & i \textcolor{purple}{love their black tea and hot chocolate selections}! \\
\multirowcell{2}{GPT-2-Medium} & i liked \textcolor{purple}{their black tea and hot chocolate selections}; i was \textcolor{purple}{impressed by the quality of the food and service} \\
GPT-2-Large & the service is \textcolor{purple}{wonderful!} \\
GPT-2-XL & the service is \textcolor{purple}{great}, and the restaurant is \textcolor{purple}{great!} \\
GPT-Neo-1.3B & the service and the restaurant are \textcolor{purple}{just average}. \\
GPT-Neo-2.7B & the service and the restaurant are \textcolor{purple}{just wonderful!} \\
GPT-J-6B & the service is \textcolor{purple}{remarkable} and the restaurant is \textcolor{purple}{amazing}. \\
\bottomrule
\end{tabular}
}
\caption{Qualitative examples of sentiment transfer on \yelp-clean. We manually highlight in blue and red phrases that we find to be associated with the input and target styles, respectively.}
\label{tab:generations_yelp}
\end{table*}

\newpage

\begin{table*}[!h]
\small 
\centering
\scalebox{1.0}{
\begin{tabular}{l | p{0.75\textwidth}}
\toprule
\bf{\quad Model} & \bf{\quad \quad\quad\quad\quad \quad\quad\quad\quad \quad \quad \quad\quad Text} \\ 
\toprule
\textcolor{gray}{\bf{Input \emph{(N$\to$P)}}} & \textcolor{gray}{it works for \textcolor{cyan}{maybe a few minutes} and wears off \textcolor{cyan}{quickly}.} \\
\cdashlinelr{1-2}
GPT-2-Small & it \textcolor{purple}{does not wear} or work well when driving. \\
GPT-2-Medium & i think that it may have something to do with this brand. \\
GPT-2-Large & it works \textcolor{purple}{well for a while}, but it wears off very quickly.  \\
GPT-2-XL & it only does \textcolor{purple}{a few minutes' worth of work and then gives up}. \\
GPT-Neo-1.3B & it can be used \textcolor{purple}{for a few minutes}, but wears off quickly.  \\
GPT-Neo-2.7B & it \textcolor{purple}{lasts a lifetime} and wears off \textcolor{purple}{after a few hours}. \\
GPT-J-6B & it works \textcolor{purple}{for a few hours} and wears \textcolor{purple}{well}. \\
\midrule
\textcolor{gray}{\bf{Input \emph{(N$\to$P)}}} & \textcolor{gray}{i am only giving this item \textcolor{cyan}{num\_num stars} because of the \textcolor{cyan}{absurd} price.} \\
\cdashlinelr{1-2}
GPT-2-Small & i am only giving this item num\_num stars because of the absurd price.  \\
GPT-2-Medium & i will \textcolor{purple}{buy} some num\_num stars and will use them to have an extra day off when i work. \\
GPT-2-Large & i am only giving this item \textcolor{purple}{stars} because it got a \textcolor{purple}{good }review. \\
GPT-2-XL & i will NOT purchase this item again. \\
GPT-Neo-1.3B & i \textcolor{purple}{really liked it} and i will \textcolor{purple}{do it more often}. \\
GPT-Neo-2.7B & i am giving this item num\_num stars, because of the absurd price. \\
GPT-J-6B & i am only giving this item \textcolor{purple}{5 stars} because it \textcolor{purple}{works perfect} and the \textcolor{purple}{price is reasonable}. \\
\midrule
\textcolor{gray}{\bf{Input \emph{(N$\to$P)}}} & \textcolor{gray}{would have been \textcolor{cyan}{alot nicer if} it simply threaded in with the \textcolor{cyan}{correct} alignment.} \\
\cdashlinelr{1-2}
GPT-2-Small & would have been alot nicer if it simply threaded in with the correct alignment.  \\
GPT-2-Medium & it \textcolor{purple}{works perfectly}. \\
GPT-2-Large & would have been \textcolor{purple}{so much better} if the plastic was \textcolor{purple}{a bit thicker}. \\
GPT-2-XL & would have been \textcolor{purple}{great} if it just threaded in \textcolor{purple}{properly}. \\
GPT-Neo-1.3B & would have been \textcolor{purple}{alot nicer} if it simply threaded in with the correct alignment. \\
GPT-Neo-2.7B & it is \textcolor{purple}{very nice but} it does \textcolor{purple}{not fit in properly} at all.  \\
GPT-J-6B & it \textcolor{purple}{works just fine} and \textcolor{purple}{there is zero frustration with the way the threading is implemented}. \\
\midrule
\textcolor{gray}{\bf{Input \emph{(P$\to$N)}}} & \textcolor{gray}{it \textcolor{cyan}{fulfills} my modest requirements, and \textcolor{cyan}{exceeds} my expectations in appearance.} \\
\cdashlinelr{1-2}
GPT-2-Small & it \textcolor{purple}{isn't good enough}. I would buy from this brand again.  \\
GPT-2-Medium & it \textcolor{purple}{doesn't fit my needs} and \textcolor{purple}{falls short in other areas}. \\
\multirowcell{2}{GPT-2-Large} & it \textcolor{purple}{did not fulfill} my modest requirements, and \textcolor{purple}{did not exceed my expectations in appearance}, and \textcolor{purple}{would not fit well in the car}. \\
GPT-2-XL & it \textcolor{purple}{does not meet my expectations} in appearance. \\
GPT-Neo-1.3B & it \textcolor{purple}{does not fit my expectations} and it \textcolor{purple}{does not fulfill} my modest requirements in appearance. \\
GPT-Neo-2.7B & it \textcolor{purple}{doesn t fulfill} my modest requirements, and i found it to be \textcolor{purple}{unsatisfactory}. \\
GPT-J-6B & it \textcolor{purple}{fails to meet my needs}, and is \textcolor{purple}{disappointing to look at}. \\
\midrule
\textcolor{gray}{\bf{Input \emph{(P$\to$N)}}} & \textcolor{gray}{\textcolor{cyan}{great} headset, \textcolor{cyan}{solid} build quality and \textcolor{cyan}{blocks out the wind.}} \\
\cdashlinelr{1-2}
GPT-2-Small & \textcolor{purple}{i hate it} \\
GPT-2-Medium & lots of dust on here and can be worn. \\
GPT-2-Large & \textcolor{purple}{not worth} buying and \textcolor{purple}{not worth the price}. It's just \textcolor{purple}{too large}. \\
GPT-2-XL & \textcolor{purple}{i hate it}, it is \textcolor{purple}{terrible}. \\
GPT-Neo-1.3B & \textcolor{purple}{dreadful} headset, \textcolor{purple}{terrible} quality and \textcolor{purple}{wind blocks out} the audio quality. \\
GPT-Neo-2.7B & \textcolor{purple}{lousy} headset, \textcolor{purple}{hard to unblock the wind}. \\
GPT-J-6B & \textcolor{purple}{great} headset, but it \textcolor{purple}{does not work great in the wind}.  \\
\midrule
\textcolor{gray}{\bf{Input \emph{(P$\to$N)}}} & \textcolor{gray}{i dropped phone \textcolor{cyan}{once} and the case held up \textcolor{cyan}{perfectly}.} \\
\cdashlinelr{1-2}
GPT-2-Small & i dropped phone once and the case held up perfectly.  \\
GPT-2-Medium & i dropped phone as a result. \\
GPT-2-Large & i never dropped my phone once and the case held up fine. \\
GPT-2-XL & i dropped phone \textcolor{purple}{twice} and the case was \textcolor{purple}{damaged}.  \\
GPT-Neo-1.3B & i dropped it and it was \textcolor{purple}{very broken}.  \\
GPT-Neo-2.7B & i dropped phone \textcolor{purple}{as many times as} I dropped a brick. \\
GPT-J-6B & i dropped the phone and the case \textcolor{purple}{broke into pieces}.  \\
\bottomrule
\end{tabular}
}
\caption{Qualitative examples of sentiment transfer on \amazon-clean. We manually highlight in blue and red phrases that we find to be associated with the input and target styles, respectively.}
\label{tab:generations_amaozon}
\end{table*}

\begin{table*}[!h]
\small 
\centering
\scalebox{1.0}{
\begin{tabular}{l | p{0.75\textwidth}}
\toprule
\bf{\quad Model} & \bf{\quad \quad\quad\quad\quad \quad\quad\quad\quad \quad \quad \quad\quad Text} \\ 
\toprule
\textcolor{gray}{\bf{Input}} & \textcolor{gray}{follow \textcolor{cyan}{ur} heart... it's \textcolor{cyan}{neva} wrong} \\
\cdashlinelr{1-2}
GPT-2-Small & follow \textcolor{purple}{my} heart. \\
GPT-2-Large & follow \textcolor{purple}{your} heart... its \textcolor{purple}{nova} wrong... \\
GPT-2-Medium & follow \textcolor{purple}{your} heart... its \textcolor{purple}{nova} wrong \\
GPT-2-XL & follow ur heart and \textcolor{purple}{be the man you always wanted to be and never had}. \\
GPT-J-6B & follow \textcolor{purple}{your} heart... it is \textcolor{purple}{never} wrong \\
GPT-Neo-1.3B & follow \textcolor{purple}{your} heart... its \textcolor{purple}{neva} wrong \\
GPT-Neo-2.7B & follow \textcolor{purple}{your} heart... it's \textcolor{purple}{never} wrong \\
\midrule
\textcolor{gray}{\bf{Input}} & \textcolor{gray}{if \textcolor{cyan}{u you} act like \textcolor{cyan}{dat} all \textcolor{cyan}{da} time she might think \textcolor{cyan}{u} desperate.} \\
\cdashlinelr{1-2}
\multirowcell{2}{GPT-2-Small} & if u act\textcolor{purple}{s} like dat all, she might think that u wanna be happy with something and u dont want dat to break up with you, you will regret it, and she will become jealous. it is good... \\
GPT-2-Medium & if u u are always like dat.. \\
GPT-2-Large & if \textcolor{purple}{you }act like dat all da time she might think \textcolor{purple}{you} desperate... \\
GPT-2-XL & if u u act like dat all da time she might think u desperate :)  \\
GPT-Neo-1.3B & if u you act like dat everytime she thinks u desperately searching. \\
GPT-Neo-2.7B & if u you act like dat all da time she might think u desperate... \\
GPT-J-6B & if \textcolor{purple}{you} act like \textcolor{purple}{that} all \textcolor{purple}{the} time she might think \textcolor{purple}{you} desperate \\
\midrule
\textcolor{gray}{\bf{Input}} & \textcolor{gray}{\textcolor{cyan}{dont} do things \textcolor{cyan}{tht} can get you hurt.} \\
\cdashlinelr{1-2}
GPT-2-Small & but im sorry that u can do it and u can't leave \\
GPT-2-Medium & do things which i'm sure u will like. \\
GPT-2-Large & \textcolor{purple}{don't} do things \textcolor{purple}{that} can get \textcolor{purple}{you} hurt \\
GPT-2-XL & dont do things \textcolor{purple}{that} can end up getting u hurt. \\
GPT-Neo-1.3B & do things that can get you hurt. \\
GPT-Neo-2.7B & \textcolor{purple}{do not} do things \textcolor{purple}{that} can get \textcolor{purple}{you} hurt. \\
GPT-J-6B & dont do things \textcolor{purple}{that} \textcolor{purple}{might hurt you} \\
\bottomrule
\end{tabular}
}
\caption{Qualitative examples of sentiment transfer on \gyafc-clean. We manually highlight in blue and red phrases that we find to be associated with the input and target styles, respectively.}
\vspace{2mm}
\label{tab:generations_gyafc}
\end{table*}

\begin{table*}[!h]
\small 
\centering
\scalebox{1.0}{
\begin{tabular}{l | p{0.75\textwidth}}
\toprule
\bf{\quad Model} & \bf{\quad \quad\quad\quad\quad \quad\quad\quad\quad \quad \quad \quad\quad Text} \\ 
\toprule
\multirowcell{2}[0pt][l]{\textcolor{gray}{\bf{Input}}} & \textcolor{gray}{\textcolor{cyan}{Fo example}, in this \textcolor{cyan}{genaration} people watch movies and \textcolor{cyan}{listen to musics} more than read books and novels.} \\
\cdashlinelr{1-2}
\multirowcell{2}[0pt][l]{GPT-2-Small} & Fo example, in this genaration people watch movies and listen to musics more than read books and novels. \\
GPT-2-Medium & \textcolor{purple}{In this generation people watch movies more than read books and novels}. \\
\multirowcell{2}[0pt][l]{GPT-2-Large} & Fo example, in this genaration people watch movies and listen to musics more than read books and novels. \\
GPT-2-XL & Fo example, in this \textcolor{purple}{exampl}e people watch movies and \textcolor{purple}{listen to music} more than read books. \\
\multirowcell{2}[0pt][l]{GPT-Neo-1.3B} & Example, in this genaration people watch movies and listen to musics more than read books and novels. \\
\multirowcell{2}[0pt][l]{GPT-Neo-2.7B} & Examples, in this \textcolor{purple}{generation}, people watch movies and listen to music, \textcolor{purple}{whereas,} read\textcolor{purple}{ing} books and novels \textcolor{purple}{is limited}. \\
\multirowcell{2}[0pt][l]{GPT-J-6B} & \textcolor{purple}{For example}, in this \textcolor{purple}{generation}, people \textcolor{purple}{are} watch\textcolor{purple}{ing} movies and listen\textcolor{purple}{ing} to music\textcolor{purple}{s} more than read\textcolor{purple}{ing} books and novels. \\
\midrule
\textcolor{gray}{\bf{Input}} & \textcolor{gray}{they are the \textcolor{cyan}{boney} and vital part \textcolor{cyan}{o} the society.} \\
\cdashlinelr{1-2}
GPT-2-Small & They are \textcolor{purple}{the most important part of} the society. \\
GPT-2-Medium & they are the \textcolor{purple}{vital part of} the \textcolor{purple}{kingdom}. \\
GPT-2-Large & they are the vital and boney part\textcolor{purple}{s} \textcolor{purple}{of} the society. \\
GPT-2-XL & they are the vital part, the boney part, and the society. \\
GPT-Neo-1.3B & they are the \textcolor{purple}{vital part of} society.  \\
GPT-Neo-2.7B & they are the boney and vital part \textcolor{purple}{of} society. \\
GPT-J-6B & they are the \textcolor{purple}{bone} and vital part \textcolor{purple}{of} the society. \\
\midrule
\textcolor{gray}{\bf{Input}} & \textcolor{gray}{And I \textcolor{cyan}{am completely disagree} with what the \textcolor{cyan}{auther want} to say.} \\
\cdashlinelr{1-2}
GPT-2-Small & And I am completely disagree with what the auther want to say. \\
GPT-2-Medium & And I am \textcolor{purple}{entirely} disagree with what the auther want to say. \\
GPT-2-Large & And I am completely disagree with what the auther want to say \\
GPT-2-XL & And I \textcolor{purple}{think} the auther want to say \textcolor{purple}{completely differently}. \\
GPT-Neo-1.3B & And I \textcolor{purple}{disagree with what \textcolor{purple}{he} say}. \\
GPT-Neo-2.7B & And I \textcolor{purple}{absolutely} disagree with what the \textcolor{purple}{author} want to say. \\
GPT-J-6B & I \textcolor{purple}{completely} disagree with what the \textcolor{purple}{author} \textcolor{purple}{wants} to say. \\
\bottomrule
\end{tabular}
}
\caption{Qualitative examples on \jfleg-clean. We manually highlight in blue and red phrases that we find to be associated with the input and target styles, respectively.}
\label{tab:generations_jfleg}
\end{table*}

\begin{table*}[!h]
\small 
\centering
\scalebox{0.90}{
\begin{tabular}{l | p{0.88\textwidth}}
\toprule
\bf{\quad Model} & \bf{\quad \quad\quad\quad\quad \quad\quad\quad\quad \quad \quad \quad\quad Text} \\ 
\toprule
\textcolor{gray}{\bf{Input}} & \textcolor{gray}{\textcolor{cyan}{r}omeo, \textcolor{cyan}{doff thy} name; and for that name, which is no part of \textcolor{cyan}{thee}, take all myself} \\
\cdashlinelr{1-2}
GPT-2-Small & speak you, my dear nurse, to me... \\
GPT-2-Medium & \textcolor{purple}{R}omeo, Romeo, Romeo. \\
GPT-2-Large & romo, doff thy name, take all myself, all i'll say is, you're a dirty old man. \\
GPT-2-XL & roome, doff thy name; and for that name, which is no part of thee, take all myself. \\
GPT-Neo-1.3B & romeo, get out of my sight; get out, i'll not have it. \\
GPT-Neo-2.7B & rommy, \textcolor{purple}{get rid of} \textcolor{purple}{your} name; and because \textcolor{purple}{your} name, which is no part of \textcolor{purple}{you}, needs to get rid of, take all of \textcolor{purple}{me}. \\
GPT-J-6B & romeo, \textcolor{purple}{take off your} name, and for that name, which is not part of \textcolor{purple}{you}, take all of \textcolor{purple}{me}. \\
\midrule
\textcolor{gray}{\bf{Input}} & \textcolor{gray}{now \textcolor{cyan}{art thou sociable}, now \textcolor{cyan}{art thou r}omeo; now \textcolor{cyan}{art thou} what \textcolor{cyan}{thou art}, by \textcolor{cyan}{art} as well as by nature.} \\
\cdashlinelr{1-2}
GPT-2-Small & now art thou romeo, now art thou romeo; now art thee what thou art, by art as well as by nature. \\
GPT-2-Medium & what I have said in thy speech thou hast said enough of me, wherewith to destroy thy soul. \\
GPT-2-Large & now art thou sociable, now art thou romeo; now art thou what thou art, by art as well as by nature. \\
GPT-2-XL & now is thou sociable, now is thou Romeo; now is thou what thou art, by nature as well as by nature. \\
GPT-Neo-1.3B & and now art thou what thou art, by art as well as by nature. \\
GPT-Neo-2.7B & now \textcolor{purple}{are you} sociable, now \textcolor{purple}{are you} romeo; now \textcolor{purple}{are you} what \textcolor{purple}{you are}, by nature. \\
GPT-J-6B & now \textcolor{purple}{you are friendly}, now \textcolor{purple}{you are Romeo}; now \textcolor{purple}{you are who you are}, by \textcolor{purple}{nature} as well as by \textcolor{purple}{art}. \\
\midrule
\textcolor{gray}{\bf{Input}} & \textcolor{gray}{the sweetest honey is \textcolor{cyan}{loathsome} in \textcolor{cyan}{his} own deliciousness and in the taste \textcolor{cyan}{confounds} the appetite.} \\
\cdashlinelr{1-2}
GPT-2-Small & the sweetest honey is loathsome in his own deliciousness and in the taste confounds the appetite. \\
GPT-2-Medium & sweetest honey, sweetest bread, sweetest wine, sweetest dainties, do them to the soul's delights. \\
GPT-2-Large & the sweetest honey is loathsome in his own deliciousness; in the taste confounds the appetite. \\
GPT-2-XL & the \textcolor{purple}{most sweet and clearest} honey is loathsome in \textcolor{purple}{the taste} and in the very \textcolor{purple}{smell} confounds the appetite. \\
\multirowcell{2}{GPT-Neo-1.3B} & the honey is sweeter than ever it was in the honeycomb, and is almost enough to make one lose their senses and die a corpse. \\
GPT-Neo-2.7B & the sweetest milk is foul and sickly in his own deliciousness and in the taste confounds the appetite. \\
GPT-J-6B & the sweetest honey is \textcolor{purple}{disgusting} in \textcolor{purple}{its} own deliciousness and in its taste \textcolor{purple}{it} confounds the appetite. \\
\midrule
\textcolor{gray}{\bf{Input}} & \textcolor{gray}{unless philosophy can make a \textcolor{cyan}{j}uliet, \textcolor{cyan}{displant} a town, reverse a prince's doom, it helps not, it prevails not.} \\
\cdashlinelr{1-2}
GPT-2-Small & what am I thinking of?\textsuperscript{*} \\
\multirowcell{2}{GPT-2-Medium} & if philosophy can make a juliet, supplant your \textcolor{purple}{King}, reverse his doom, reverse a prince's doom, reverse a \textcolor{purple}{king}'s doom. \\
GPT-2-Large & unless philosophy can make a juliet, displant a town, reverse a prince's doom, it helps not, it prevails not. \\
GPT-2-XL & unless it can convert a whore into a knight, it does not make it jolly. \\
GPT-Neo-1.3B & unless philosophy can make a juliet, a town, reverse a prince's doom, it helps not, it prevails not. \\
\multirowcell{2}{GPT-Neo-2.7B} & unless philosophy can make a \textcolor{purple}{J}uliet, \textcolor{purple}{cause a town to explode}, reverse a prince's \textcolor{purple}{downfall}, it helps not, it prevails not. \\
GPT-J-6B & unless philosophy can make a \textcolor{purple}{J}uliet, \textcolor{purple}{overturn} a town, reverse a prince's doom, it \textcolor{purple}{doesn't} help, it \textcolor{purple}{doesn't} work. \\
\midrule
\textcolor{gray}{\bf{Input}} & \textcolor{gray}{\textcolor{cyan}{night's candles are burnt out}, and \textcolor{cyan}{jocund} day \textcolor{cyan}{stands tiptoe} on the misty mountain tops.} \\
\cdashlinelr{1-2}
GPT-2-Small & not here \\
GPT-2-Medium & and \textcolor{purple}{when night comes}, the \textcolor{purple}{sun sets} and \textcolor{purple}{stars all flicker} in the distance. \\
GPT-2-Large & night's candles are burnt out, and jocund day stands \textcolor{purple}{in a field of snow}. \\
GPT-2-XL & night's candles are burn\textcolor{purple}{ing} out, and jocund day is stand\textcolor{purple}{ing still} on the misty mountain tops. \\
GPT-Neo-1.3B & night's \textcolor{purple}{fires} are out and the day \textcolor{purple}{stands up above} the foggy mountains. \\
GPT-Neo-2.7B & night's candles are burnt out; \textcolor{purple}{day's gone to bed and evening's at hand}; \textcolor{purple}{the sun is setting on} the mountain tops. \\
GPT-J-6B & \textcolor{purple}{day is over, and night is approaching}, and it stands on \textcolor{purple}{its tiptoes} over the mist\textcolor{purple}{ed} mountain tops. \\
\midrule
\textcolor{gray}{\bf{Input}} & \textcolor{gray}{nor tears nor prayers shall \textcolor{cyan}{purchase out abuses}, therefore use none.} \\
\cdashlinelr{1-2}
GPT-2-Small & nor tears nor prayers shall purchase out abuses, therefore use none \\
GPT-2-Medium & nor tears nor prayers shall purchase out abuses, therefore use none. \\
GPT-2-Large & use none, for abuse is in the eyes of a man. \\
GPT-2-XL & \textcolor{purple}{no} tears \textcolor{purple}{nor} prayers shall \textcolor{purple}{buy} out abuses, therefore \textcolor{purple}{I will not} use \textcolor{purple}{any}. \\
GPT-Neo-1.3B & nor tears nor prayers shall \textcolor{purple}{have any power}, therefore use none. \\
GPT-Neo-2.7B & nor prayers nor tears will \textcolor{purple}{do any good}, therefore use none. \\
GPT-J-6B & and tears and prayers \textcolor{purple}{won't work for you}, so \textcolor{purple}{don't use them}. \\
\midrule
\textcolor{gray}{\bf{Input}} & \textcolor{gray}{and here stands \textcolor{cyan}{all your state}: either \textcolor{cyan}{be gone before the watch be set}, or \textcolor{cyan}{by the break of day disguised from hence}.} \\
\cdashlinelr{1-2}
GPT-2-Small & and here stands all your state: either be gone before the watch be set, or by the break of day disguised from hence. \\
GPT-2-Medium & and this day there is nothing but the city. \\
GPT-2-Large & but where is the watch? \\
GPT-2-XL & and here stands all your state: either be gone before the watch, or by the break of day disguised from hence. \\
GPT-Neo-1.3B & and here stands all your state: either be gone before the watch be set, or by the break of day disguised from hence. \\
GPT-Neo-2.7B & and \textcolor{purple}{this is all you get}: either go before the \textcolor{purple}{watch is set}, or before the break of day \textcolor{purple}{dressed like a thief} from hence. \\
GPT-J-6B & and here \textcolor{purple}{is all your stuff}: either \textcolor{purple}{leave now} or \textcolor{purple}{you'll have to deal with us when it's morning}. \\
\bottomrule
\end{tabular}
}
\caption{Qualitative examples on \shakespeare-clean. We manually highlight in blue and red phrases that we find to be associated with the input and target styles, respectively. (Footnote \textsuperscript{*}: Pray tell us, what are you thinking of right now?)}
\label{tab:generations_shakespeare}
\end{table*}

\end{document}